\documentclass[lettersize,journal]{IEEEtran}
\usepackage{amsmath,amsfonts}
\usepackage{algorithmic}
\usepackage{algorithm}
\usepackage{array}
\usepackage[caption=false,font=normalsize,labelfont=sf,textfont=sf]{subfig}
\usepackage{textcomp}
\usepackage{stfloats}
\usepackage{url}
\usepackage{verbatim}
\usepackage{graphicx}
\usepackage{cite}
\usepackage{bm}
\usepackage{threeparttable}
\usepackage{multirow}      
\usepackage{multicol}
\usepackage{booktabs}
\usepackage{makecell}
\newcommand{\etal}{\textit{et al.}}
\newcommand{\ie}{\textit{i.e.}}

\begin{document}

\title{XGrad: Boosting Gradient-Based Optimizers with Weight Prediction}

\author{Lei Guan,~\IEEEmembership{Member,~IEEE,} Dongsheng Li, Yanqi Shi, Jian Meng
\thanks{Manuscript received April 19, 2021; revised August 16, 2021.}
\thanks{This work is sponsored in part by the National Natural Science Foundation of China under Grant 62025208, and in part by the State Administration of Science, Technology and Industry for National Defence under Grant WDZC20235250118.}
\thanks{Lei Guan is with the Department of Mathematics, National University of Defense Technology, Changsha, China. (E-mail: guanleimath@163.com)}
\thanks{ Dongsheng Li and Yanqi Shi are with the National Key Laboratory of Parallel and Distributed Computing, National University of Defense Technology. (E-mails: dsli@nudt.edu.cn,sconcer@outlook.com)} 
\thanks{Jian Meng is with the Department of Mathematics, National University of Defense Technology, Changsha, China. (E-mail:mengjian23@nudt.edu.cn)}
\thanks{Lei Guan is the corresponding author.}
}

\markboth{Journal of \LaTeX\ Class Files,~Vol.~14, No.~8, August~2021}%
{Shell \MakeLowercase{\textit{et al.}}: A Sample Article Using IEEEtran.cls for IEEE Journals}


\maketitle

\begin{abstract}
In this paper, we propose a general deep learning training framework XGrad which introduces weight prediction into the popular gradient-based optimizers to boost their convergence and generalization when training the deep neural network (DNN) models. In particular, ahead of each mini-batch training,  the future weights are predicted according to the update rule of the used optimizer and are then applied to both the forward pass and backward propagation. In this way, during the whole training period, the optimizer always utilizes the gradients w.r.t.  the future weights to update the DNN parameters, making the gradient-based optimizer achieve better convergence and generalization compared to the original optimizer without weight prediction.  XGrad is rather straightforward to implement yet pretty effective in boosting the convergence of gradient-based optimizers and the accuracy of DNN models. Empirical results concerning five popular optimizers including SGD with momentum, Adam, AdamW, AdaBelief, and AdaM3 demonstrate the effectiveness of our proposal. The experimental results validate that XGrad can attain higher model accuracy than the baseline optimizers when training the DNN models.  The code of XGrad will be available at: \url{https://github.com/guanleics/XGrad}.
\end{abstract}


\section{Introduction}
\IEEEPARstart{T}{he} training of deep neural network (DNN) models is to find the optimal parameters using an optimizer which has a decisive influence on the accuracy of the models. The gradient-based optimization methods are currently of core practical importance in deep learning as they can attain rapid training of modern deep neural network models. Among all gradient-based optimizers, stochastic gradient descent (SGD) with momentum~\cite{qian1999momentum,sutskever2013importance} and adaptive methods such as Adam~\cite{kingma2014adam} and AdamW~\cite{loshchilov2017decoupled} are the most popular optimizers and have become the default choices for training many DNN models including convolutional neural networks (CNN)~\cite{simonyan2014very,szegedy2015going,he2016deep,krizhevsky2017imagenet}, recurrent neural networks (RNN), graph neural networks (GNN)~\cite{li2022graph,chen2019graph,choudhary2021atomistic}, generative adversarial networks (GAN)~\cite{goodfellow2020generative} and lots of transformer-based DNN models such as Transformer~\cite{vaswani2017attention}, BERT~\cite{devlin2018bert}, Vision Transformer~\cite{dosovitskiy2020image,han2022survey} and GPT-2/3~\cite{radford2019language,brown2020language}, etc.

When training a DNN model using the gradient-based optimizers, each mini-batch training generally consists of one forward pass and one backward propagation, where the gradients w.r.t. all the parameters (also known as weights) are computed during the backward propagation. After that, the generated gradients are utilized by the optimizer to calculate the update values for all parameters, which are finally applied to updating the DNN weights.  We exemplify this using the training process of mini-batch SGD~\cite{bottou2018optimization}. We assume that the mini-batch size is $b$, the available neural network weights at the $t$-th iteration are $\bm \theta_{t-1}$, and the loss function is $f(\cdot)$. Given a mini-batch training data $({\bf x}_i, {\bf y}_i)$, the forward pass first computes the loss with ${\bf z}_i = f({\bm \theta}_{t-1}, {\bf x}_i, {\bf y}_i)$. Then, the gradients generated in the backward propagation can be calculated as
\begin{equation}
	\label{equ:gradient}
	{\bf g}_t = \frac{1}{b}\sum_{i=1}^{b}\nabla_{\bm \theta} f({\bm \theta}_{t-1}, {\bf z}_i).
\end{equation}
After that, given the learning rate $\gamma_t$, the DNN weights are updated via
\begin{equation}
	\label{equ:sgd}
	{\bm \theta}_{t} = {\bm \theta}_{t-1} - \gamma_t{\bf g}_t.  
\end{equation}

\begin{figure}[htb]
	\centering
	\includegraphics[width=.43\textwidth]{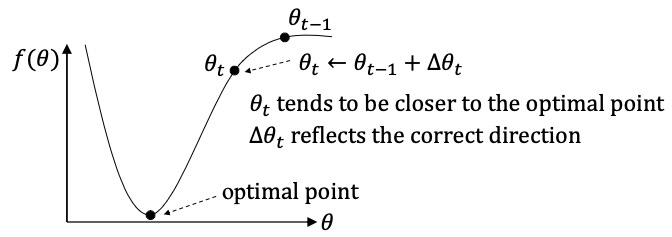}
	\caption{The optimization of DNN weights.}
	\label{fig:optimization}
\end{figure}

 For gradient-based optimization methods, the differences among different optimization methods lie in that the ways using the gradients generated by~\eqref{equ:gradient} to update model parameters are different. To be general, the weights are updated by $\bm\theta_{t} \leftarrow \bm\theta_{t-1} + \Delta \bm\theta_t$, where $\Delta \bm\theta_t$ denotes the relative increments of $\bm \theta_{t}$ over $\bm \theta_{t-1}$ and is computed by the used optimizer. Figure~\ref{fig:optimization} illustrates the optimization of DNN weights. The remarkable features of existing gradient-based optimizers can be summarized as follows. First, the updates of weights are continuous. Second, each mini-batch uses the currently available weights to do both forward pass and backward propagation. Third, in comparison with the current weights, the updated weights tend to be closer to the optimal point. In other words, in each mini-batch training, the weights tend to be updated in a ``correct'' direction to move towards the optimal point.

Motivated by the fact that DNN weights are updated in a continuous manner and the update values calculated by each gradient-based optimizer should reflect the ``correct'' direction for updating the weights, we introduce weight prediction~\cite{guan2019xpipe,chen2018efficient} into the DNN training and, in particular, propose a general framework XGrad to boost the convergence of gradient-based optimizers and to improve the accuracy of DNN models.  
The XGrad framework is not only very straightforward to implement but also works well for many commonly used deep learning optimizers. Previously, we explored weight prediction for DNN training in a short conference paper~\cite{guan2023weight} where we introduced weight prediction into the DNN training when using AdamW as an optimizer. In this paper, we study weight prediction in detail and enable it to cover a large class of gradient-based optimizers including SGD with momentum (SGDM) and adaptive methods such as RMSprop~\cite{tieleman2012lecture}, Adam~\cite{kingma2014adam}, AdamW~\cite{loshchilov2017decoupled}, AdaBelief~\cite{zhuang2020adabelief}, and AdaM3~\cite{wang2021rethinking}. 

Generally, given the available weights $\bm \theta_t$ and the learning rate $\gamma$, XGrad uses the following formula to predict the weights:
\begin{equation}
	\hat{\bm \theta}_{t+s} \approx {\bm \theta}_t - \gamma \cdot s \cdot\Delta {\bm \theta}_{t+1},
	\label{eq:xgrad_weight}
\end{equation}
where ${\hat {\bm \theta}_{t+s}}$ denote the approximately predicted weights used for both forward pass and backward propagation, $s$ is known as weight prediction steps, and $\Delta \bm \theta_t$ can be calculated according to the update rule of the used optimizers. In each mini-batch training, XGrad consists of the following three steps: 1) Ahead of the forward pass, cache the current weights $\bm \theta_t$ and apply Equation~\eqref{eq:xgrad_weight} to predict the future weights based on the specific update rule of the used gradient-based optimizer. 2) Use the predicted weights $\hat{\bm \theta}_{t+s}$ to perform both the forward pass and backward propagation after which the gradients w.r.t. the parameters of each layer are generated. 3) Recover the cached weights $\bm \theta_t$ and then update all parameters with the newly generated gradients.

We conduct extensive experiments to validate the effectiveness of our proposal. The experiment results demonstrate that XGrad can improve the model accuracy compared with the original optimizer. For example, XGrad achieves an average of 0.98\% top-1 accuracy improvement over the SGDM optimizer when training on the CIFAR-10 datset. Compared to Adam, XGrad also averages a 0.76\% accuracy improvement and obtains a 0.74 higher BLEU score when training GNMT-8 on WMT-16 EN$\rightarrow$De dataset. Similar conclusions can be drawn when comparing XGrad with AdamW, AdaBelief, and AdaM3.


The contributions of this paper can be summarized as follows:
\begin{itemize} 
	\item [(1)] We, for the first time, construct the mathematical relationship between currently available weights and future weights after several continuous updates when using six popular deep learning optimizers including SGDM, RMSprop, Adam, AdamW, AdaBelief, and AdaM3.
	\item [(2)] We devise a general workflow for incorporating weight prediction into the DNN training. To the best of our knowledge, this is the first time that applies weight prediction strategy to boost the convergence and generalization of popular gradient-based optimizers.
	
	
	\item [(3)] We conducted extensive experimental evaluations by using 19 different DNN models spanning image classification, natural language processing, and image generalization tasks to validate the effectiveness of our proposal. The experiment results demonstrate that XGrad works well for boosting the convergence and generalization of both SGDM and adaptive methods such as Adam, AdamW, AdaBelief, and AdaM3.
	
\end{itemize}

The rest of this paper is organized as follows:  Section~\ref{sec:grad} lays the foundation of weight prediction where we build up the mathematical relationship between the initial weights and the future weights after $s$ continuous updates when respectively using SGDM, RMSprop, Adam, AdamW, AdaBelief, and AdaM3 as the optimization method. Section~\ref{sec:xgrad} further constructs the XGrad framework on the basis of Section~\ref{sec:grad}.  Extensive experiments are conducted in Section~\ref{sec:experiment} to validate the effectiveness of our proposal. Following that, Section~\ref{sec:related_work} briefly presents the related research work about gradient-based optimizers. Finally, we conclude this paper and discuss the future work in Section~\ref{sec:conclusion}.

\section{Gradient-based Optimizers}\label{sec:grad}
In this section, we construct the mathematical relationship between the initial weights (denoted as ${\bm \theta}_0$) and the future weights after $s$ ($s>1$) times of continuous updates  (dubbed as ${{\bm \theta}_{s}}$) when training the DNN models using popular gradient-based optimizers including SGDM, RMSprop, Adam, AdamW, AdaBelief, and AdaM3. 

Ahead of any $t$-th ($t\geq1$) iteration, we assume that the current available DNN weights are ${\bm \theta}_{t-1}$.  Throughout this paper, we always let  $\gamma$ ($\gamma \in \mathbb{R}$) denote the learning rate and refer to $\lambda$ ($\lambda \in \mathbb{R}$) as the weight decay. 

\subsection{SGD with momentum}
We first reformulate the update of SGDM as
\begin{equation}
	\begin{aligned}
		& {\bm \theta}_{t} =  {\bm \theta}_{t-1} -\gamma \cdot {\bf v}_t, \\
		\textnormal{s.t.} 
		& \left\{\begin{array}{ll}
			\mathbf{g}_t = \nabla_{\bm \theta}f_t({\bm \theta}_{t-1}),\\
			\mathbf{g}_t = \mathbf{g}_t + \lambda {\bm \theta}_{t-1},\\
			{\mathbf v}_{t}= u\cdot \mathbf v_{t-1} + (1-\tau)\cdot \mathbf{g}_t.
		\end{array}\right.
	\end{aligned}
	\label{sgd_update}
\end{equation}
where $u$ is the momentum factor and $\tau$ is the dampening for momentum.

Letting ${\bm \theta}_0$ denote the initial weights of a DNN model, then in the following $s$ times of continuous mini-batch training, the DNN weights are updated  via
\begin{equation}
	\begin{aligned}
		& {\bm \theta}_1 = {\bm \theta}_0 - \gamma \cdot {\bf v}_1, \\
		& {\bm \theta}_2 = {\bm \theta}_1 - \gamma \cdot {\bf v}_2, \\
		& \cdots \\
		& {\bm \theta}_s = {\bm \theta}_{s-1} - \gamma \cdot {\bf v}_{s},\\
	\end{aligned}
	\label{sgd_update_series}
\end{equation}
where for any $ i \in\{1,2,\cdots, s\}$, we have
\begin{equation}
	\begin{aligned}
		\left\{\begin{array}{ll}
			\mathbf{g}_i = \nabla_{\bm \theta}f_i({\bm \theta}_{i-1}), \\
			\mathbf{g}_i = \mathbf{g}_i + \lambda {\bm \theta}_{i-1},\\
			\mathbf{v}_i = u\cdot \mathbf{v}_{i-1} + (1-\tau)\cdot \mathbf{g}_i. \\
		\end{array}\right.
	\end{aligned}
\end{equation}

When summing up all weight update equations in~\eqref{sgd_update_series}, we have
\begin{equation}
	\begin{aligned}
		& {\bm \theta}_s = {\bm \theta}_0 - \gamma\sum_{i=1}^{s}{\bf v}_i,\\
		\textnormal{s.t.}	& 
		\left\{\begin{array}{ll}
			\mathbf{g}_i = \nabla_{\bm \theta}f_i({\bm \theta}_{i-1}), \\
			\mathbf{g}_i = \mathbf{g}_i + \lambda {\bm \theta}_{i-1},\\
			\mathbf{v}_i = u\cdot \mathbf{v}_{i-1} + (1-\tau)\cdot \mathbf{g}_i. \\
		\end{array}\right.
	\end{aligned}
	\label{sgd_sum}
\end{equation}

\subsection{RMSprop}
We reformulate the update of RMSprop as
\begin{equation}
	\begin{aligned}
		& {\bm \theta}_{t} = {\bm \theta} _{t-1} - \frac{\gamma \cdot {\bf g}_t}{\sqrt{{\bf v}_t} + \epsilon}, \\
		\textnormal{s.t.} 
		& \left\{\begin{array}{ll}
			\mathbf{g}_t = \nabla_{\bm \theta}f_t({\bm \theta}_{t-1}),\\
			\mathbf{v}_t = \alpha\cdot \mathbf{v}_{t-1} + (1-\alpha)\cdot \mathbf{g}_t^2, \\
		\end{array}\right.
	\end{aligned}
	\label{rmsprop_update}
\end{equation}
where $a$ denotes the smoothing constant and $\epsilon$ (default: $1e^{-8}$) is used to improve numerical stability.

Likewise, during the first $s$ times of continuous mini-batch training, the DNN weights are updated  via
\begin{equation}
	\begin{aligned}
		& {\bm \theta}_1 = {\bm \theta}_0 - \frac{\gamma \cdot {\bf g}_1}{\sqrt{{\bf v}_1} + \epsilon}, \\
		& {\bm \theta}_2 = {\bm \theta}_1 - \frac{\gamma \cdot {\bf g}_2}{\sqrt{{\bf v}_2} + \epsilon}, \\
		& \cdots \\
		& {\bm \theta}_s = {\bm \theta}_{s-1} -\frac{\gamma \cdot {\bf g}_s}{\sqrt{{\bf v}_s} + \epsilon},\\
	\end{aligned}
	\label{rmsprop_update_series}
\end{equation}
where for any $ i \in\{1,2,\cdots, s\}$, we have
\begin{equation}
	\begin{aligned}
		\left\{\begin{array}{ll}
			\mathbf{g}_i = \nabla_{\bm \theta}f_i({\bm \theta}_{i-1}), \\
			\mathbf{v}_i = \alpha\cdot \mathbf{v}_{i-1} + (1-\alpha)\cdot \mathbf{g}_i^2.\\
		\end{array}\right.
	\end{aligned}
\end{equation}

When summing up all weight update equations in~\eqref{rmsprop_update_series}, we have
\begin{equation}
	\begin{aligned}
		& {\bm \theta}_s = {\bm \theta}_0 - \sum_{i=1}^s\frac{\gamma \cdot {\bf g}_i}{\sqrt{{\bf v}_i} + \epsilon},\\
		\textnormal{s.t.} & 
		\left\{\begin{array}{ll}
			\mathbf{g}_i = \nabla_{\bm \theta}f_i({\bm \theta}_{i-1}), \\
			\mathbf{v}_i = \alpha\cdot \mathbf{v}_{i-1} + (1-\alpha)\cdot \mathbf{g}_i^2. \\
		\end{array}\right.
	\end{aligned}
	\label{rmsprop_sum}
\end{equation}

\subsection{Adam}
We reformulate the update of Adam as
\begin{equation}
	\begin{aligned}
		& {\bm \theta}_{t} = {\bm \theta} _{t-1} - \frac{\gamma{\hat{\bf m}_t}}{\sqrt{\hat{\bf v}_t}+\epsilon}, \\
		\textnormal{s.t.} 
		& \left\{\begin{array}{ll}
			\mathbf{g}_t = \nabla_{\bm \theta}f_t({\bm \theta}_{t-1}),\\
			{\mathbf m}_{t}= \beta_1\cdot \mathbf m_{t-1} + (1-\beta_1) \cdot \mathbf{g}_t,\\
			\mathbf{v}_t = \beta_2\cdot \mathbf{v}_{t-1} + (1-\beta_2)\cdot \mathbf{g}_t^2, \\
			\hat{\mathbf m}_{t}= \frac{\mathbf m_{t}}{1-\beta_1^t},\\
			\hat{\mathbf v}_{t}= \frac{\mathbf v_t}{1-\beta_2^t}.
		\end{array}\right.
	\end{aligned}
	\label{adam_update}
\end{equation}
In~\eqref{adam_update}, ${\mathbf m}_{t}$ and ${\mathbf v}_{t}$ refer to the exponential moving average (EMA) of ${\mathbf g}_t$ and ${\mathbf g}_t^2$ respectively, $\beta_1$ and $\beta_2$ are coefficients used for computing ${\bf m}_t$ and ${\bf v}_t$ respectively, $\epsilon$ is the smoothing term that can prevent division by zero.

When training the DNN weights from ${\bm \theta}_0$, in the following $s$ times of continuous mini-batch training, the DNN weights are updated  via
\begin{equation}
	\begin{aligned}
		& {\bm \theta}_1 = {\bm \theta}_0 - \frac{\gamma{\hat{\bf m}_1}}{\sqrt{\hat{\bf v}_1}+\epsilon}, \\
		& {\bm \theta}_2 = {\bm \theta}_1 - \frac{\gamma{\hat{\bf m}_2}}{\sqrt{\hat{\bf v}_2}+\epsilon}, \\
		& \cdots \\
		& {\bm \theta}_s = {\bm \theta}_{s-1} - \frac{\gamma{\hat{\bf m}_s}}{\sqrt{\hat{\bf v}_s}+\epsilon},\\
	\end{aligned}
	\label{adam_update_series}
\end{equation}
where for any $ i \in\{1,2,\cdots, s\}$, we have
\begin{equation}
	\begin{aligned}
		\left\{\begin{array}{ll}
			\mathbf{g}_i = \nabla_{\bm \theta}f_i({\bm \theta}_{i-1}), \\
			{\mathbf m}_{i}= \beta_1\cdot \mathbf m_{i-1} + (1-\beta_1) \cdot \mathbf{g}_i,\\
			\mathbf{v}_i = \beta_2\cdot \mathbf{v}_{i-1} + (1-\beta_2)\cdot \mathbf{g}_i^2, \\
			\hat{\mathbf m}_{i}= \frac{\mathbf m_{i}}{1-\beta_1^i},\\
			\hat{\mathbf v}_{i}= \frac{\mathbf v_i}{1-\beta_2^i}. \\
		\end{array}\right.
	\end{aligned}
\end{equation}

When summing up all weight update equations in~\eqref{adam_update_series}, we have
\begin{equation}
	\begin{aligned}
		& {\bm \theta}_s = {\bm \theta}_0 - \sum_{i=1}^s\frac{\gamma \hat{\bf m}_i}{\sqrt{\hat{\bf v}_i}+\epsilon},\\
		\textnormal{s.t.} & 
		\left\{\begin{array}{ll}
			\mathbf{g}_i = \nabla_{\bm \theta}f_i({\bm \theta}_{i-1}), \\
			{\mathbf m}_{i}= \beta_1\cdot \mathbf m_{i-1} + (1-\beta_1) \cdot \mathbf{g}_i,\\
			\mathbf{v}_i = \beta_2\cdot \mathbf{v}_{i-1} + (1-\beta_2)\cdot \mathbf{g}_i^2, \\
			\hat{\mathbf m}_{i}= \frac{\mathbf m_{i}}{1-\beta_1^i},\\
			\hat{\mathbf v}_{i}= \frac{\mathbf v_i}{1-\beta_2^i}. \\
		\end{array}\right.
	\end{aligned}
	\label{adam_sum}
\end{equation}

\subsection{AdamW}
Given the momentum factor $\beta_1 \in \mathbb{R}$ and $\beta_2 \in \mathbb{R}$, we reformulate the update of AdamW~\cite{loshchilov2017decoupled} as
\begin{equation}
	\begin{aligned}
		& {\bm \theta}_{t} = (1-\gamma\lambda){\bm \theta} _{t-1} - \frac{\gamma{\hat{\bf m}_t}}{\sqrt{\hat{\bf v}_t}+\epsilon}, \\
		\textnormal{s.t.} 
		& \left\{\begin{array}{ll}
			\mathbf{g}_t = \nabla_{\bm \theta}f_t({\bm \theta}_{t-1}),\\
			{\mathbf m}_{t}= \beta_1\cdot \mathbf m_{t-1} + (1-\beta_1) \cdot \mathbf{g}_t,\\
			\mathbf{v}_t = \beta_2\cdot \mathbf{v}_{t-1} + (1-\beta_2)\cdot \mathbf{g}_t^2, \\
			\hat{\mathbf m}_{t}= \frac{\mathbf m_{t}}{1-\beta_1^t},\\
			\hat{\mathbf v}_{t}= \frac{\mathbf v_t}{1-\beta_2^t}.
		\end{array}\right.
	\end{aligned}
	\label{adamw_update}
\end{equation}

Likewise, the first $s$ times of continuous mini-batch training can be formulated as
\begin{equation}
	\begin{aligned}
		& {\bm \theta}_1 = (1-\gamma\lambda){\bm \theta}_0 - \frac{\gamma{\hat{\bf m}_1}}{\sqrt{\hat{\bf v}_1}+\epsilon}, \\
		& {\bm \theta}_2 = (1-\gamma\lambda){\bm \theta}_1 - \frac{\gamma{\hat{\bf m}_2}}{\sqrt{\hat{\bf v}_2}+\epsilon}, \\
		& \cdots \\
		& {\bm \theta}_s = (1-\gamma\lambda){\bm \theta}_{s-1} - \frac{\gamma{\hat{\bf m}_s}}{\sqrt{\hat{\bf v}_s}+\epsilon},\\
	\end{aligned}
	\label{adamw_update_series}
\end{equation}
where for any $ i \in\{1,2,\cdots, s\}$, we have
\begin{equation}
	\begin{aligned}
		\left\{\begin{array}{ll}
			\mathbf{g}_i = \nabla_{\bm \theta}f_i({\bm \theta}_{i-1}), \\
			{\mathbf m}_{i}= \beta_1\cdot \mathbf m_{i-1} + (1-\beta_1) \cdot \mathbf{g}_i,\\
			\mathbf{v}_i = \beta_2\cdot \mathbf{v}_{i-1} + (1-\beta_2)\cdot \mathbf{g}_i^2, \\
			\hat{\mathbf m}_{i}= \frac{\mathbf m_{i}}{1-\beta_1^i},\\
			\hat{\mathbf v}_{i}= \frac{\mathbf v_i}{1-\beta_2^i}. \\
		\end{array}\right.
	\end{aligned}
\end{equation}

When summing up all weight update equations in~\eqref{adamw_update_series}, we have
\begin{equation}
	\begin{aligned}
		& {\bm \theta}_s = {\bm \theta}_0 - \gamma\lambda\sum_{i=0}^{s-1}{\bm \theta}_i- \sum_{i=1}^s\frac{\gamma \hat{\bf m}_i}{\sqrt{\hat{\bf v}_i}+\epsilon},\\
		\textnormal{s.t.} & 
		\left\{\begin{array}{ll}
			\mathbf{g}_i = \nabla_{\bm \theta}f_i({\bm \theta}_{i-1}), \\
			{\mathbf m}_{i}= \beta_1\cdot \mathbf m_{i-1} + (1-\beta_1) \cdot \mathbf{g}_i,\\
			\mathbf{v}_i = \beta_2\cdot \mathbf{v}_{i-1} + (1-\beta_2)\cdot \mathbf{g}_i^2, \\
			\hat{\mathbf m}_{i}= \frac{\mathbf m_{i}}{1-\beta_1^i},\\
			\hat{\mathbf v}_{i}= \frac{\mathbf v_i}{1-\beta_2^i}. \\
		\end{array}\right.
	\end{aligned}
	\label{adamw_sum}
\end{equation}

It is well known that the weight decay value $\lambda$ is generally set to an extremely small value (e.g., $5e^{-4}$), and the learning rate $\gamma$ is commonly set to a value smaller than 1 (e.g., $1e^{-3}$). Consequently, $\gamma\lambda$ is pretty close to zero, and thus, the second term of the right-hand of~\eqref{adamw_sum} can be neglected. This, therefore, generates the following equation:
\begin{equation}
	\begin{aligned}
		& {\bm \theta}_s \approx {\bm \theta}_0  - \sum_{i=1}^s\frac{\gamma \hat{\bf m}_i}{\sqrt{\hat{\bf v}_i}+\epsilon}. \\
		\textnormal{s.t.} & 
		\left\{\begin{array}{ll}
			\mathbf{g}_i = \nabla_{\bm \theta}f_i({\bm \theta}_{i-1}), \\
			{\mathbf m}_{i}= \beta_1\cdot \mathbf m_{i-1} + (1-\beta_1) \cdot \mathbf{g}_i,\\
			\mathbf{v}_i = \beta_2\cdot \mathbf{v}_{i-1} + (1-\beta_2)\cdot \mathbf{g}_i^2, \\
			\hat{\mathbf m}_{i}= \frac{\mathbf m_{i}}{1-\beta_1^i},\\
			\hat{\mathbf v}_{i}= \frac{\mathbf v_i}{1-\beta_2^i}. \\
		\end{array}\right.
	\end{aligned}
	\label{adamw_sum_approx}
\end{equation}

\subsection{AdaBelief}
We reformulate the update of AdaBelief~\cite{zhuang2020adabelief} as
\begin{equation}
	\begin{aligned}
		& {\bm \theta}_{t} = {\bm \theta} _{t-1} - \frac{\gamma{\hat{\bf m}_t}}{\sqrt{\hat{\bf v}_t}+\epsilon}, \\
		\textnormal{s.t.} 
		& \left\{\begin{array}{ll}
			\mathbf{g}_t = \nabla_{\bm \theta}f_t({\bm \theta}_{t-1}),\\
			{\mathbf m}_{t}= \beta_1\cdot \mathbf m_{t-1} + (1-\beta_1) \cdot \mathbf{g}_t,\\
			\mathbf{v}_t = \beta_2\cdot \mathbf{v}_{t-1} + (1-\beta_2)\cdot (\mathbf{g}_t-\mathbf{m}_t)^2 + \epsilon, \\
			\hat{\mathbf m}_{t}= \frac{\mathbf m_{t}}{1-\beta_1^t},\\
			\hat{\mathbf v}_{t}= \frac{\mathbf v_t}{1-\beta_2^t}.
		\end{array}\right.
	\end{aligned}
	\label{adabelief_update}
\end{equation}
In~\eqref{adabelief_update}, ${\mathbf m}_{t}$ and ${\mathbf v}_{t}$ refer to the EMA of ${\mathbf g}_t$ and ${(\mathbf{g}_t-\mathbf{m}_t)^2}$ respectively, $\beta_1$ and $\beta_2$ are coefficients used for computing ${\bf m}_t$ and ${\bf v}_t$ respectively, $\epsilon$ is the smoothing term that can prevent division by zero.

When training the DNN weights from ${\bm \theta}_0$, in the following $s$ times of continuous mini-batch training, the DNN weights are updated  via
\begin{equation}
	\begin{aligned}
		& {\bm \theta}_1 = {\bm \theta}_0 - \frac{\gamma{\hat{\bf m}_1}}{\sqrt{\hat{\bf v}_1}+\epsilon}, \\
		& {\bm \theta}_2 = {\bm \theta}_1 - \frac{\gamma{\hat{\bf m}_2}}{\sqrt{\hat{\bf v}_2}+\epsilon}, \\
		& \cdots \\
		& {\bm \theta}_s = {\bm \theta}_{s-1} - \frac{\gamma{\hat{\bf m}_s}}{\sqrt{\hat{\bf v}_s}+\epsilon},\\
	\end{aligned}
	\label{adabelief_update_series}
\end{equation}
where for any $ i \in\{1,2,\cdots, s\}$, we have
\begin{equation}
	\begin{aligned}
		\left\{\begin{array}{ll}
			\mathbf{g}_i = \nabla_{\bm \theta}f_i({\bm \theta}_{i-1}), \\
			{\mathbf m}_{i}= \beta_1\cdot \mathbf m_{i-1} + (1-\beta_1) \cdot \mathbf{g}_i,\\
			\mathbf{v}_i = \beta_2\cdot \mathbf{v}_{i-1} + (1-\beta_2)\cdot (\mathbf{g}_i-\mathbf{m}_i)^2+\epsilon, \\
			\hat{\mathbf m}_{i}= \frac{\mathbf m_{i}}{1-\beta_1^i},\\
			\hat{\mathbf v}_{i}= \frac{\mathbf v_i}{1-\beta_2^i}. \\
		\end{array}\right.
	\end{aligned}
\end{equation}

When summing up all weight update equations in~\eqref{adabelief_update_series}, we have
\begin{equation}
	\begin{aligned}
		& {\bm \theta}_s = {\bm \theta}_0 - \sum_{i=1}^s\frac{\gamma \hat{\bf m}_i}{\sqrt{\hat{\bf v}_i}+\epsilon},\\
		\textnormal{s.t.} & 
		\left\{\begin{array}{ll}
			\mathbf{g}_i = \nabla_{\bm \theta}f_i({\bm \theta}_{i-1}), \\
			{\mathbf m}_{i}= \beta_1\cdot \mathbf m_{i-1} + (1-\beta_1) \cdot \mathbf{g}_i,\\
			\mathbf{v}_i = \beta_2\cdot \mathbf{v}_{i-1} + (1-\beta_2)\cdot (\mathbf{g}_i-\mathbf{m}_i)^2 + \epsilon, \\
			\hat{\mathbf m}_{i}= \frac{\mathbf m_{i}}{1-\beta_1^i},\\
			\hat{\mathbf v}_{i}= \frac{\mathbf v_i}{1-\beta_2^i}. \\
		\end{array}\right.
	\end{aligned}
	\label{adabelief_sum}
\end{equation}

\subsection{AdaM3}
We reformulate the update of AdaM3~\cite{wang2021rethinking} as
\begin{equation}
	\begin{aligned}
		& {\bm \theta}_{t} = {\bm \theta} _{t-1} - \frac{\gamma{\hat{\bf m}_t}}{\sqrt{\hat{\bf v}_t}}, \\
		\textnormal{s.t.} 
		& \left\{\begin{array}{ll}
			\mathbf{g}_t = \nabla_{\bm \theta}f_t({\bm \theta}_{t-1}),\\
			{\mathbf m}_{t}= \beta_1\cdot \mathbf m_{t-1} + (1-\beta_1) \cdot \mathbf{g}_t,\\
			\mathbf{v}_t = \beta_2\cdot \mathbf{v}_{t-1} + (1-\beta_2)\cdot \mathbf{m}_t^2 + \epsilon, \\
			\hat{\mathbf m}_{t}= \frac{\mathbf m_{t}}{1-\beta_1^t},\\
			\hat{\mathbf v}_{t}= \frac{\mathbf v_t}{1-\beta_2^t}.
		\end{array}\right.
	\end{aligned}
	\label{adam3_update}
\end{equation}
where $\eta$ is the learning rate decay and $\epsilon$ is used to improve numerical stability.

The DNN weights are updated via
\begin{equation}
	\begin{aligned}
		& {\bm \theta}_1 = {\bm \theta}_0 - \frac{\gamma{\hat{\bf m}_1}}{\sqrt{\hat{\bf v}_1}}, \\
		& {\bm \theta}_2 = {\bm \theta}_1 - \frac{\gamma{\hat{\bf m}_2}}{\sqrt{\hat{\bf v}_2}}, \\
		& \cdots \\
		& {\bm \theta}_s = {\bm \theta}_{s-1} - \frac{\gamma{\hat{\bf m}_s}}{\sqrt{\hat{\bf v}_s}},\\
	\end{aligned}
	\label{adam3_update_series}
\end{equation}
where for any $ i \in\{1,2,\cdots, s\}$, we have
\begin{equation}
	\begin{aligned}
		\left\{\begin{array}{ll}
			\mathbf{g}_i = \nabla_{\bm \theta}f_i({\bm \theta}_{i-1}), \\
			{\mathbf m}_{i}= \beta_1\cdot \mathbf m_{i-1} + (1-\beta_1) \cdot \mathbf{g}_i,\\
			\mathbf{v}_i = \beta_2\cdot \mathbf{v}_{i-1} + (1-\beta_2)\cdot \mathbf{m}_i^2+\epsilon, \\
			\hat{\mathbf m}_{i}= \frac{\mathbf m_{i}}{1-\beta_1^i},\\
			\hat{\mathbf v}_{i}= \frac{\mathbf v_i}{1-\beta_2^i}. \\
		\end{array}\right.
	\end{aligned}
\end{equation}

Summing up all equations in~\eqref{adam3_update_series} further yields
\begin{equation}
	\begin{aligned}
		& {\bm \theta}_s = {\bm \theta}_0 - \sum_{i=1}^s\frac{\gamma \hat{\bf m}_i}{\sqrt{\hat{\bf v}_i}},\\
		\textnormal{s.t.} & 
		\left\{\begin{array}{ll}
			\mathbf{g}_i = \nabla_{\bm \theta}f_i({\bm \theta}_{i-1}), \\
			{\mathbf m}_{i}= \beta_1\cdot \mathbf m_{i-1} + (1-\beta_1) \cdot \mathbf{g}_i,\\
			\mathbf{v}_i = \beta_2\cdot \mathbf{v}_{i-1} + (1-\beta_2)\cdot \mathbf{m}_i^2+\epsilon, \\
			\hat{\mathbf m}_{i}= \frac{\mathbf m_{i}}{1-\beta_1^i},\\
			\hat{\mathbf v}_{i}= \frac{\mathbf v_i}{1-\beta_2^i}. \\
		\end{array}\right.
	\end{aligned}
	\label{adam3_sum}
\end{equation}

\section{The XGrad framework}\label{sec:xgrad}
\subsection{Derivation of XGrad}
When summarizing Equations~\eqref{sgd_sum},~\eqref{rmsprop_sum},~\eqref{adam_sum}, \eqref{adamw_sum_approx}, \eqref{adabelief_sum}, and~\eqref{adam3_sum} in Section~\ref{sec:grad}, we can immediately reach a common conclusion. That is, given the initial weights $\bm \theta_0$, the weights after $s$ times of continuous updates can be approximately calculated via
\begin{equation}
	{\bm \theta}_s = {\bm \theta}_0  - \gamma \sum_{i=1}^s\Delta {\bm \theta}_i,
	\label{xgrad_sum}
\end{equation}
where $\Delta {\bm \theta}_i$ represents the relative increments of ${\bm \theta}_{i}$ over ${\bm \theta}_{i-1}$. In each iteration, the gradients of stochastic objective w.r.t. the $i$-th mini-batch, \ie \ $\mathbf{g}_i = \nabla_{\bm \theta} f_i({\bm \theta }_{i-1})$, can be calculated after the backward propagation is completed. Given ${\bf g}_i$, one can easily compute $\Delta{\bm \theta}_i$ according to the update rule of the used optimization method. 

Equation~\eqref{xgrad_sum} illustrates that given the initial weights $\bm \theta _0$, ${\bm \theta}_{s}$ is calculated by letting ${\bm \theta}_0$ subtract the learning rate times the sum of $s$ continuous relative variation of the weights. Correspondingly, given ${\bm \theta}_t$, the weights of any $t$-th ($t\ge 1$) iteration, the weights after $s$ times of continuous updates can be approximately calculated via
\begin{equation}
	{\bm \theta}_{t+s} = {\bm \theta}_t  - \gamma \sum_{i=t+1}^{t+s}\Delta {\bm \theta}_i.
	\label{xgrad_sum1}
\end{equation}

Equation~\eqref{xgrad_sum1} illustrates that ${\bm \theta}_{t+s}$ can be calculated by letting ${\bm \theta}_t$ subtract the sum of $s$ continuous relative variation of the weights when ${\bm \theta}_{t}$ is available. Note that when using an effective gradient-based optimizer, the relative increments of the weights in each iteration should reflect the trend of the weight updates. Consequently, $\Delta {\bm \theta}_i$ in \eqref{xgrad_sum1} should reflect the ``correct" direction for updating the weights ${\bm \theta}_{i-1}$ because $\Delta {\bm \theta}_i$ is calculated on the basis of the update rule of the optimizer and because the weights are updated in a continuous manner and along the way of inertia directions.

We can therefore replace $\sum_{i=t+1}^{t+s}\Delta {\bm \theta}_i$ in~\eqref{xgrad_sum1} with $s\cdot \Delta {\bm \theta}_{t+1}$  in an effort to approximately compute ${\bm \theta}_{t+s}$ for the case when only ${\bm \theta}_t$ and $\Delta {\bm \theta}_{t+1}$ are available. Letting ${\hat {\bm \theta}_{t+s}}$ denote the approximately predicted weights for ${\bm \theta}_{t+s}$, we get the weight prediction formula of XGrad as follows.
\begin{equation}
	\hat{\bm \theta}_{t+s} \approx {\bm \theta}_t - \gamma \cdot s \cdot\Delta {\bm \theta}_{t+1}.
	\label{weight_update}
\end{equation}

It is worth noting that when using Equation~\eqref{weight_update} as the weight prediction formula, $s$ denotes the weight prediction step which can be manually set and $\Delta {\bm \theta}_{t+1}$ can be easily computed according to the type of used optimizer (dubbed as base optimizer).  Table \ref{tab:delta_theta} summarizes the mathematical formulas of $\Delta {\bm \theta}_{t+1}$ for SGDM,  RMSprop, Adam, AdamW, AdaBelief, and AdaM3. Here, we note that XGrad directly uses the cached optimizer states of the $t$-th iteration step to calculate $\Delta \bm \theta_{t+1}$ so as to avoid computing the gradients when performing weight prediction. Notably, other gradient-based optimizers such as AdaGrad~\cite{duchi2011adaptive}, AdaBound~\cite{luo2019adaptive}, RAdam~\cite{liu2019variance}, and Lion~\cite{chen2023symbolic} can  be easily incorporated into the proposed framework.

\begin{table}[h]
	\centering
	\caption{The calculation of $\Delta {\bm \theta}_{t+1}$ for SGDM, RMSprop, Adam, AdamW, AdaBelief, and AdaM3.}
	\begin{threeparttable} 
		\begin{tabular}{{cc}} 
			\toprule
			Optimizer & $\Delta {\bm \theta}_{t+1}$ \\
			\midrule
			\makecell{SGDM} & \makecell{ $\begin{array}{ll}
					\quad \quad {\Delta\bm \theta}_{t+1} = \mathbf{v}_t,\\  \textnormal{s.t.} 
					\left\{\begin{array}{ll}
						\mathbf{g}_t = \nabla_{\bm \theta}f_t({\bm \theta}_{t-1}),\\
						{\mathbf v}_{t}= u\cdot \mathbf v_{t-1} + (1-\tau)\cdot \mathbf{g}_t.
					\end{array}\right.\\
				\end{array}$} \\
			\midrule
			RMSprop\tnote{*} & \makecell{$\begin{array}{ll}
					\quad \Delta {\bm \theta}_{t+1}=\frac{{\bf g}_t}{\sqrt{{\bf v}_t} + \epsilon}, \\ \textnormal{s.t.} 
					\left\{\begin{array}{ll}
						\mathbf{g}_t = \nabla_{\bm \theta}f_t({\bm \theta}_{t-1}), \\
						\mathbf{v}_t = \alpha\cdot \mathbf{v}_{t-1} + (1-\alpha)\cdot \mathbf{g}_t^2. \\
					\end{array}\right. \\
				\end{array}$}\\
			\midrule
			Adam\tnote{*} & \makecell{$\begin{array}{ll}
					\quad\quad	\Delta {\bm \theta}_{t+1} = \frac{\hat{\mathbf m}_{t}}{\sqrt{\hat{\mathbf v}_t} + \epsilon}, \\ \textnormal{s.t.} \ \left\{\begin{array}{ll}
						\mathbf{g}_t = \nabla_{\bm \theta}f_t({\bm \theta}_{t-1}), \\
						\mathbf{m}_t = \beta_1 \cdot \mathbf{m}_{t-1} + (1-\beta_1)\cdot \mathbf{g}_t, \\
						{\mathbf v}_{t}= \beta_2\cdot \mathbf v_{t-1} + (1-\beta_2) \cdot \mathbf{g}_t^2,\\
						\hat{\mathbf m}_{t}= \frac{\mathbf m_t}{1-\beta_1^t}, \\
						\hat{\mathbf v}_{t}= \frac{\mathbf v_{t}}{1-\beta_2^t}.\\
					\end{array}\right.
				\end{array}$}  \\
			\midrule
			AdamW\tnote{*} & \makecell{$\begin{array}{ll}
					\quad\quad	\Delta {\bm \theta}_{t+1} = \frac{\hat{\mathbf m}_{t}}{\sqrt{\hat{\mathbf v}_t} + \epsilon}, \\ \textnormal{s.t.} \ 	\left\{\begin{array}{ll}
						\mathbf{g}_t = \nabla_{\bm \theta}f_t({\bm \theta}_{t-1}), \\
						{\mathbf m}_{t}= \beta_1\cdot \mathbf m_{t-1} + (1-\beta_1) \cdot \mathbf{g}_t,\\
						\mathbf{v}_t = \beta_2\cdot \mathbf{v}_{t-1} + (1-\beta_2)\cdot \mathbf{g}_t^2, \\
						\hat{\mathbf m}_{t}= \frac{\mathbf m_{t}}{1-\beta_1^t},\\
						\hat{\mathbf v}_{t}= \frac{\mathbf v_t}{1-\beta_2^t}. \\
					\end{array}\right.
				\end{array}$}   \\
			\midrule
			AdaBelief & \makecell{$\begin{array}{ll}
					\quad\quad	\Delta {\bm \theta}_{t+1} = \frac{\hat{\mathbf m}_{t}}{\sqrt{\hat{\mathbf v}_t} + \epsilon}, \\ \textnormal{s.t.} \ 	\left\{\begin{array}{ll}
						\mathbf{g}_t = \nabla_{\bm \theta}f_t({\bm \theta}_{t-1}), \\
						{\mathbf m}_{t}= \beta_1\cdot \mathbf m_{t-1} + (1-\beta_1) \cdot \mathbf{g}_t,\\
						\mathbf{v}_t = \beta_2\cdot \mathbf{v}_{t-1} + (1-\beta_2)\cdot (\mathbf{g}_t - \mathbf{m}_t)^2+\epsilon, \\
						\hat{\mathbf m}_{t}= \frac{\mathbf m_{t}}{1-\beta_1^t},\\
						\hat{\mathbf v}_{t}= \frac{\mathbf v_t}{1-\beta_2^t}. \\
					\end{array}\right.
				\end{array}$}   \\
			\midrule
			AdaM3& \makecell{$\begin{array}{ll}
					\quad\quad	\Delta {\bm \theta}_{t+1} = \frac{\hat{\mathbf m}_{t}}{\sqrt{\hat{\mathbf v}_t}}, \\ \textnormal{s.t.} \ 	\left\{\begin{array}{ll}
						\mathbf{g}_t = \nabla_{\bm \theta}f_t({\bm \theta}_{t-1}), \\
						{\mathbf m}_{t}= \beta_1\cdot \mathbf m_{t-1} + (1-\beta_1) \cdot \mathbf{g}_t,\\
						\mathbf{v}_t = \beta_2\cdot \mathbf{v}_{t-1} + (1-\beta_2)\cdot \mathbf{m}_t^2+\epsilon, \\
						\hat{\mathbf m}_{t}= \frac{\mathbf m_{t}}{1-\beta_1^t},\\
						\hat{\mathbf v}_{t}= \frac{\mathbf v_t}{1-\beta_2^t}. \\
					\end{array}\right.
				\end{array}$}   \\
			\bottomrule
		\end{tabular}
		\begin{tablenotes}    
			\footnotesize               
			\item[*] ${\mathbf g}_i^2$ refers to element-wise square with $\mathbf g_i^2=\mathbf g_i\odot \mathbf g_i$.          \end{tablenotes}  
	\end{threeparttable}
	\label{tab:delta_theta}
\end{table}

\begin{algorithm}[h!]
	\centering
	\caption{The workflow of XGrad}
	\label{alg1}
	\begin{algorithmic}[1]
		\REQUIRE Weight prediction step $s$, other hyper-parameters such as $\gamma$,  $\beta_1$, $\beta_2$, $\epsilon$.
		\STATE {Initialize or load DNN weights $\bm \theta_0$.}
		\STATE {$t \leftarrow{0}$.}
		\WHILE {the stopping criterion is not met}
		\STATE{Cache the current weights $\bm \theta_{t}$.}
		\STATE{Compute $\Delta {\bm \theta}_{t+1}$ according to Table~\ref{tab:delta_theta}.}
		\STATE{Calculate $\hat{\bm \theta}_{t+s}$ using \eqref{weight_update}.}
		\STATE {Do forward pass with $\hat{\bm \theta}_{t+s}$.}
		\STATE {Do backward propagation with $\hat{\bm \theta}_{t+s}$.}
		\STATE {Recover the cached weights $\bm \theta_t$.}
		\STATE {Update the weights $\bm \theta_t$ using the specific optimizer. }
		\STATE{$t\leftarrow t+1$.}
		\ENDWHILE
	\end{algorithmic}
\end{algorithm}

\subsection{Workflow of XGrad}
In the following, we showcase how to incorporate weight prediction into DNN training. Algorithm \ref{alg1} illustrates the details of training the DNN models using the XGrad framework. The weight prediction steps $s$ and other hyper-parameters are required before the DNN training starts. At each iteration, the currently available weights ${\bm \theta}_t$ should be cached before the forward pass starts (Line 4).  Then the relative increments of ${\bm \theta}_{t+1}$ over ${\bm \theta}_{t}$   are computed according to the update rule of the used base optimizer by a quick lookup of Table~\ref{tab:delta_theta} (Line 5). After that, weight prediction is performed using Equation~\eqref{weight_update} to compute $\hat{\bm \theta}_{t+s}$ (Line 6).  Following that, the predicted weights  $\hat{\bm \theta}_{t+s}$ are directly applied to both forward pass and backward propagation (Lines 7 and 8). Then, the cached weights ${\bm \theta}_t$ are required to be recovered (Line 9). Finally, the weights are updated by the optimizer using the gradients generated in the backward propagation (Line 10).

\subsection{Analysis of XGrad}
In this section, we give a brief analysis of why XGrad should be more effective than the base optimizers without weight prediction. The key point is that XGrad can be regarded as an approximation of the extragradient (EG) method~\cite{korpelevich1976extragradient} where the weight prediction step plays the role of extrapolation step.

We consider the optimization of Equation~\eqref{equ:gradient}. At the $(t+1)$-th iteration, EG updates the weights with 
\begin{equation}
	\begin{aligned}
		& {\tilde {\bm \theta}}_{t} = {\bm \theta}_t - \gamma \cdot \nabla_{\bm \theta}f_t( {\bm \theta}_t), \\
		& {\bm \theta}_{t+1} = {\bm \theta}_t -  \gamma \cdot \nabla_{\bm \theta}f_t({\tilde {\bm \theta}_t}),\\
	\end{aligned}
	\label{eq:eg}
\end{equation}
where $\gamma$ is the learning rate and the computation of ${\tilde {\bm \theta}}_{t}$ is known as the extrapolation step.

Meanwhile, for any $(t+1)$-th iteration, XGrad actually does the following update:
\begin{equation}
	\begin{aligned}
		& {\hat {\bm \theta}}_{t}  \approx {\bm \theta}_t - \gamma \cdot s \cdot \Delta \bm\theta_{t+1}, \\
		& {\bm \theta}_{t+1} = {\bm \theta}_t -  \gamma \nabla_{\bm \theta}f_t({\hat {\bm \theta}_t}).\\
	\end{aligned}
	\label{eq:xgrad-eg}
\end{equation}

When comparing Equations~\eqref{eq:eg} and~\eqref{eq:xgrad-eg},  we can see that ${\tilde {\bm \theta}}_{t}$ and ${\hat {\bm \theta}}_{t}$ are very similar in their calculation formulas. The main differences between EM and XGrad lie in the calculation of ${\tilde {\bm \theta}}_{t}$ and ${\hat {\bm \theta}}_{t}$. EG computes ${\tilde {\bm \theta}}_{t}$ by directly performing one time of gradient descent. In contrast, XGrad computes ${\hat {\bm \theta}}_{t}$ by mainly utilizing $\Delta \bm \theta_{t+1}$, which is computed according to the update rule of the used optimizer and also incorporates the gradient information. 

Similar to ${\tilde {\bm \theta}}_{t}$, the calculation of ${\hat {\bm \theta}}_{t}$ can be regarded as an extrapolation step before performing the update of $\bm \theta_t$. As a result, we can see that XGrad can be understood as an approximate approach to EG, where performing weight prediction can be seen as an extrapolation step in EG. It is well-known that EG can address many of the non-convergence issues that afflict gradient descents thanks to the extrapolation step~\cite{hsieh2020explore}. Therefore, it is very natural to expect that XGrad should be more effective than the base optimizers without weight prediction.

Here, we also note that EG requires computing two gradients for updating the model weights $\bm \theta_t$. In contrast, XGrad only requires computing one time of gradients because $\Delta \bm \theta_{t+1}$ can be directly computed by utilizing the cached optimizer states (e.g., ${\mathbf v}_t$ for SGDM in Table~\ref{tab:delta_theta}) of the previous iteration.




\section{Experiments}\label{sec:experiment}

\subsection{Experimental Settings}~\label{sec:exp-setting}

\subsubsection{Environments}
All experiments were conducted on two machines. The first machine is a multi-GPU platform that is equipped with four NVIDIA Tesla P100 GPUs, each with 16 GB of memory size. The second machine includes one NVIDIA Tesla M40 GPU with 24 GB of memory size. The CPU of both machines is an Intel Xeon E5-2680 with 128GB DDR4-2400 off-chip main memory.  We used PyTorch (v1.12.0) to train all DNN models.

\subsubsection{Benchmark models}
For experimental evaluations, we used 18 different benchmark models spanning image classification, natural language processing (NLP), and image generalization tasks. For the image classification tasks, we used nine CNN models and Vision Transformer (ViT)~\cite{dosovitskiy2020image} as the benchmark DNN models. The evaluated CNN models include LeNet~\cite{lecun1998gradient}, AlexNet~\cite{krizhevsky2017imagenet}, VGG-11~\cite{simonyan2014very}, VGG-16, ResNet-34~\cite{he2016deep}, ResNet-101, GoogleNet~\cite{szegedy2015going}, DenseNet-121~\cite{huang2017densely}, and Inception-V3~\cite{szegedy2016rethinking}. The ViT model is configured with 6 transformer blocks and a patch size of 4. The number of heads in the multi-head attention layer is 8 and the dimension of the MLP layer is 512. Both the dropout rate and embedding dropout rate are set to 0.1. The NLP tasks consist of three parts. For the language modeling task, we trained LSTM models with 1, 2, and 3 layers (denoted as LSTM-1, LSTM-2, and LSTM-3). Each layer was configured with an embedding size of 400 and 1,150 hidden activations. For machine translation task, we evaluated GNMT~\cite{wu2016google} with 8 LSTM layers and 16 LSTM layers (respectively denoted as GNMT-8 and GNMT-16). Furthermore, we also used BERT~\cite{devlin2018bert} with 12 Transformer blocks, 768 hidden sizes, and 12 self-attention blocks (denoted as BERT$_{\text{BASE}}$) as the NLP benchmark model. For image generalization tasks, we used VAE~\cite{kingma2013auto} and Wassertein-GAN (WAGN)~\cite{arjovsky2017wasserstein} as the evaluated benchmark models. 

\subsubsection{Datasets}
We used three image datasets for image classification tasks, including Fashion-MNIST~\cite{xiao2017fashion}, CIFAR-10~\cite{krizhevsky2009learning}, and CIFAR-100. For NLP tasks, the used datasets include Penn Treebank Dataset~\cite{marcinkiewicz1994building}, WMT-16~\cite{sennrich2016edinburgh} English-to-German (WMT En$\rightarrow$De), and Microsoft Research Paraphrase Corpus (MRPC). We note that MRPC belongs to the General Language Understanding Evaluation (GLUE)~\cite{wang2018glue} benchmarks. For image generalization tasks, we used MNIST and CIFAR-10 as the benchmark datasets. Table~\ref{table:model-datasets} summarizes all the evaluated DNN models and the used datasets in the experiments.

\begin{table}[!h]
	\centering
	\caption{Summary of evaluated DNN models and the datasets used in the experiments .}
	\label{table:model-datasets}
	\setlength{\tabcolsep}{1.6mm}
	\begin{tabular}{ccc}
		\toprule
		Tasks & Models & Data Sets    \\
		\midrule
		\multirow{4}{*}{\makecell{Image \\ Classification}} & LeNet & FashionMNIST  \\
		  & \makecell{Alexnet, VGG-11, VGG-16, \\ ResNet-34, ResNet-101, GoogleNet, \\ DenseNet-121, Inception-V3, ViT} & \makecell{CIFAR-10, \\ CIFAR-100} \\
		\hline
		\multirow{3}{*}{\makecell{NLP}}& \makecell{LSTM-1, LSTM-2, LSTM-3} & Penn Treebank \\
		& \makecell{GNMT-8, GNMT-16} & WMT-16 En$\rightarrow$De \\
		& BERT$_{\text{BASE}}$& MRPC \\
		\hline
		\multirow{2}{*}{\makecell{Image \\ Generalizaiton}} & VAE &  MNIST  \\
		 & WGAN & CIFAR-10 \\
		\bottomrule
	\end{tabular}
\end{table}

\subsubsection{Comparison settings}
To validate the effectiveness of our proposal, we mainly compare XGrad with five widely used optimizers including SGDM, Adam, AdamW, AdaBelief, and AdaM3. We evaluated our proposal with four different weight prediction steps (\ie, $s=1$, $s=2$, $s=3$, and $s=4$), where we respectively denoted them as XGrad-S1, XGrad-S2, XGrad-S3, and XGrad-S4 for convenience purposes.  We divided the comparisons into five groups, that is, SGDM vs. XGrad, Adam vs. XGrad, AdamW vs. XGrad, AdaBelief vs. XGrad, and AdaM3 vs. XGrad. Table~\ref{table:mthods-comp} summarizes the comparison settings. It's worth noting that XGrad automatically selects the base optimizer according to the type of compared optimizer. For example, when comparing with SGDM, XGrad automatically selected SGDM as the base optimizer and means SGDM with weight prediction. When comparing with Adam, AdamW, AdaBelief, and AdaM3, XGrad correspondingly means Adam, AdamW, AdaBelief, and AdaM3 with weight prediction, respectively.


\begin{table}[!h]
	\centering
	\caption{Summarization of comparison groups.}
	\label{table:mthods-comp}
	\setlength{\tabcolsep}{2.5mm}
	\begin{tabular}{ccc}
		\toprule
		No. & Comparisons  & Base optimizer of XGrad   \\
		\midrule
		1 & SGDM vs.  XGrad-S1/S2/S3/S4  &  SGDM \\
		2 & Adam vs.  XGrad-S1/S2/S3/S4 & Adam \\
		3 &  AdamW vs. XGrad-S1/S2/S3/S4 & AdamW \\
		4 &  AdaBelief vs. XGrad-S1/S2/S3/S4 & AdaBelief \\
		5 &  AdaM3 vs. XGrad-S1/S2/S3/S4 & AdaM3 \\
		\bottomrule
	\end{tabular}
\end{table}



\subsubsection{Model accuracy measurement}~\label{secsec:meansure}
For image classification tasks, the model accuracy means the top-1 accuracy (higher is better). For training LSTM-1, LSTM-2, and LSTM-3, the model accuracy is denoted by the perplexity (lower is better). For training GNMT models, the model accuracy refers to the BLEU score (higher is better). For training BERT$_{\text{BASE}}$ on MRPC, the model accuracy is defined as Dev set accuracy. For training VAE, we use the sum of reconstruction error and Kullback-Leibler divergence, denoted as total loss (lower is better), to assess the model accuracy. For training WGAN, we computed the Frechet Inception Distance (FID) score (lower is better) between the fake images and the real dataset to assess the generative models.

\subsection{Convergence and generalization} 

\subsubsection{Comparisons of XGrad and SGDM}\label{sec:comp-xgrad-sgdm}
In this section, we report the experimental results when training eight CNN models on the CIFAR-10 and CIFAR-100 datasets respectively. For all CNN models and ViT, we trained them for 200 epochs with a mini-batch size of 128. We initialized the learning rate with $1e^{-2}$ and decreased it by a factor of 10 at the 120th and 150th epochs. For both XGrad and SGDM, we set momentum with 0.9 and weight decay with $5e^{-4}$.

\begin{figure*}[h!]
	\centering
	\subfloat[AlexNet]{
		\includegraphics[width=.28\textwidth]{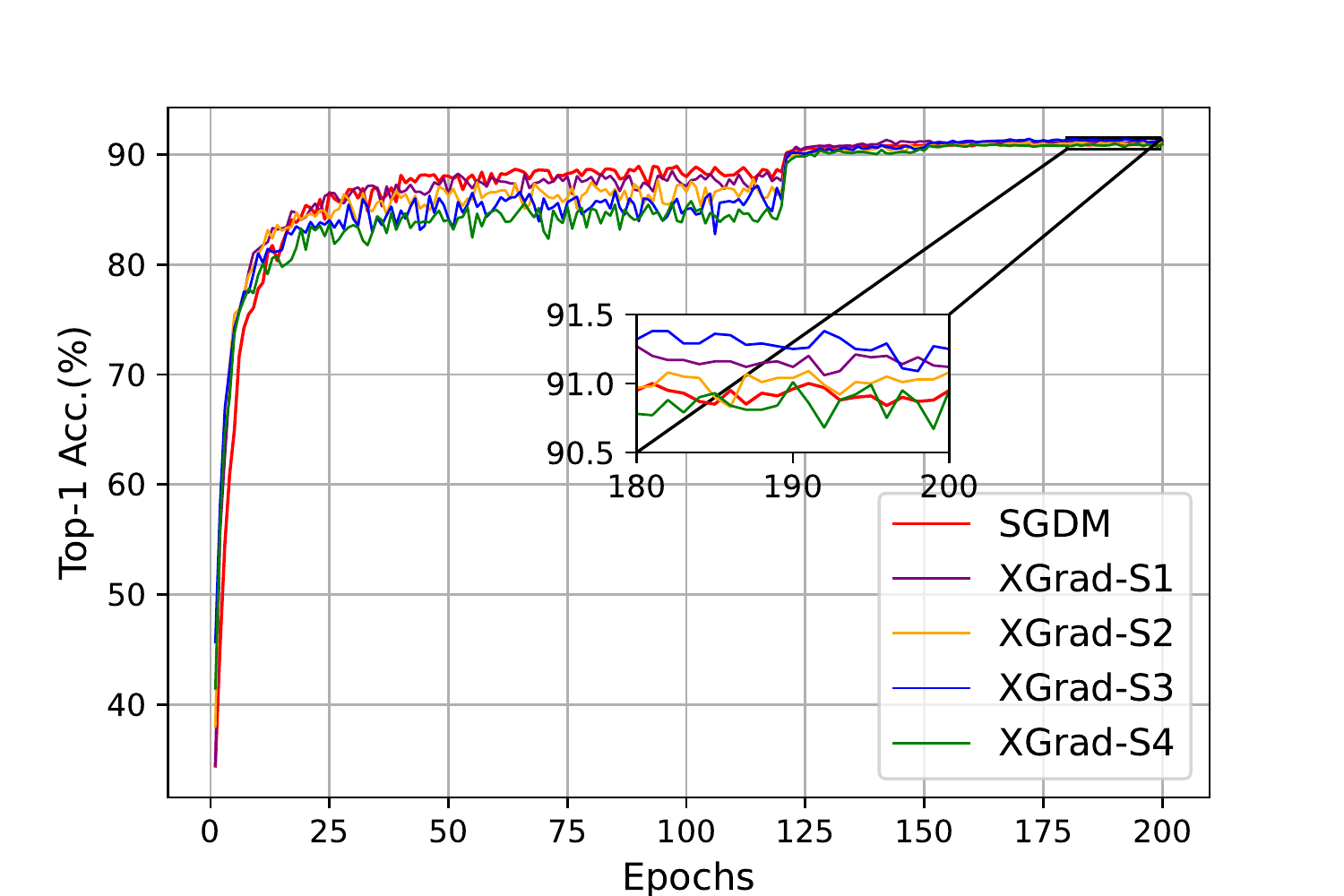}\label{comp-alexnet-acc}}
	\subfloat[VGG-11]{\includegraphics[width=.28\textwidth]{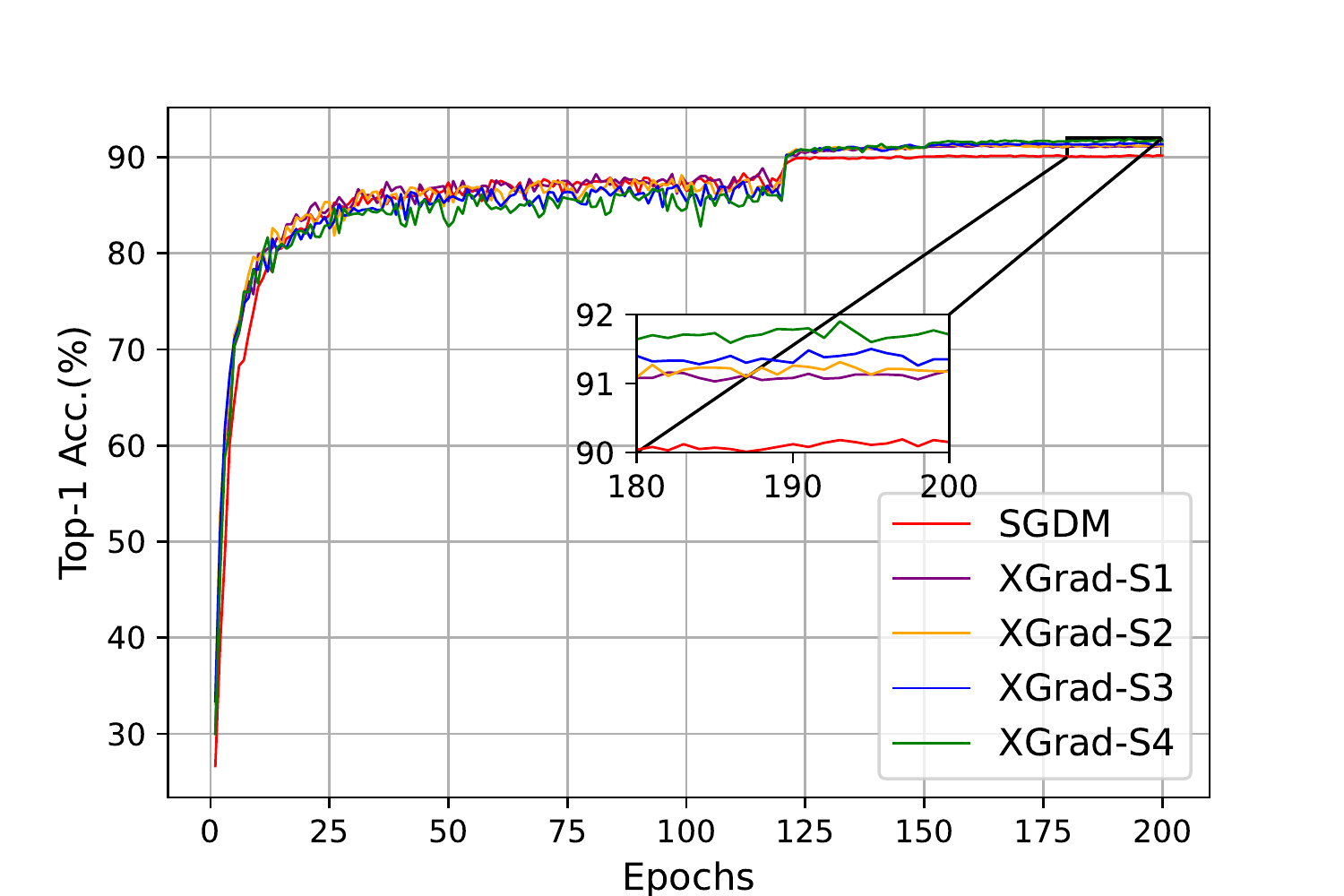}\label{comp-vgg11-acc}}
	\subfloat[VGG-16]{\includegraphics[width=.28\textwidth]{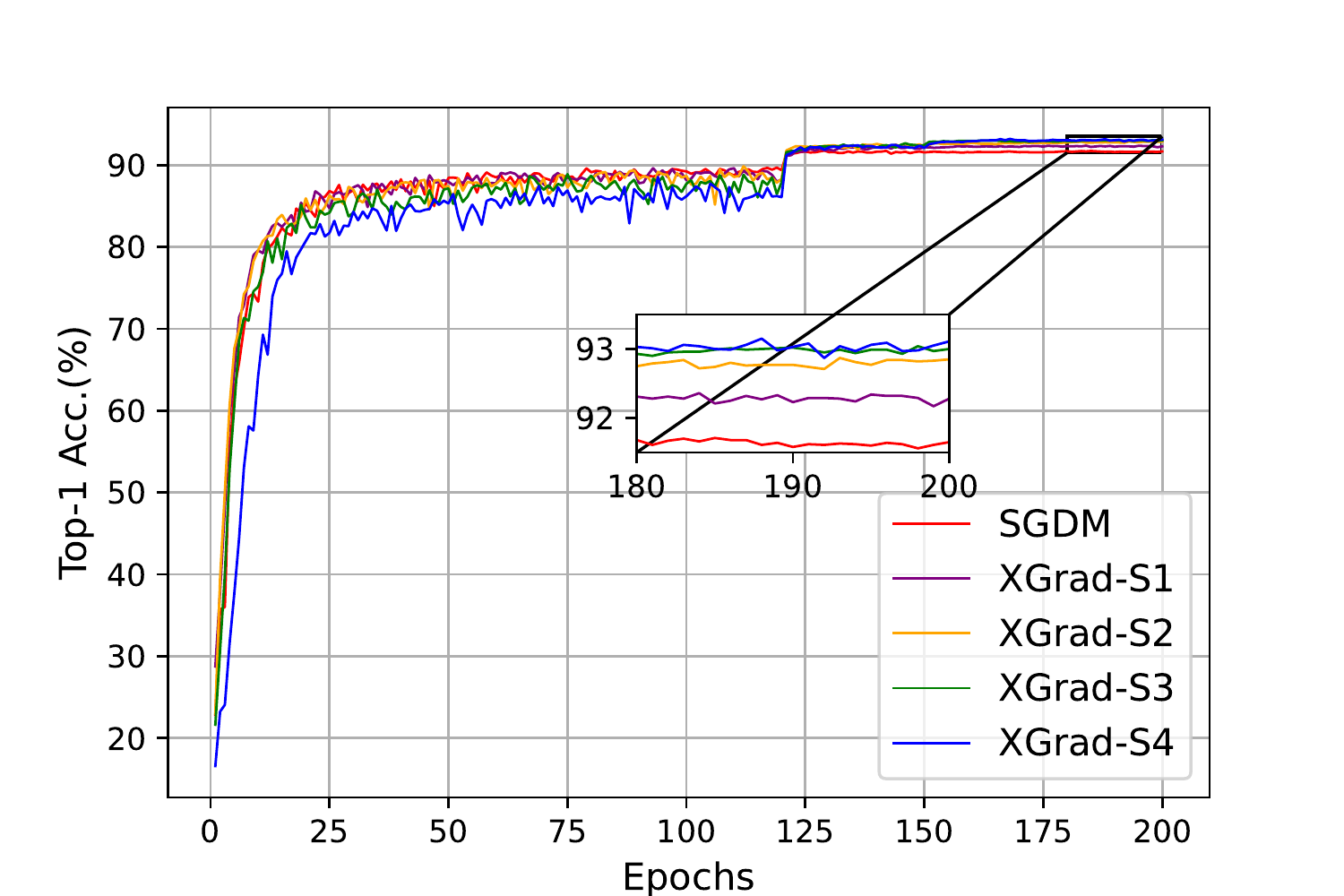}\label{comp-vgg16-acc}}
	\quad
	\subfloat[ResNet-34]{\includegraphics[width=.28\textwidth]{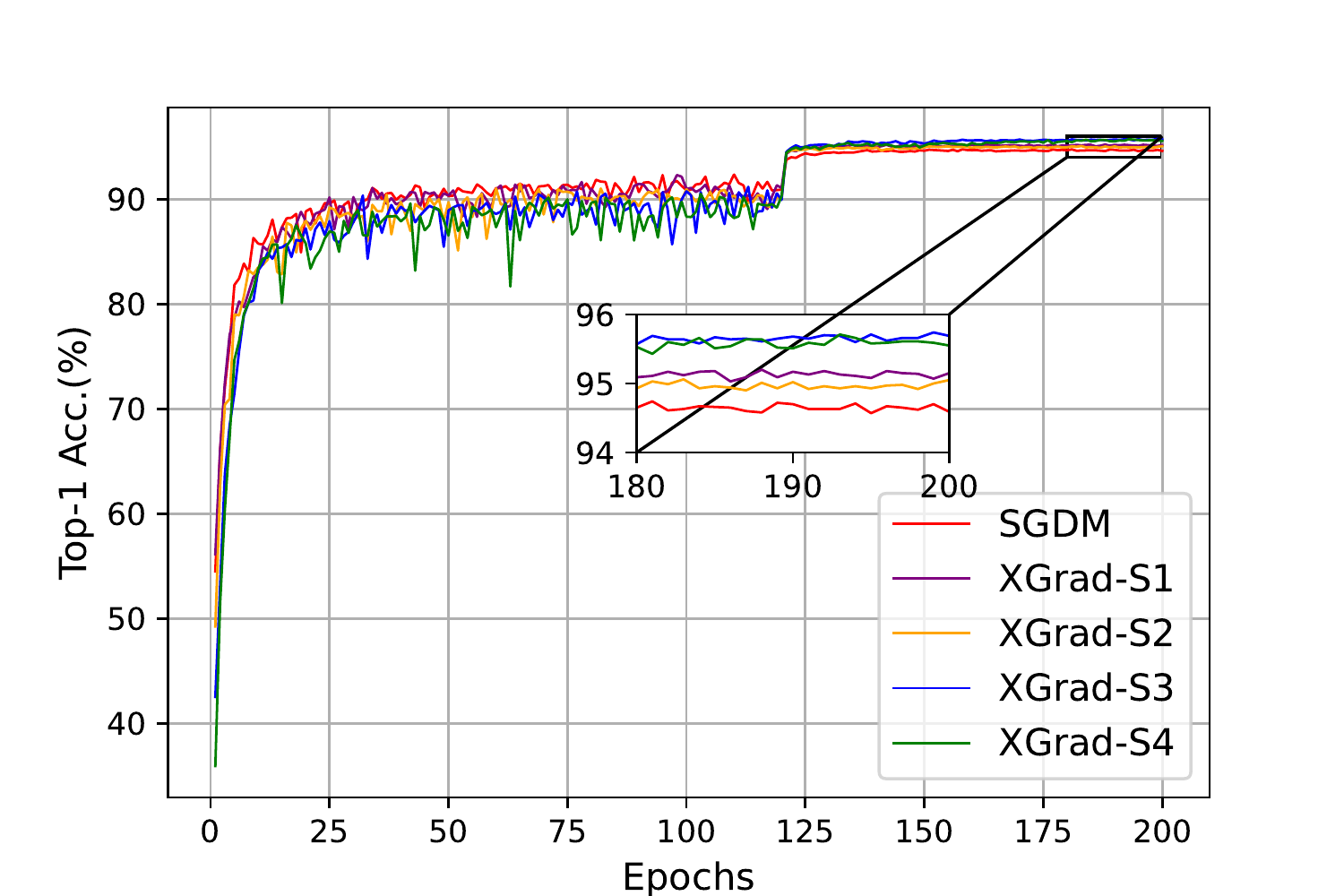}\label{comp-resnet34-acc}}
	\subfloat[ResNet-101]{\includegraphics[width=.28\textwidth]{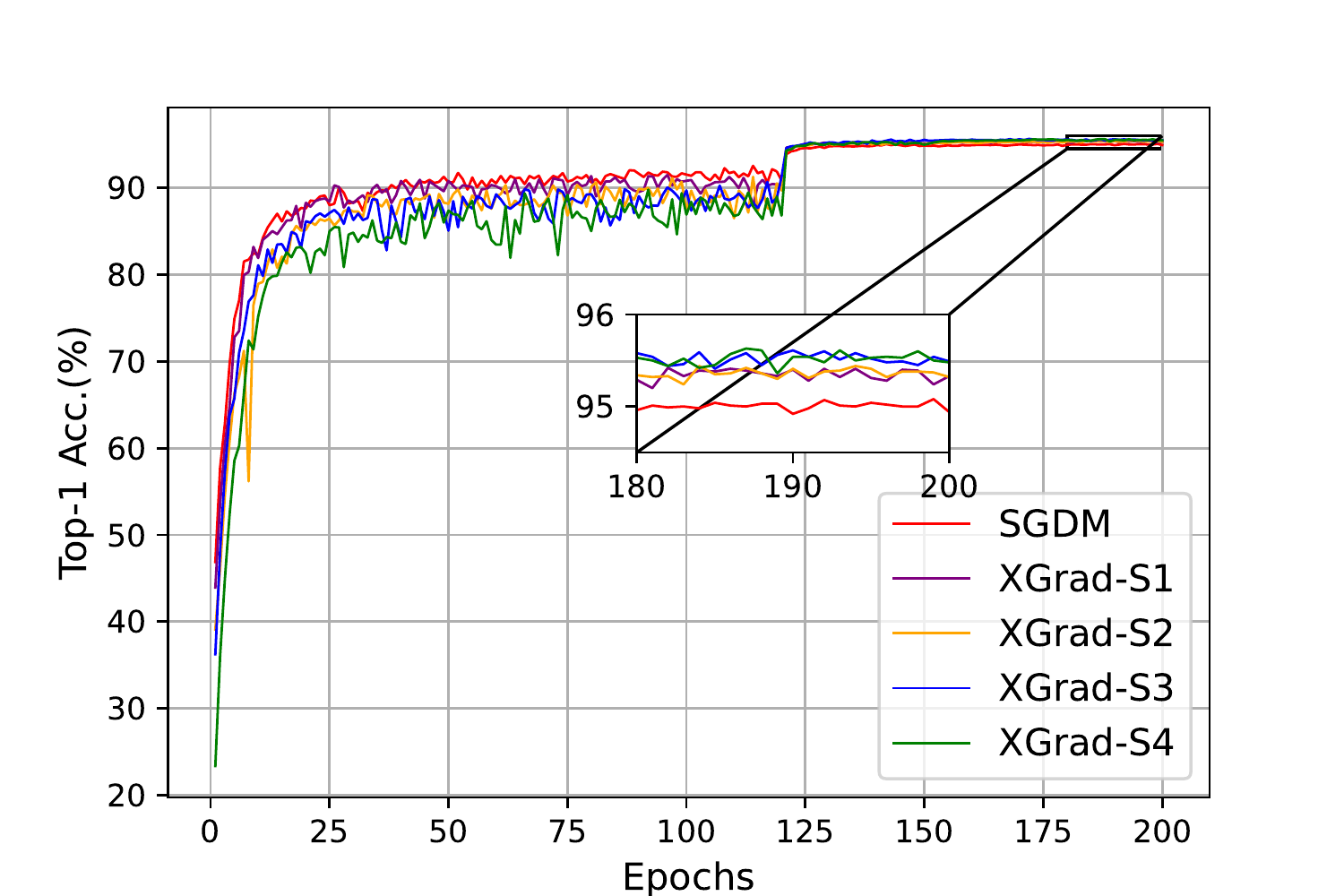}\label{comp-resnet101-acc}}
	\subfloat[GoogleNet]{\includegraphics[width=.28\textwidth]{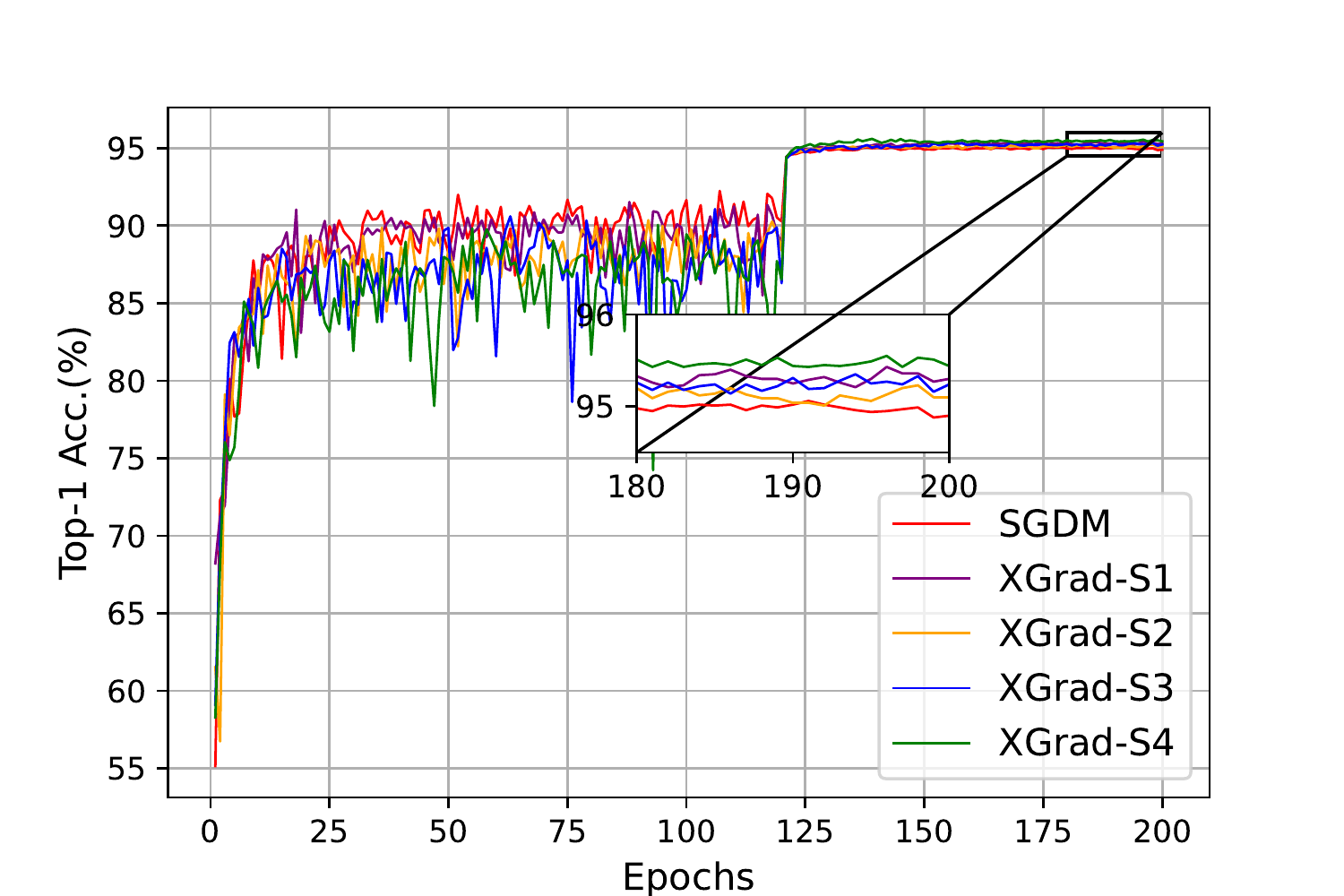}\label{comp-googlenet-acc}}
	\quad
	\subfloat[DenseNet-121]{\includegraphics[width=.28\textwidth]{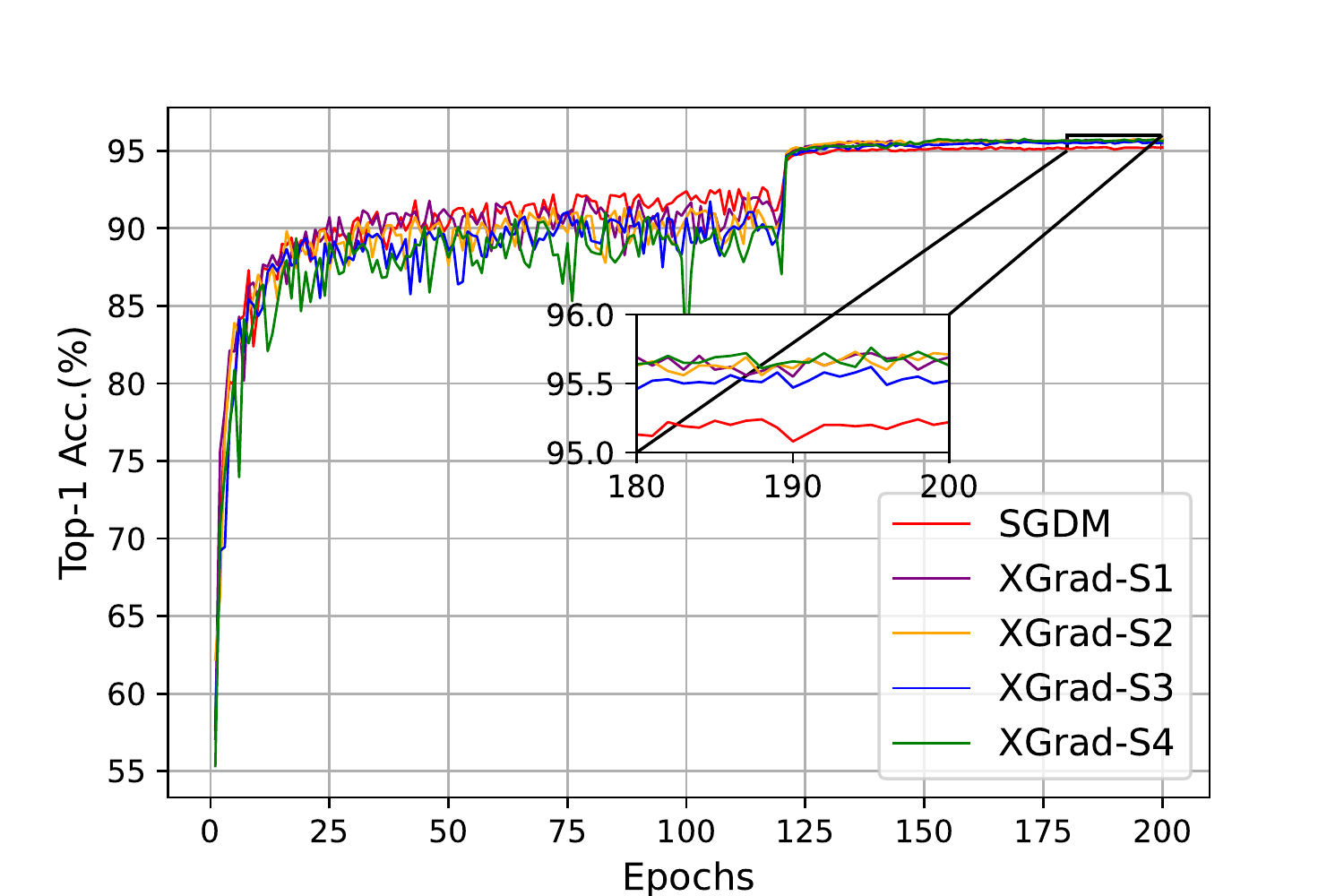}\label{comp-densenet-acc}}
	\subfloat[Inception-V3]{\includegraphics[width=.28\textwidth]{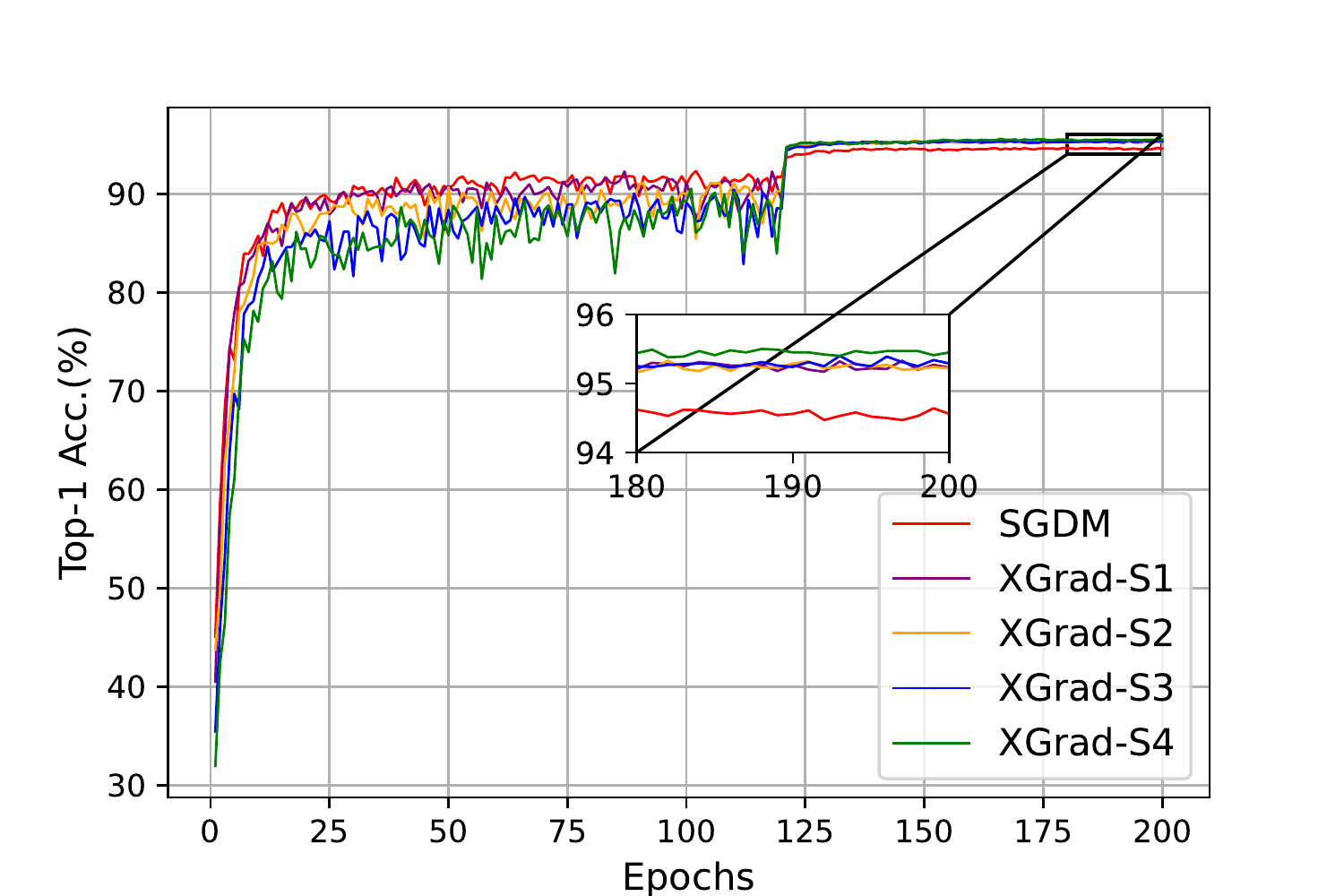}\label{comp-inceptionv3-acc}}
	\subfloat[ViT]{\includegraphics[width=.28\textwidth]{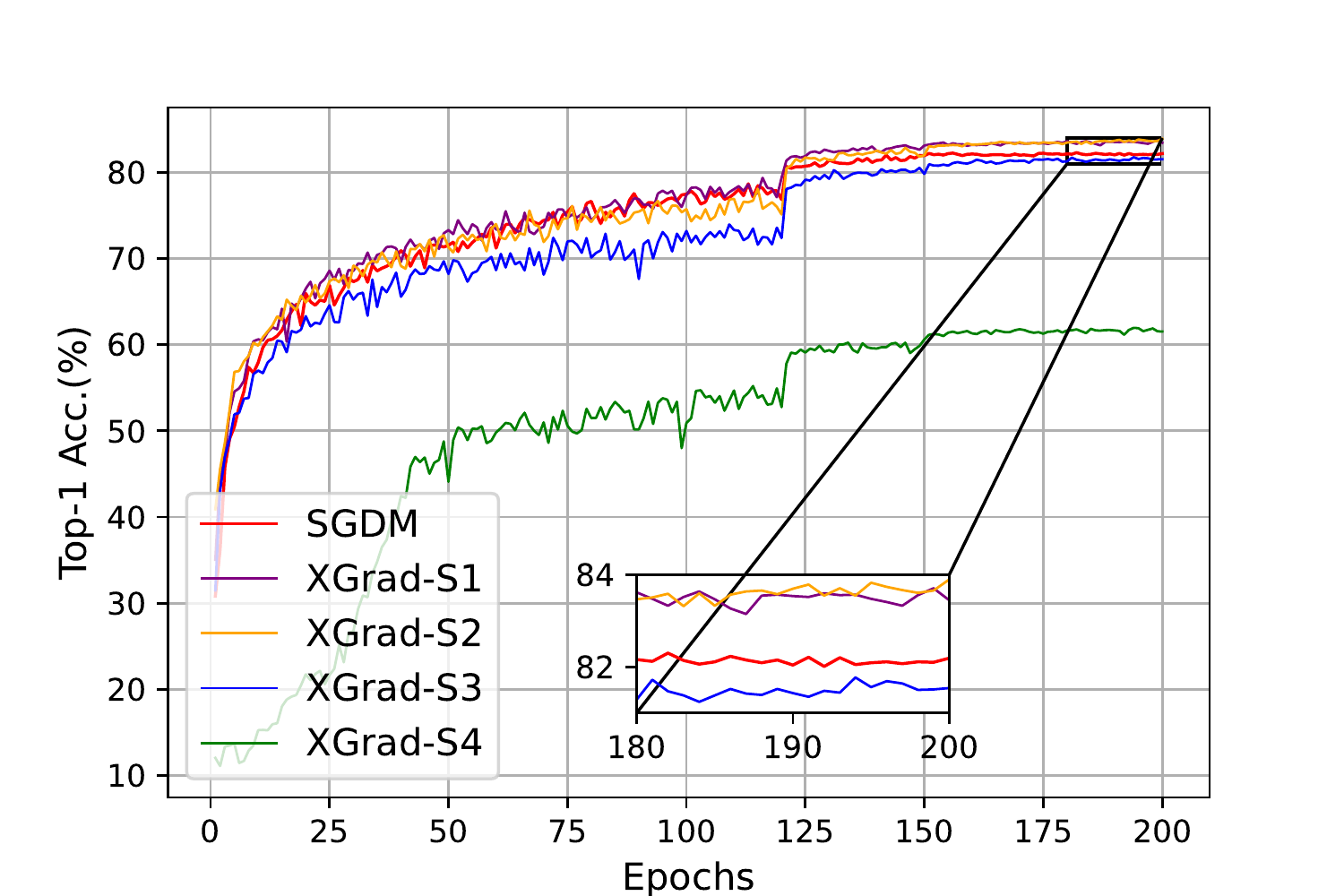}\label{comp-vit-cifar10-acc}}
	\caption{Validation Top-1 accuracy (higher is better) vs. epochs when training CNNs and ViT on CIFAR-10.}
	\label{comp-acc-cifar10}
\end{figure*}

\begin{table*}[h!]
	\centering
	\caption{Comparisons of XGrad and SGDM when training on CIFAR-10. The best results are highlighted in boldface.}
	\label{table:sgd-cifar10}
	\setlength{\tabcolsep}{2.8mm}
	\begin{tabular}{c|ccccccccc}
		\toprule
		Optimizers &  AlexNet & VGG-11 &  VGG-16  & ResNet-34   & ResNet-101  & GoogleNet & DenseNet-121&  Inception-V3 & ViT  \\
		\midrule
		\multicolumn{10}{c}{Maximum Top-1 Accuracy } \\
		\midrule
		SGDM & 91.00\% & 90.09\% & 91.76\% & 94.74\% & 95.08\%&  95.07\%& 95.25\%& 94.64\% &  82.30\% \\
		XGrad-S1 &91.32\%  & 91.23\% & 92.36\% & 95.21\% & 95.42\%& 95.43\% & 95.72\%& 95.36\% & 83.71\%\\
		XGrad-S2&91.09\% & 91.34\% &  92.87\% &  95.06\% & 95.44\%& 95.23\%& 95.73\% & 95.39\% & \textbf{83.90}\% \\
		XGrad-S3 & \textbf{91.42}\% & 91.50\% & 93.04\% &   \textbf{95.74}\%  & \textbf{95.65}\%&  95.37\%& 95.62\%& 95.40\%  & 81.77\% \\
		XGrad-S4 & 91.01\% & \textbf{91.90}\% &  \textbf{93.19}\%&  95.71\%& 95.63\%& \textbf{95.60}\%& \textbf{95.77}\%& \textbf{95.56}\% & 61.96\% \\
		\bottomrule
	\end{tabular}
\end{table*}

\begin{figure*}[h!]
	\centering
	\subfloat[AlexNet]{\includegraphics[width=.28\textwidth]{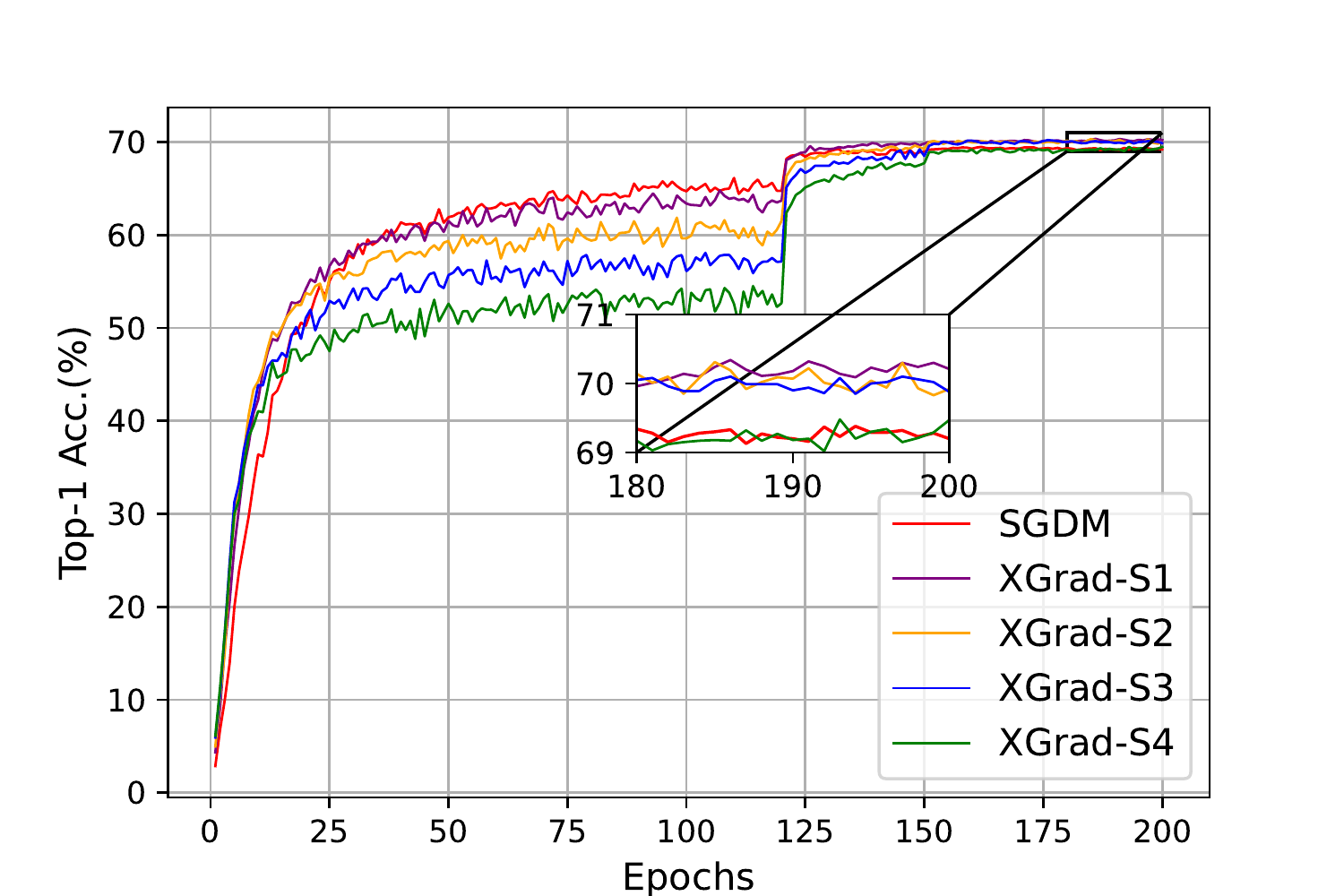}\label{comp-alexnet-cifar100-acc}}
	\subfloat[VGG-11]{\includegraphics[width=.28\textwidth]{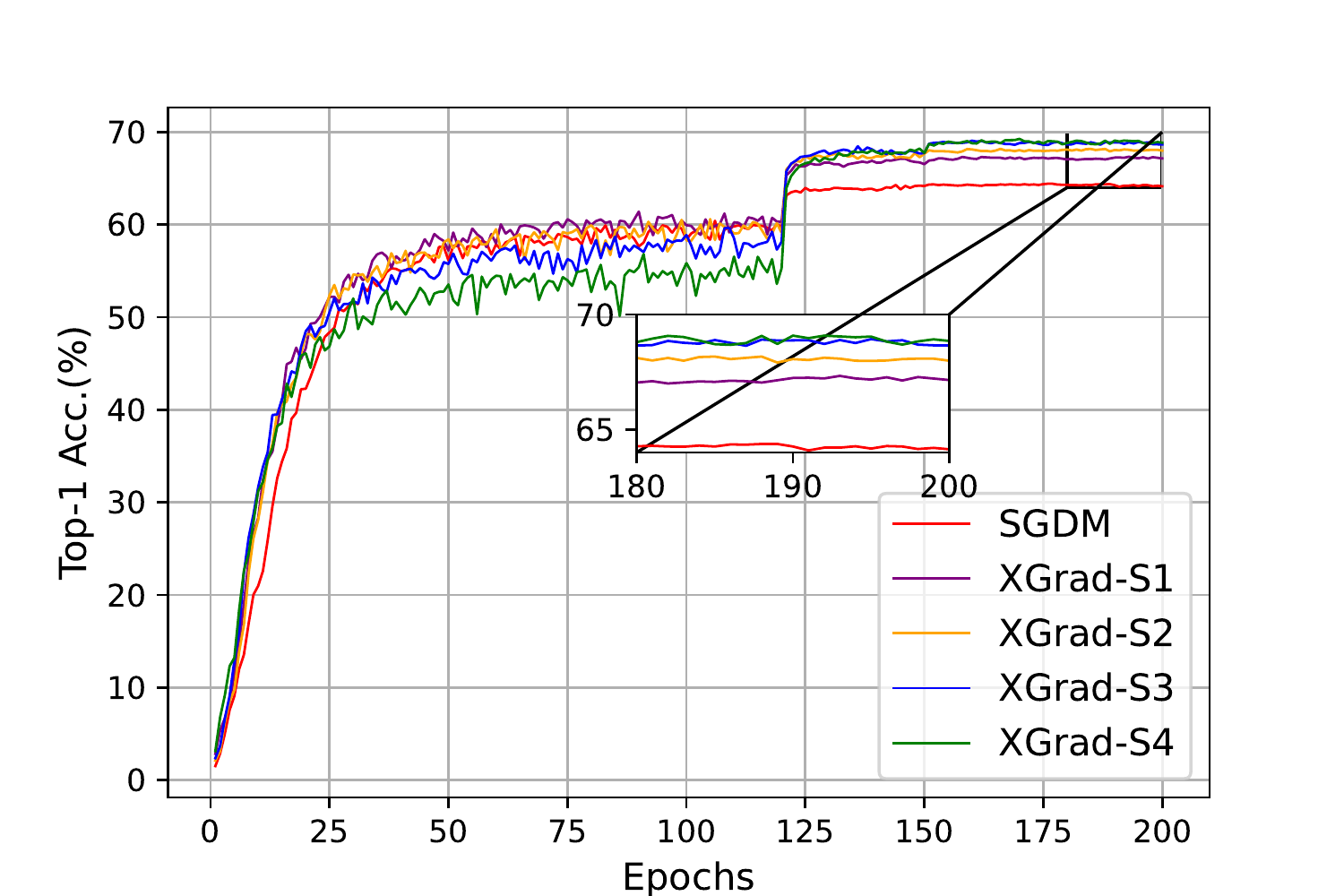}\label{comp-vgg11-cifar100-acc}}
	\subfloat[VGG-16]{\includegraphics[width=.28\textwidth]{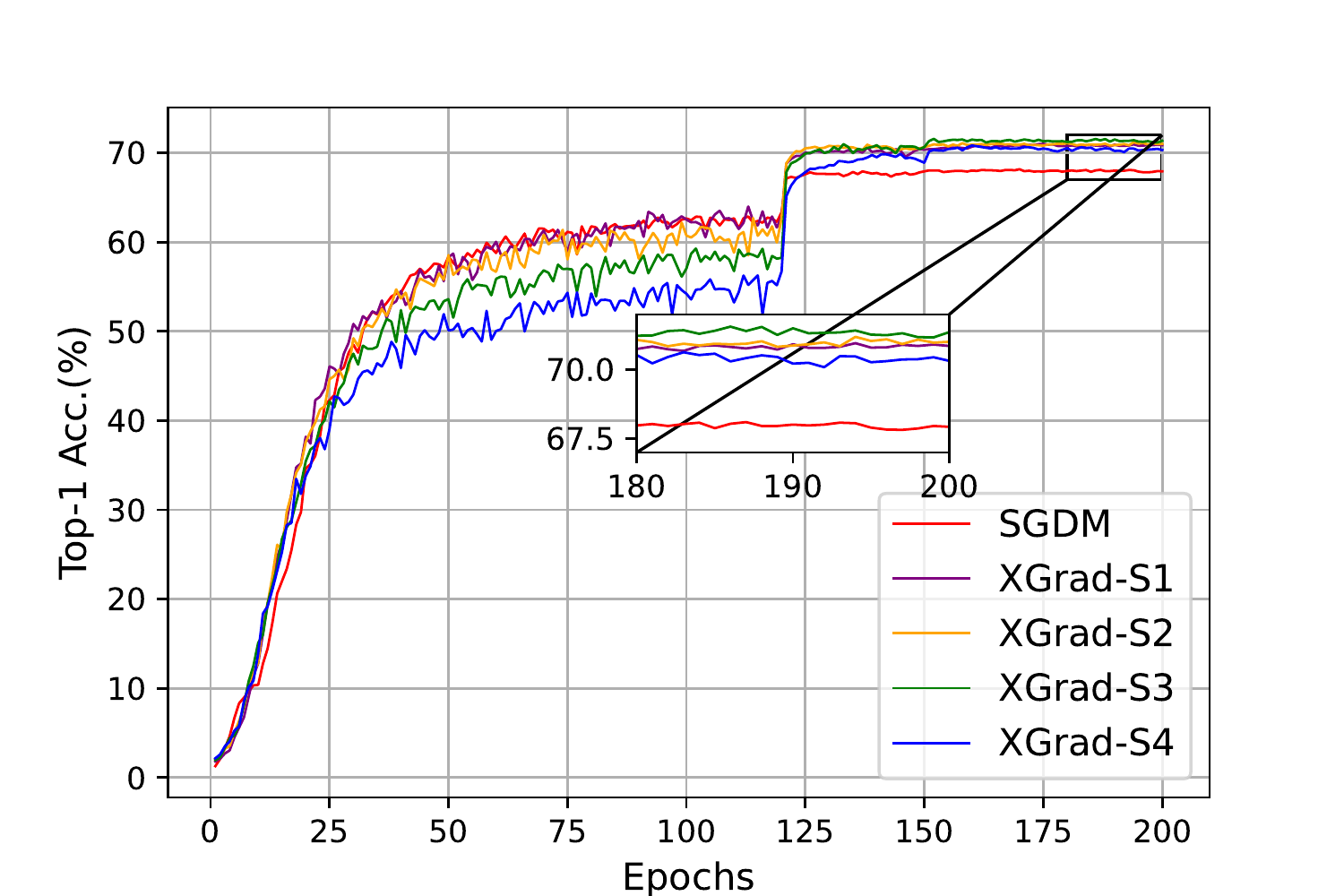}\label{comp-vgg16-cifar100-acc}}
	\quad
	\subfloat[ResNet-34]{\includegraphics[width=.28\textwidth]{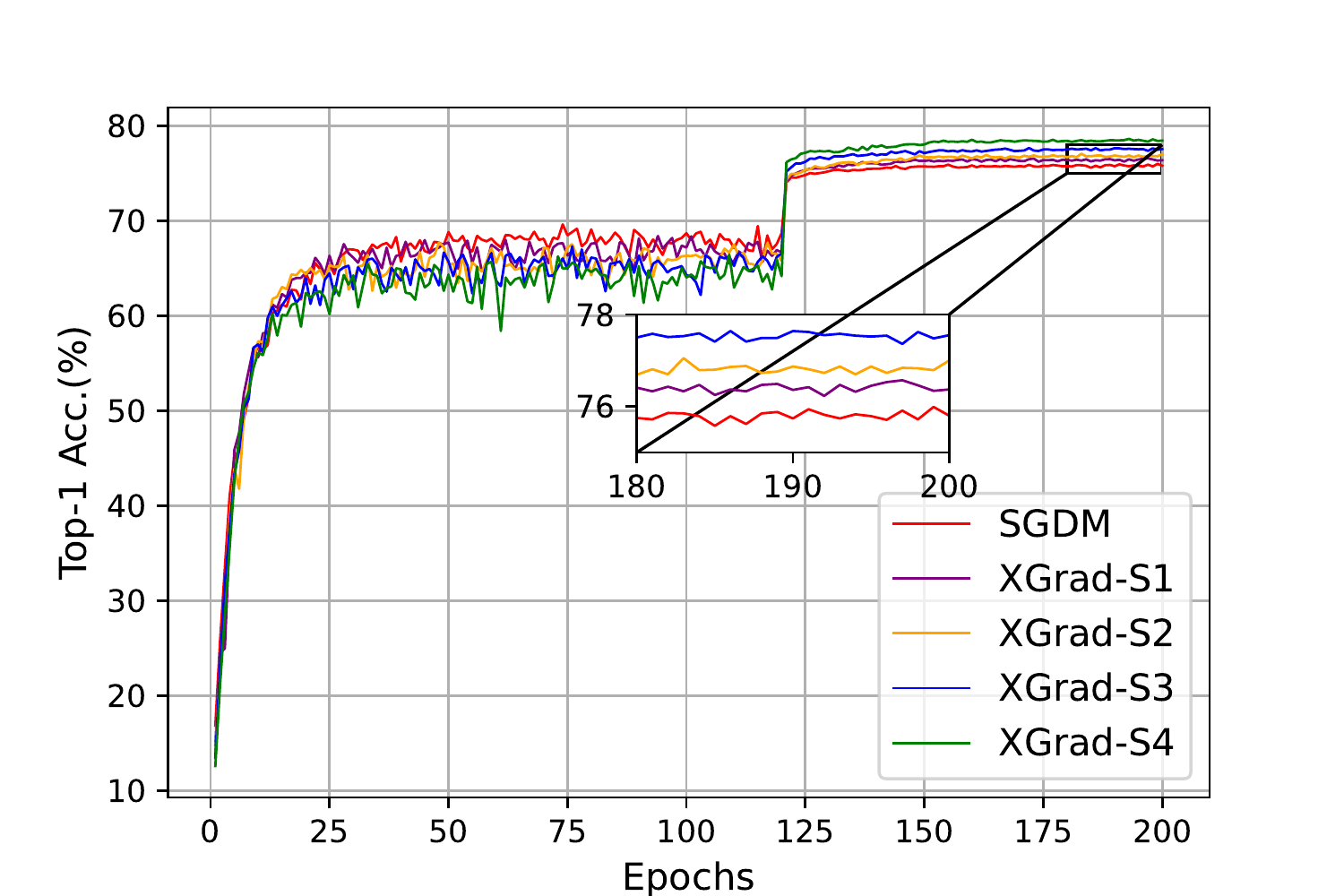}\label{comp-resnet34-cifar100-acc}}
	\subfloat[ResNet-101]{\includegraphics[width=.28\textwidth]{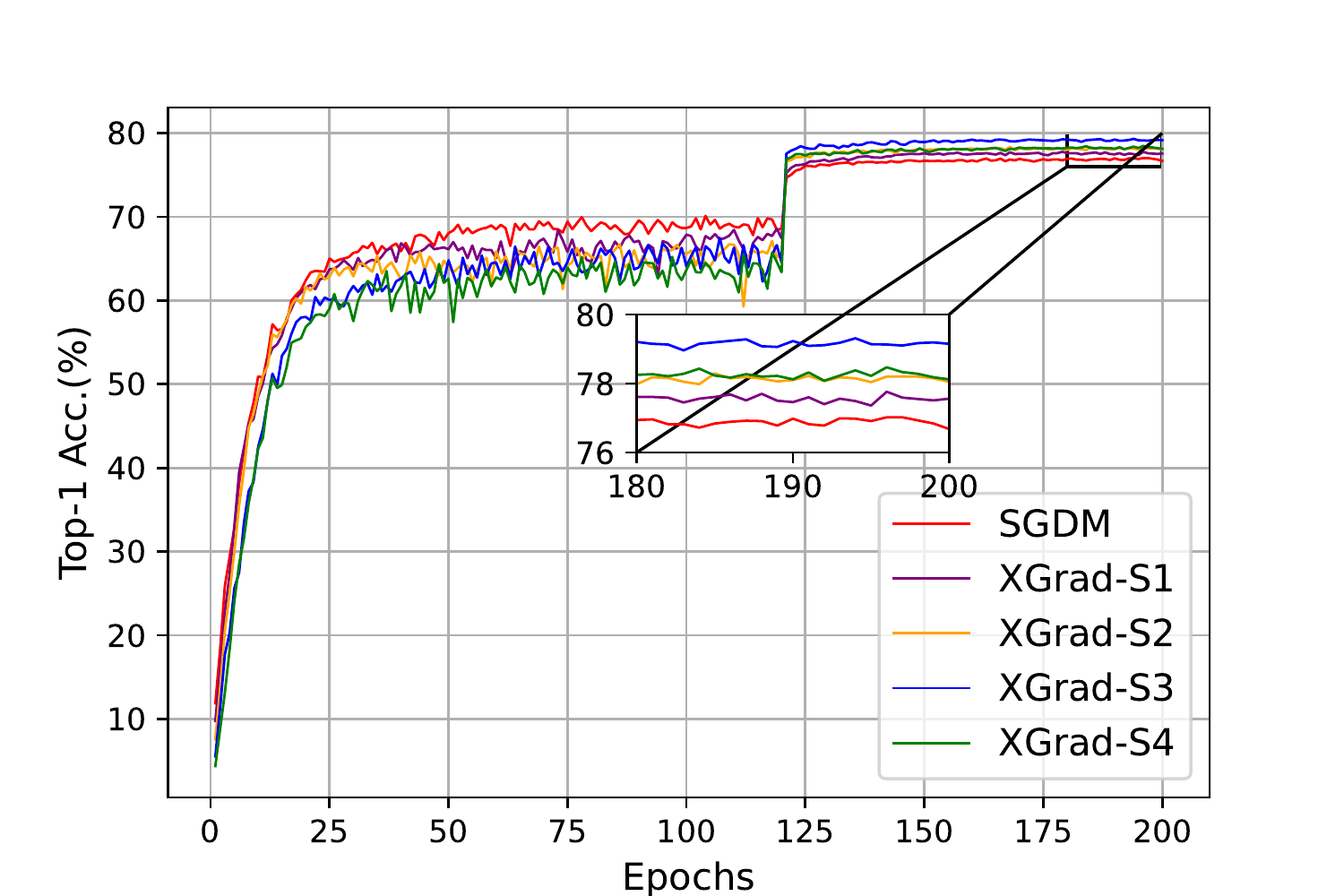}\label{comp-resnet101-cifar100-acc}}
	\subfloat[GoogleNet]{\includegraphics[width=.28\textwidth]{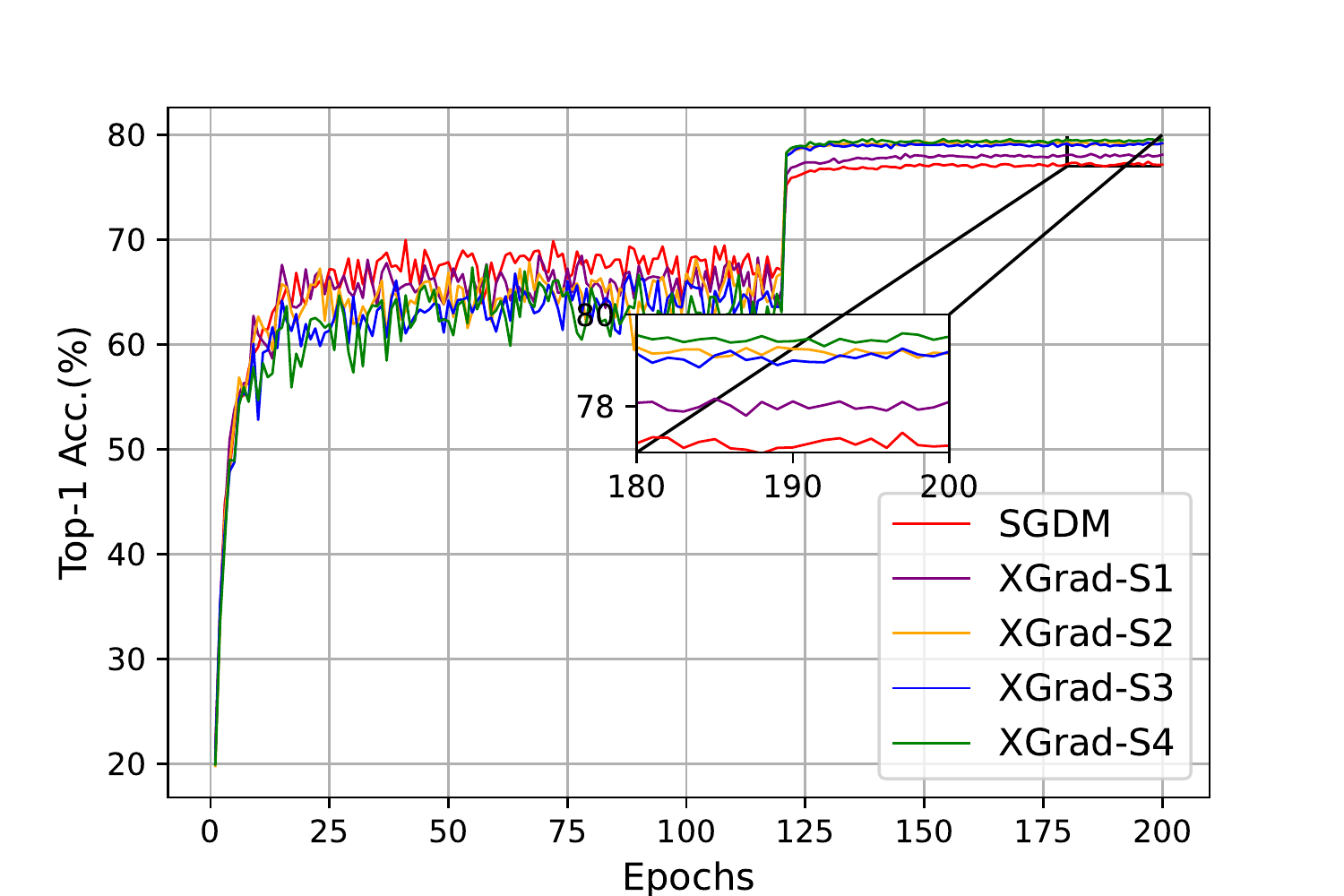}\label{comp-googlenet-cifar100-acc}}
	\quad
	\subfloat[DenseNet-121]{\includegraphics[width=.28\textwidth]{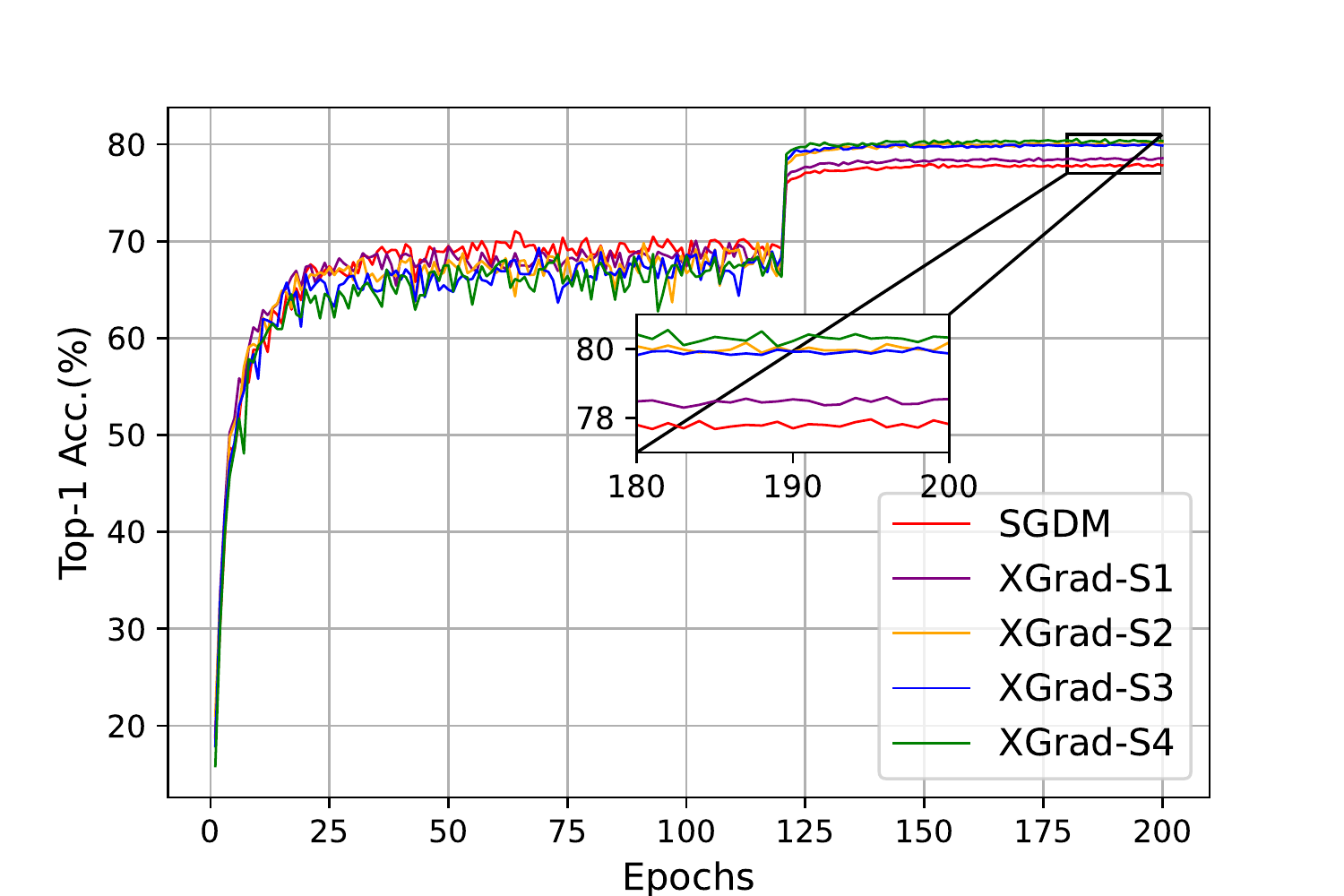}\label{comp-densenet-cifar100-acc}}
	\subfloat[Inception-V3]{\includegraphics[width=.28\textwidth]{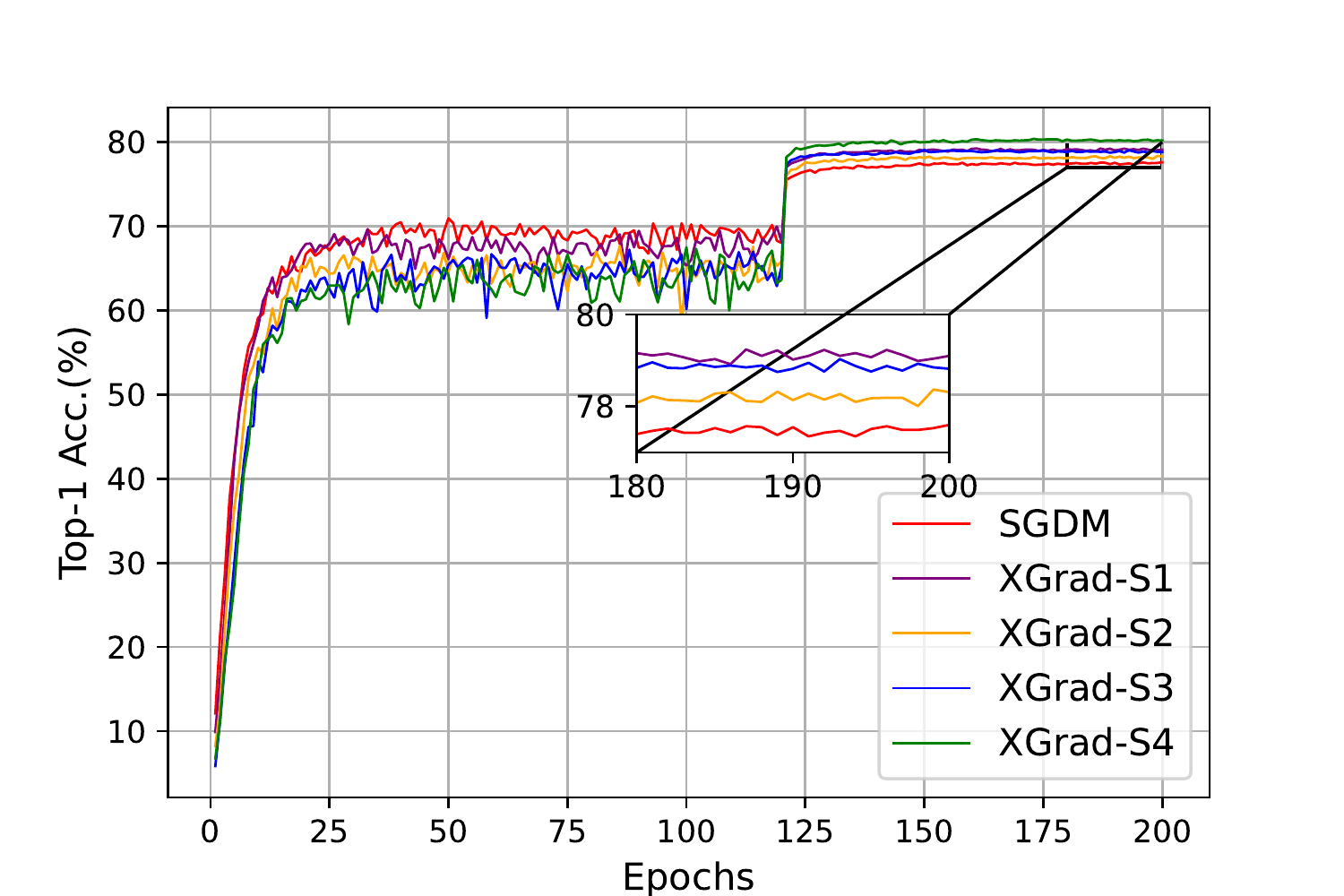}\label{comp-inceptionv3-cifar100-acc}}
	\subfloat[ViT]{\includegraphics[width=.28\textwidth]{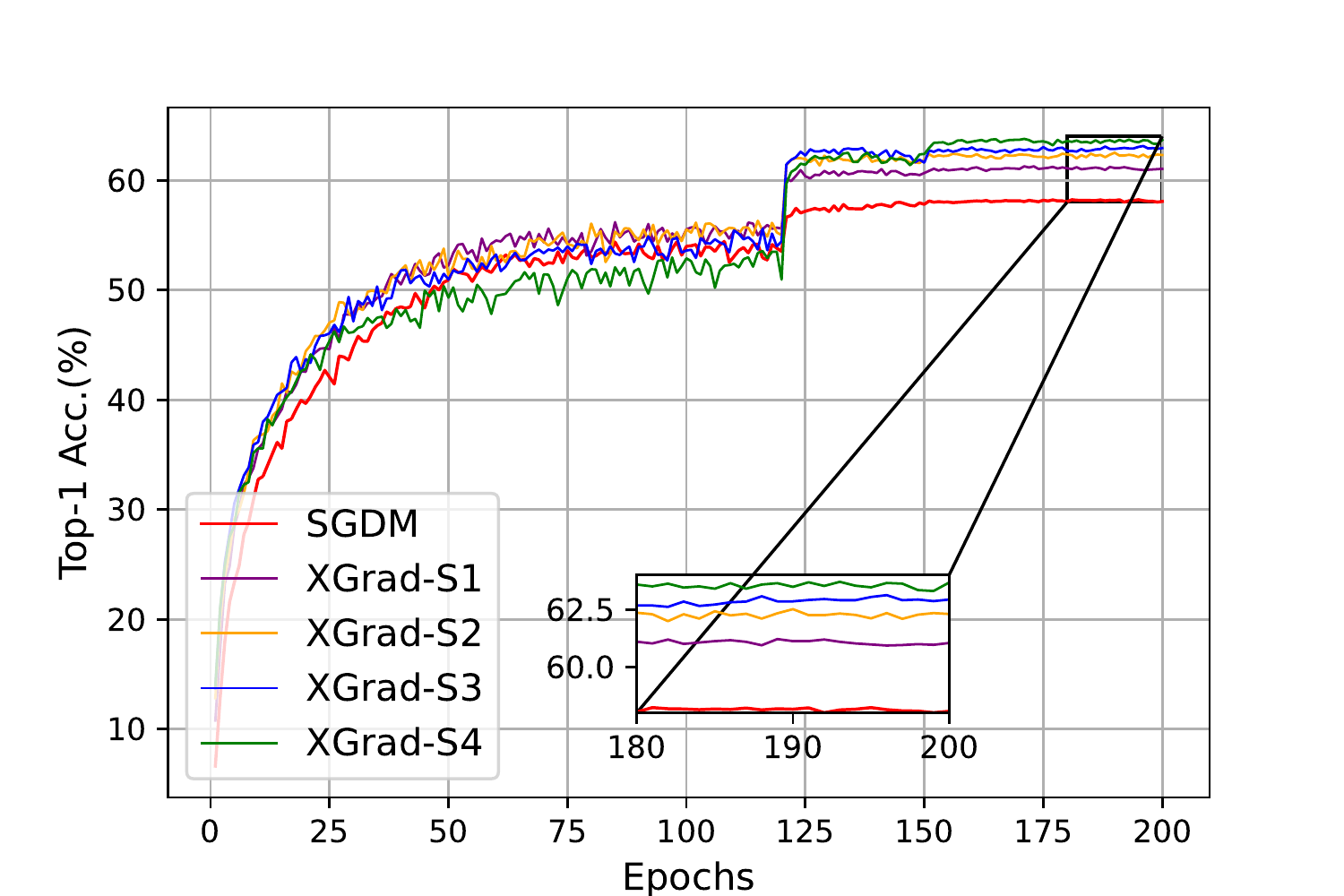}\label{comp-vit-cifar100-acc}}
	\caption{Validation Top-1 accuracy (higher is better) vs. epochs of training CNNs and ViT on CIFAR-100.}
	\label{comp-acc-cifar100}
\end{figure*}

\begin{table*}[h!]
	\centering
	\caption{Comparisons of XGrad and SGDM when training on CIFAR-100. The best results are highlighted in boldface.}
	\label{table:sgd-cifar100}
	\setlength{\tabcolsep}{2.8mm}
	\begin{tabular}{c|ccccccccc}
		\toprule
		Optimizers & AlexNet & VGG-11 &  VGG-16  & ResNet-34   & ResNet-101 & GoogleNet  & DenseNet-121&  Inception-V3 & ViT  \\
		\midrule
		\multicolumn{10}{c}{Maximum Top-1 Accuracy} \\
		\midrule
		SGDM & 69.48\% & 64.42\% & 68.17\% &75.99\% & 77.02\% &  77.43\% &  77.96\% &  77.60\% & 58.23\% \\
		XGrad-S1 & \textbf{70.34}\%  & 67.33\% & 71.01\% &  76.57\% &  77.76\%& 78.17\%  & 78.60\% & 79.25\% & 61.27\% \\
		XGrad-S2&70.31\% & 68.18\% &  71.19\% & 77.05\% & 78.34\% & 79.40\% & 80.19\%  & 78.37\% & 62.51\% \\
		XGrad-S3 & 70.21\% & 69.06\% & \textbf{71.57}\% &  77.66\% &  \textbf{79.31}\% & 79.26\% & 80.05\% & 79.09\% & 63.12\% \\
		XGrad-S4 & 69.48\% & \textbf{69.26}\% & 70.83\% &  \textbf{78.62}\% &  78.47\% & \textbf{79.61}\% & \textbf{80.55}\% & \textbf{80.39}\% & \textbf{63.77}\%\\
		\bottomrule
	\end{tabular}
\end{table*}

Figure~\ref{comp-acc-cifar10} presents the learning curves for training CNN and ViT models on the CIFAR-10 dataset. Table~\ref{table:sgd-cifar10} summarizes the obtained maximum validation top-1 accuracy. Figure~\ref{comp-acc-cifar100} shows the learning curves for training CNN models on the CIFAR-100 dataset. Table~\ref{table:sgd-cifar100} summarizes the obtained maximum  top-1 validation accuracy on CIFAR-100.  In all figures of Figures~\ref{comp-acc-cifar10} and~\ref{comp-acc-cifar100}, we let the red lines illustrate the learning curves of SGDM and let other colored lines represent the learning curves of XGrad with different weight prediction steps. 

Based on the observation of Table~\ref{table:sgd-cifar10} and Figure~\ref{comp-acc-cifar10}, we can immediately reach the following conclusions. First, the learning curves of XGrad with different weight prediction steps tend to be lower than that of SGDM at the beginning training period. Yet, as the decaying of the learning rate, XGrad gradually surpasses SGDM and always gets higher top-1 accuracy than SGDM at the end of the training period.  This demonstrates that XGrad tends to be inferior to SGDM with a large learning rate but outperforms SGDM with a small learning rate.  Second, Table~\ref{table:sgd-cifar10} illustrates that XGrad outperforms SGDM on all evaluated DNN models in terms of the obtained top-1 validation accuracy. In particular, XGrad achieves consistently higher top-1 validation accuracy than SGDM. 
Compared to SGDM, XGrad respectively achieves an improvement of 0.42\%, 1.81\%, 1.43\%, 1.00\%, 0.57\%, 0.53\%, 0.52\%, 0.92\%, and 1.60\% top-1 accuracy for training AlexNet, VGG-11, VGG-16, ResNet-34, ResNet-101, GoogleNet, DenseNet-121, Inception-V3, and ViT. On average, our proposal yields 0.98\% (up to 1.81\%) top-1 accuracy improvement over SGDM.  Third, 
comparing the experimental results of XGrad-S1, XGrad-S2, XGrad-S3, and XGrad-S4, we can see that XGrad with different weight prediction steps consistently gets good results for all CNN models. The special case is training ViT on CIFAR-10, where XGrad-S4 performs poorly and obtains the minimum top-1 accuracy.


We can draw similar conclusions from the observation of the experiment results shown in Table~\ref{table:sgd-cifar100} and Figure~\ref{comp-acc-cifar100}. First, the learning curves of SGDM seem to be higher than that of XGrad at the beginning of the training but lower than that of XGrad at the end of the training. 
Second, Table~\ref{table:sgd-cifar100} illustrates that XGrad achieves consistently higher top-1 validation accuracy than SGDM. Specially, XGrad yields 0.86\%, 4.84\%, 3.40\%, 2.63\%, 2.29\%, 2.18\%, 2.59\%, 2.79\%, and 5.54\% more top-1 accuracy than SGDM when training AlexNet, VGG-11, VGG-16, ResNet-34, ResNet-101, GoogleNet, DenseNet-121 and Inception-V3, respectively. That is to say, XGrad can achieve an average of 3.01\% (up to 5.54\%) top-1 accuracy improvement over SGDM. Third, when comparing the experimental results of XGrad-S1, XGrad-S2, XGrad-S3, and XGrad-S4, we can see that XGrad always achieves comparable high performance.


\begin{figure*}[h!]
	\centering
	\subfloat[LeNet]{\includegraphics[width=.28\textwidth]{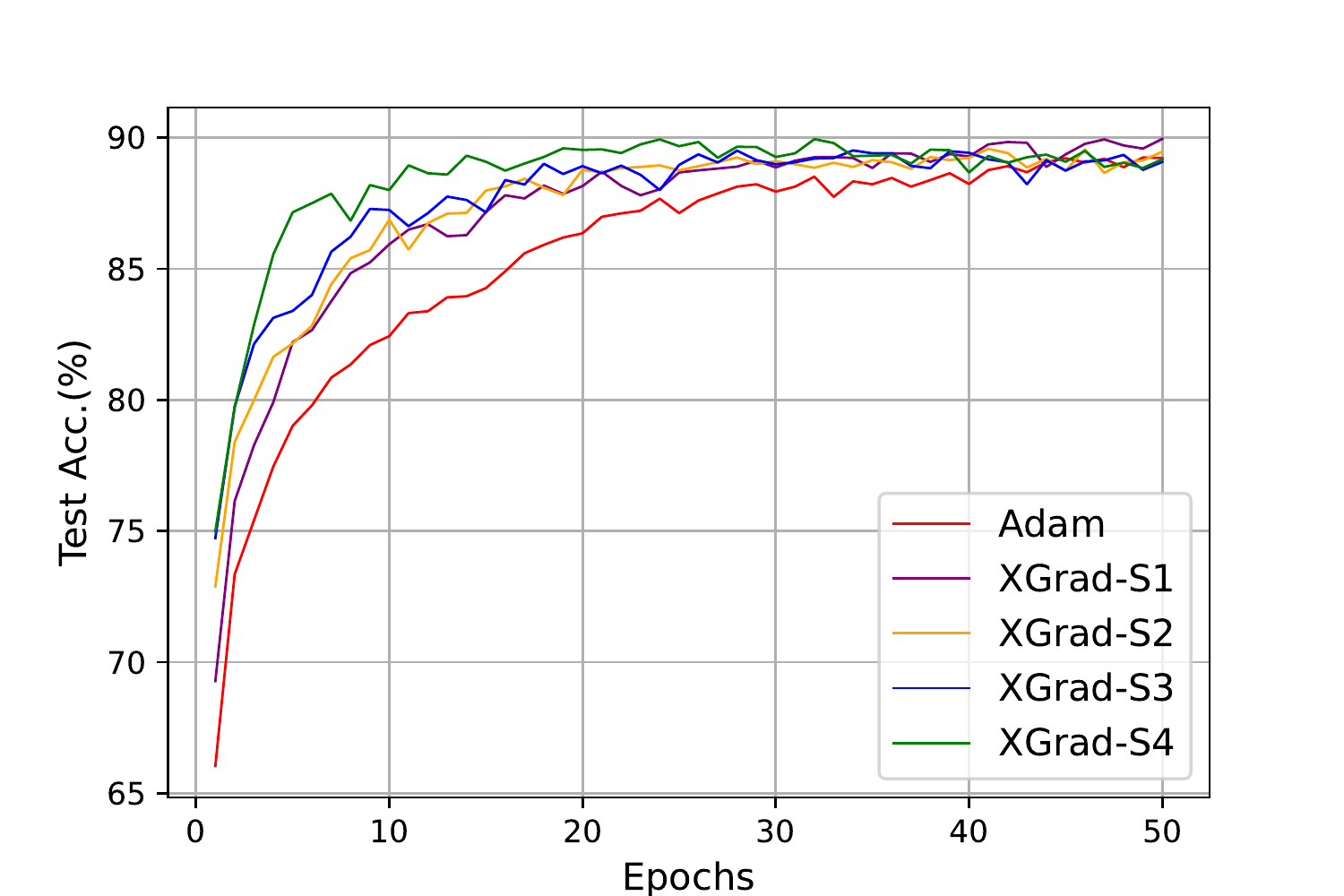}\label{comp-adam-lenet-acc}}
	\subfloat[ResNet-34]{\includegraphics[width=.28\textwidth]{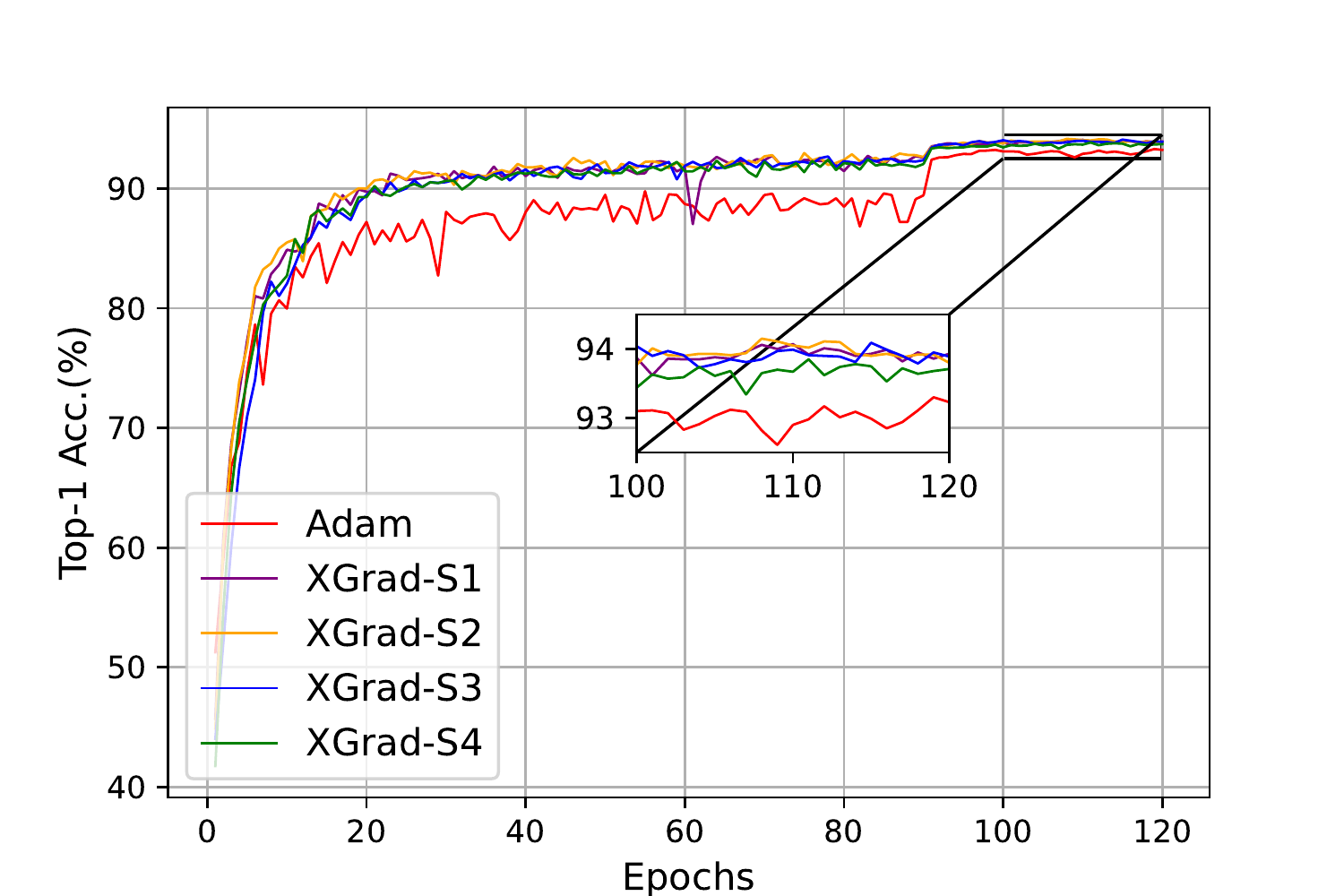}\label{comp-adam-resnet34-acc}}
	\subfloat[DenseNet-121]{\includegraphics[width=.28\textwidth]{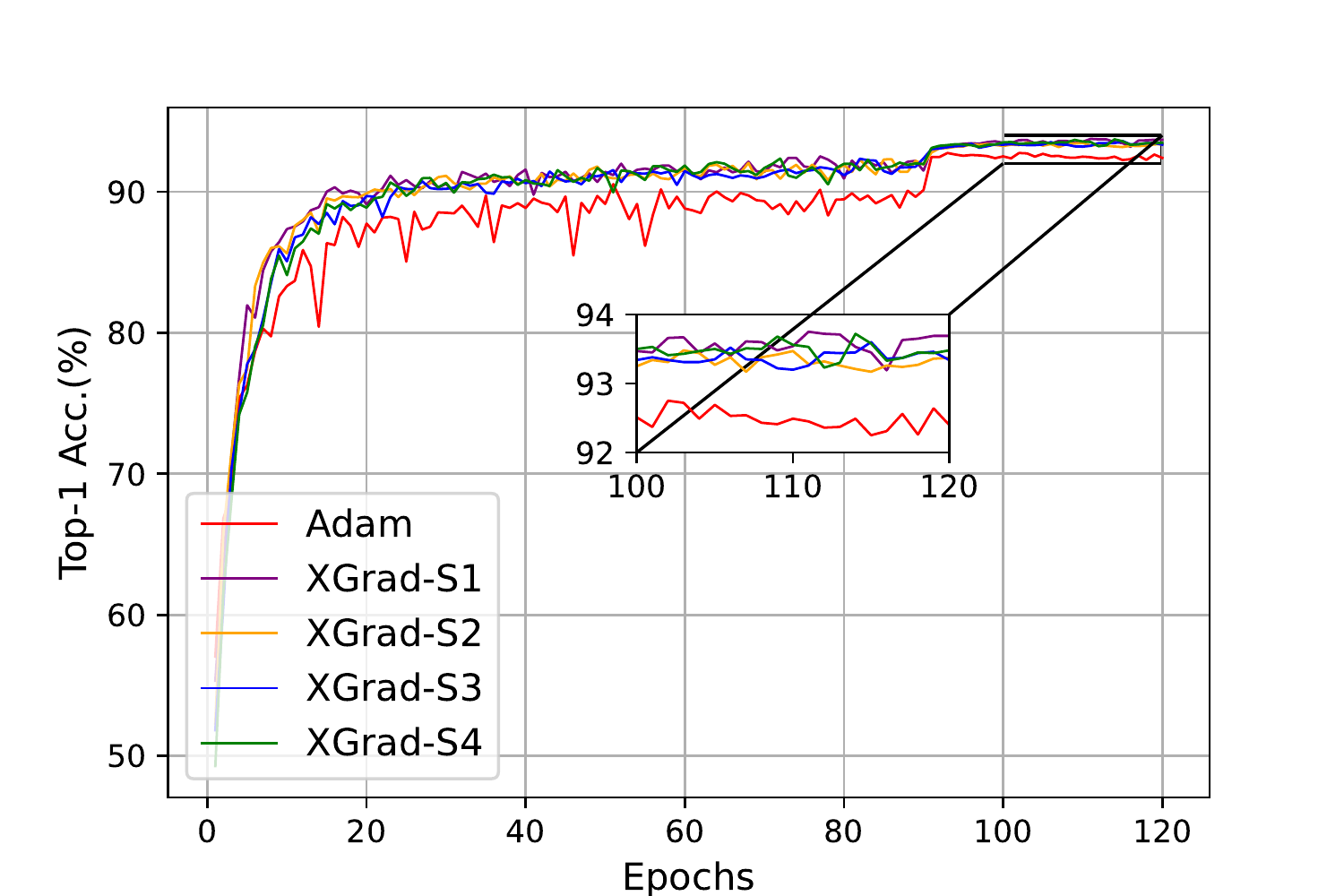}\label{comp-adam-densenet-acc}}
	\quad
	\subfloat[Inception-V3]{\includegraphics[width=.28\textwidth]{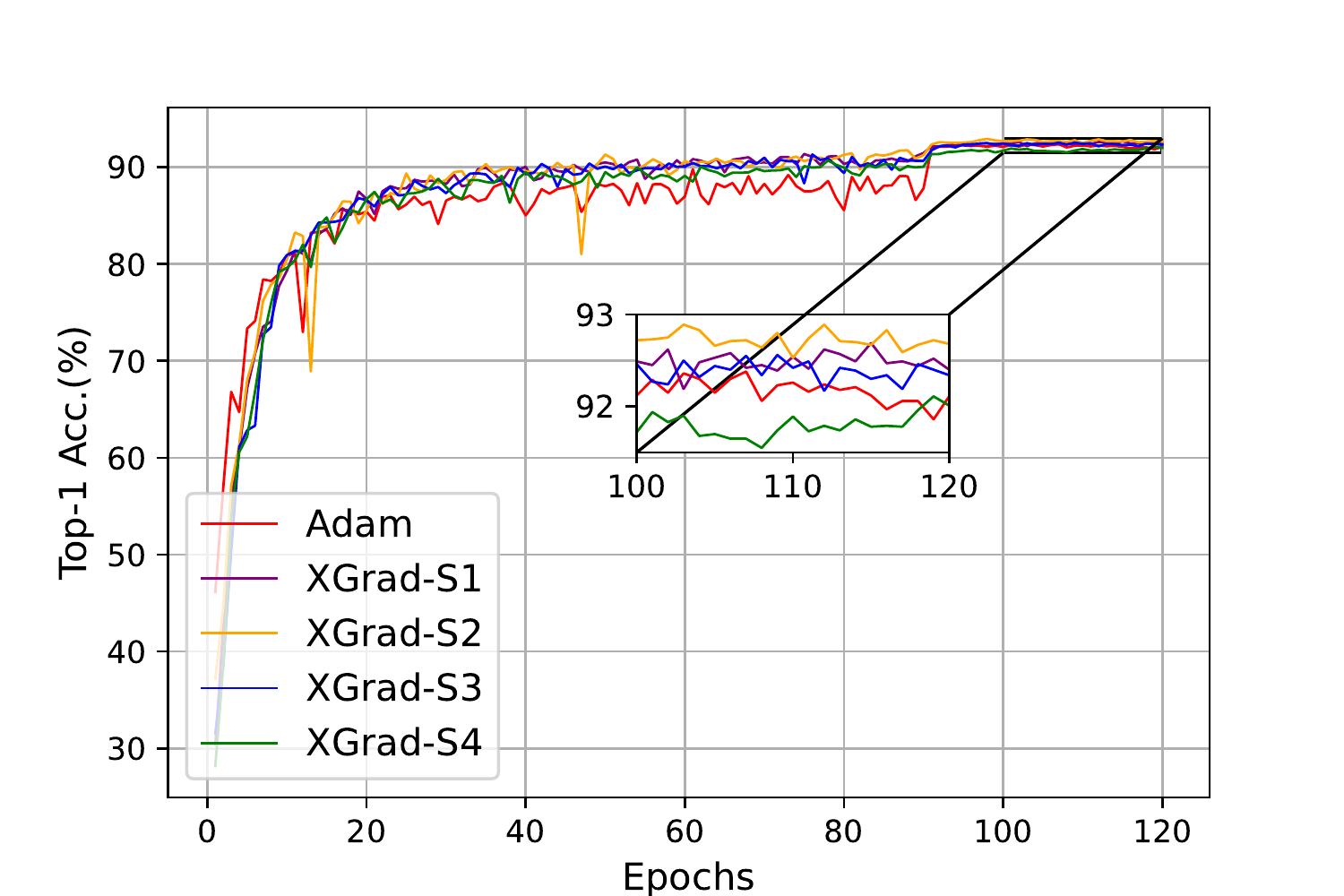}\label{comp-adam-inceptionv3-acc}}
	\subfloat[LSTM-1]{\includegraphics[width=.28\textwidth]{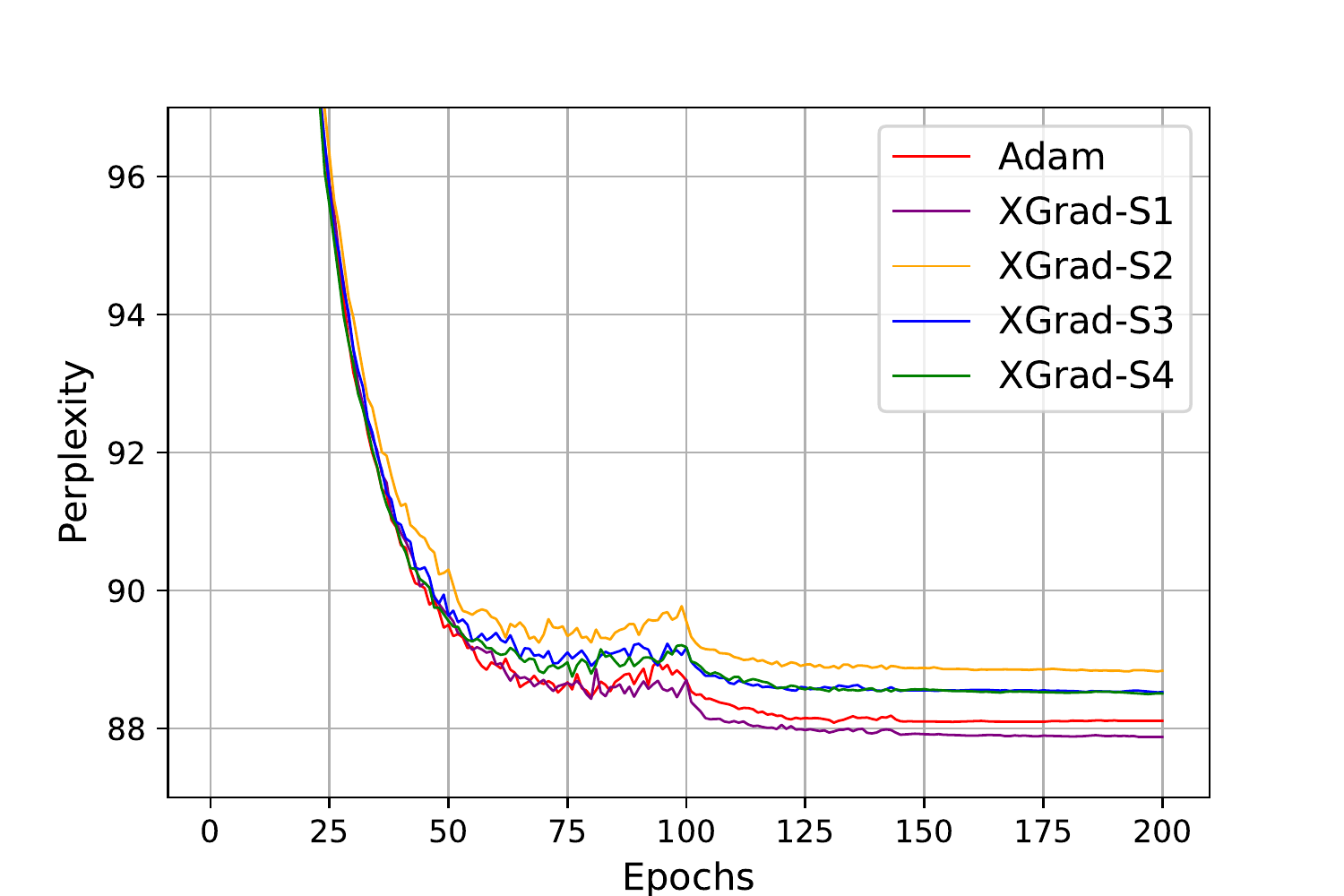}\label{comp-adam-lstm1}}
	\subfloat[LSTM-2]{\includegraphics[width=.28\textwidth]{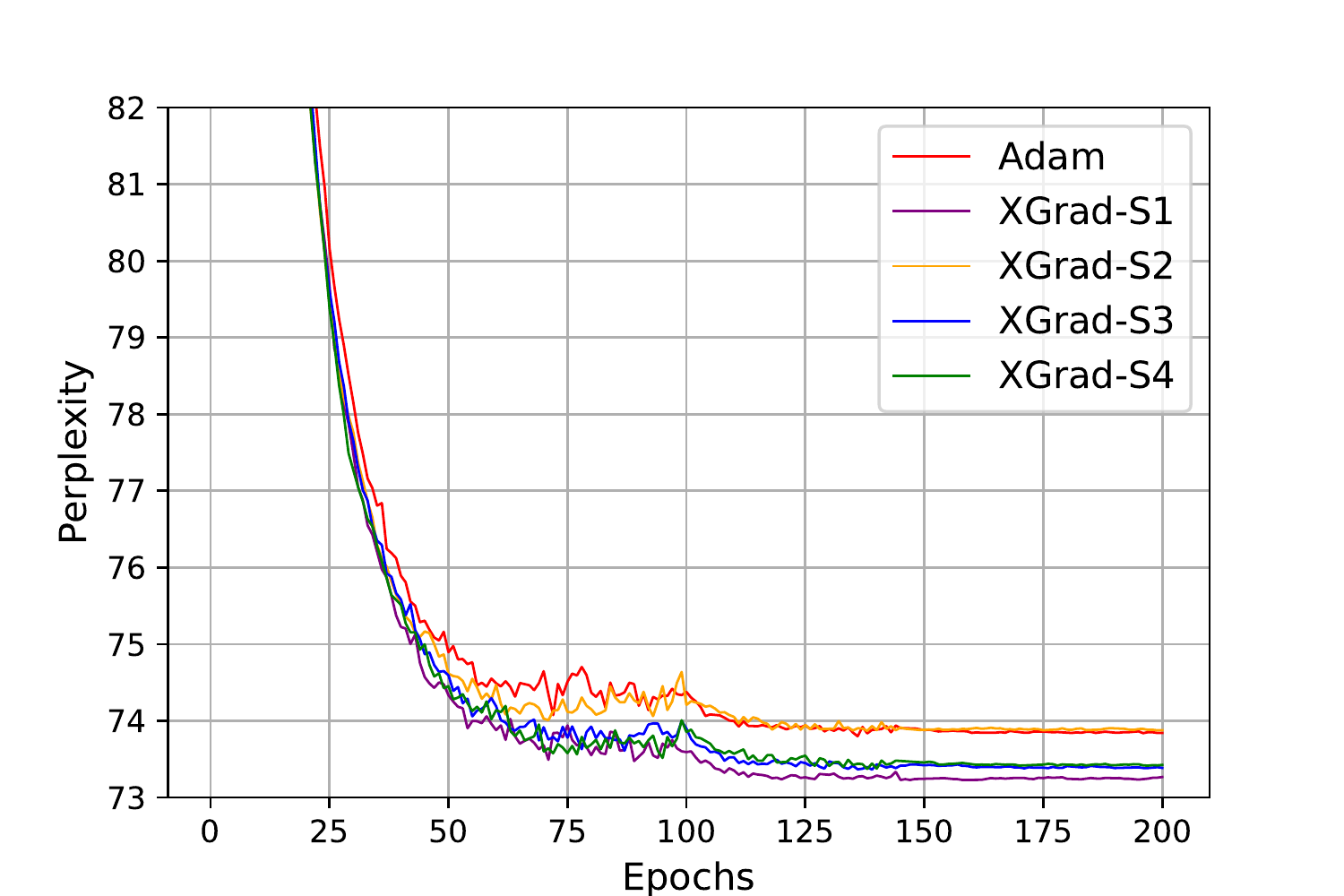}\label{comp-adam-lstm2}}
	\quad
	\subfloat[LSTM-3]{\includegraphics[width=.28\textwidth]{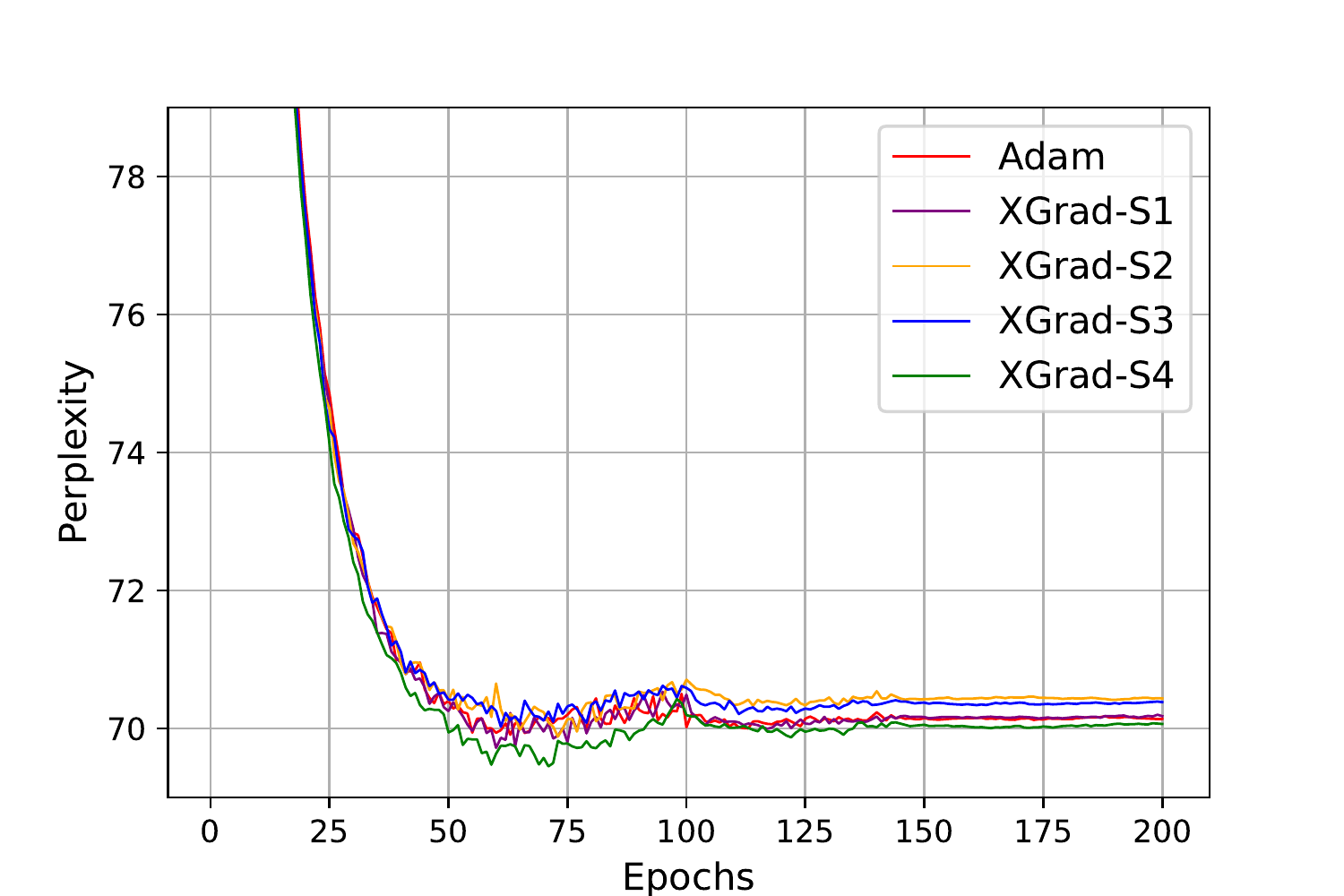}\label{comp-adam-lstm3}}
	\subfloat[GNMT-8]{\includegraphics[width=.28\textwidth]{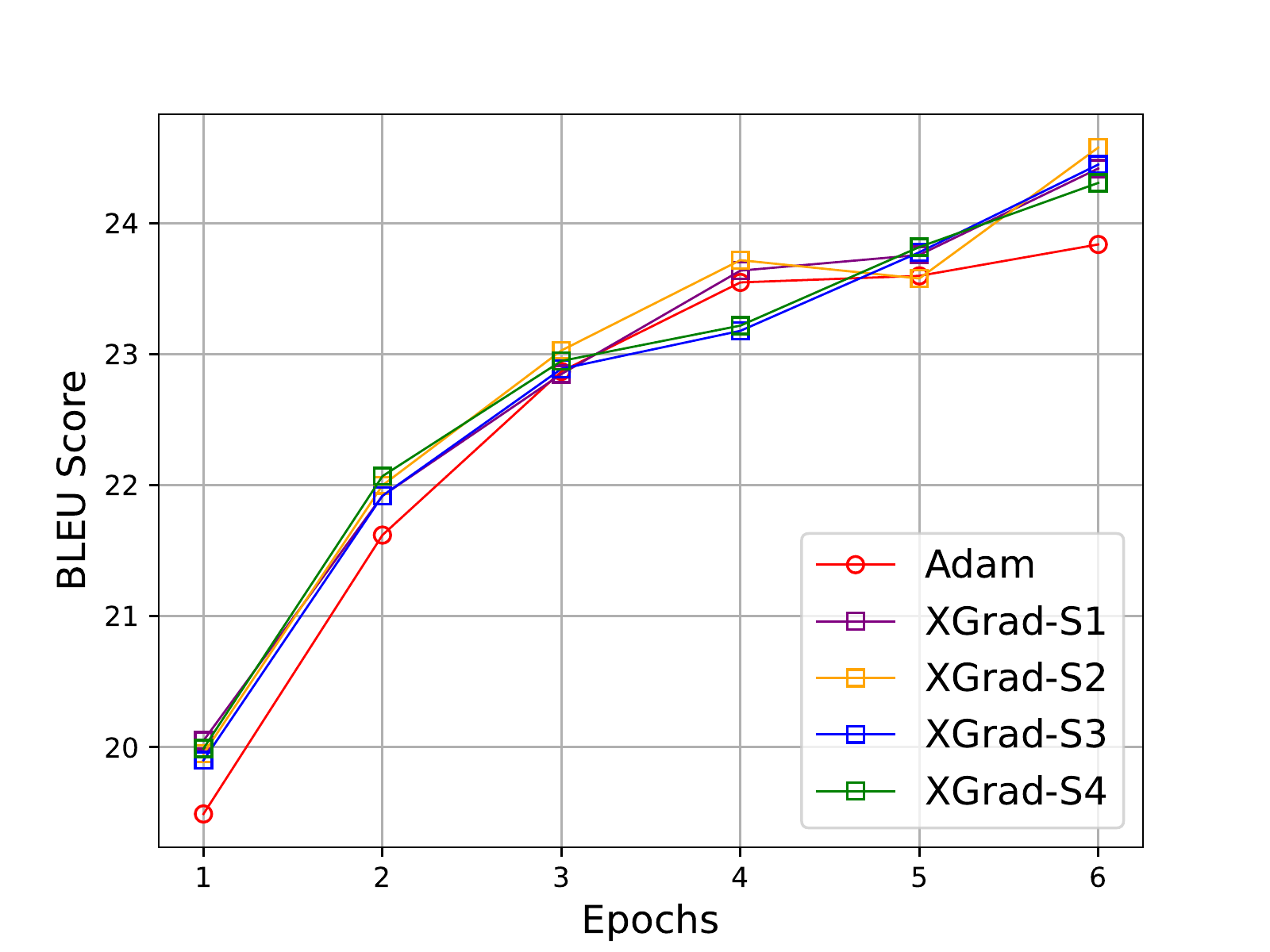}\label{comp-adam-gnmt8}}
	\subfloat[VAE]{\includegraphics[width=.28\textwidth]{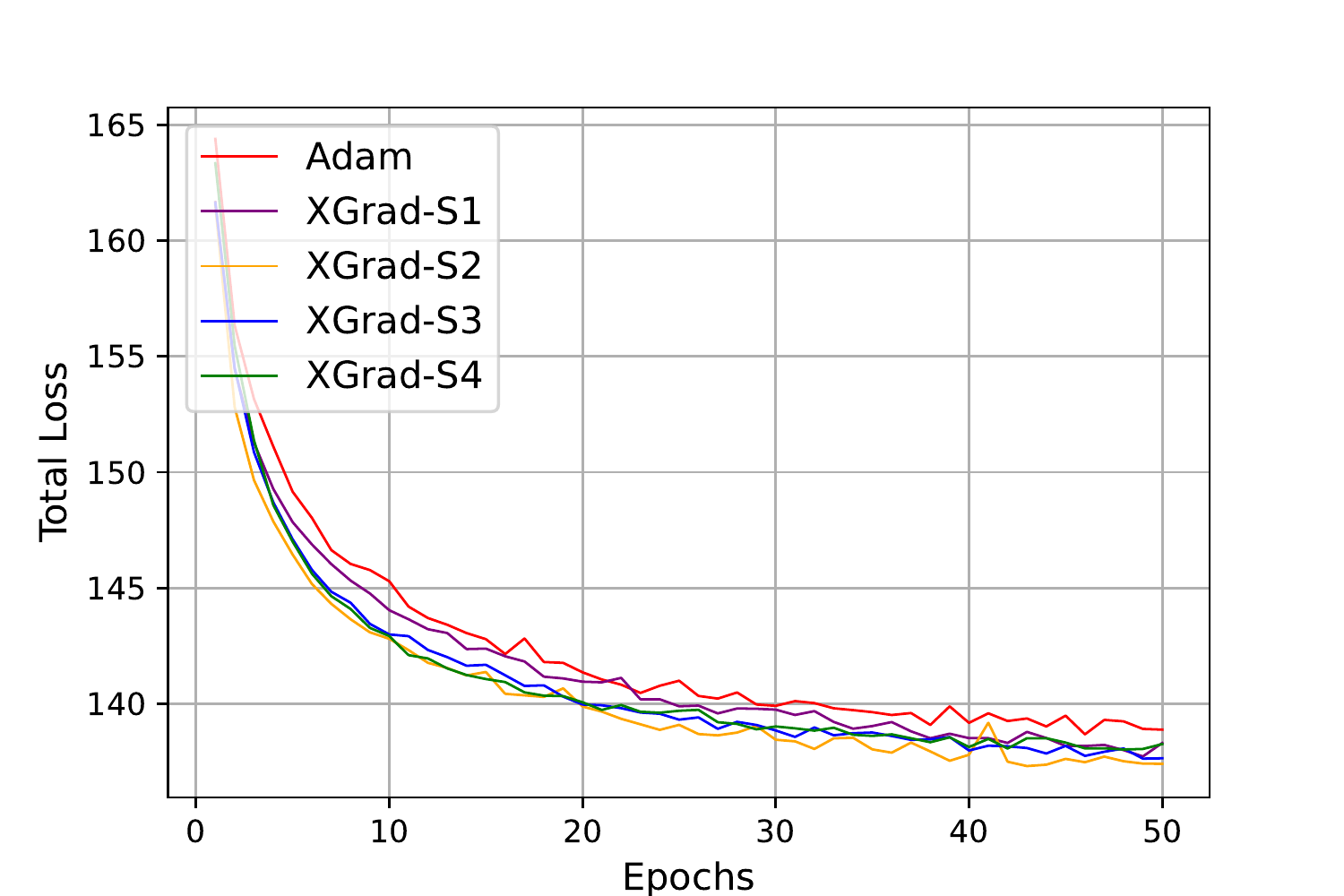}\label{comp-adam-vae}}
	\caption{Comparison of Adam and XGrad. Figures~\ref{comp-adam-lenet-acc}, ~\ref{comp-adam-resnet34-acc}, ~\ref{comp-adam-densenet-acc}, and~\ref{comp-adam-inceptionv3-acc}: Top-1 accuracy (higher is better) vs. Epochs; Figures~\ref{comp-adam-lstm1}, \ref{comp-adam-lstm2}, and~\ref{comp-adam-lstm3}: Perplexity (lower is better) vs. Epochs; Figure~\ref{comp-adam-gnmt8}: BLEU score (higher is better) vs. Epochs; Figure~\ref{comp-adam-vae}: Total loss (lower is better) vs. Epochs.}
	\label{comp-adam-acc-cifar10}
\end{figure*}

\begin{table*}[h!]
	\centering
	\caption{Summarization of best model accuracy of XGrad and Adam. Maximum top-1 accuracy for LeNet, ResNet-34, DenseNet-121,  and Incepiton-V3; Minimum perplexity for LSTM-1/2/3; Maximum BLEU score for GNMT-8; Maximum Dev set accuracy for BERT$_{\text{BASE}}$; Minimum total loss for VAE; Minimum FID score for WGAN. The best model accuracy results are highlighted in boldface.}
	\label{table:adam}
	\setlength{\tabcolsep}{1.8mm}
	\begin{tabular}{c|ccccccccccc}
		\toprule
		Optimizers & LeNet & ResNet-34 & DenseNet-121 & Inception-V3 & LSTM-1 & LSTM-2  & LSTM-3&  GNMT-8 & BERT$_{\text{BASE}}$ & VAE & WGAN \\
		\midrule
		\multicolumn{12}{c}{Best Accuracy} \\
		\midrule
		\makecell{Adam} & 89.24\%  & 93.30\% & 92.75\% & 92.47\%  &88.08& 73.79& 69.91 &  23.84 & 86.76\% & 138.67 & 95.60 \\
		XGrad-S1 & \textbf{89.95}\%& 94.07\%  & \textbf{93.75}\% & 92.69\%   &  \textbf{87.88} & \textbf{73.22} & 69.71  & 24.42 &  \textbf{87.25}\% & 137.72 & 75.12   \\
		XGrad-S2 & 89.57\% & \textbf{94.15}\% & 93.48\% &  \textbf{92.94}\%   & 88.83 & 73.87& 69.88 &  \textbf{24.58}   & 86.52\% & \textbf{137.30} & \textbf{72.90}   \\
		XGrad-S3 & 89.51\% & 94.09\% & 93.60\% & 92.56\% &88.52 & 73.36&  70.02& 24.45 & 87.01\% & 137.62 & 87.17      \\
		XGrad-S4 & 89.94\% & 93.85\% & 93.72\% &  92.11\% & 88.50& 73.37 & \textbf{69.45}& 24.31  &86.52\%  & 138.02 &  84.61 \\
		\bottomrule
	\end{tabular}
\end{table*}

\subsubsection{Comparisons of XGrad with Adam}\label{sec:comp-xgrad-adam}
In this section, we selected LeNet, ResNet-34, DenseNet-121, Inception-V3, LSTM-1, LSTM-2, LSTM-3, GNMT-8, BERT$_{\text{BASE}}$, VAE, and WGAN as the benchmark models. Concretely, we trained LeNet on Fashion-MNIST for 50 epochs with a fixed learning rate of $1e^{-3}$ and a mini-batch size of 128. We trained ResNet-34, DenseNet-121, and Inception-V3 on CIFAR-10 for 120 epochs with a mini-batch size of 128.  The initial learning rate was $1e^{-3}$ and was divided by 10 at the 90th epoch. We trained LSTM-1, LSTM-2, and LSTM-3 on the Penn Treebank Dataset for 200 epochs with a mini-batch size of 20. The initial learning rate was $1e^{-3}$ and was decayed by 0.1 at the 100th and 145th epochs. We trained GNMT-8 on WMT-16 En$\rightarrow$De for 6 epochs with a steady learning rate of $3e^{-4}$ and a mini-batch size of 128. We trained BERT$_{\text{BASE}}$ on MRPC for 3 epochs with a learning rate of $2e^{-5}$. We trained VAE on MNIST for 50 epochs with a learning rate of $1e^{-3}$ and a mini-batch size of 128. We trained WGAN for 100 epochs with a learning rate of $2e^{-4}$. When training WGAN with XGrad, we empirically only applied weight prediction to the generative model and not to the discriminative model. Furthermore, for both XGrad and Adam, unless specially mentioned, we always evaluated them with the default settings, \ie, $\gamma=1e^{-3}$ and $(\beta_1, \beta_2)$=(0.9, 0.999).  For training LSTM models, we set $\lambda=1.2e^{-4}$ and $\epsilon=1e^{-12}$.  For training WGAN, we set $(\beta_1, \beta_2)=(0.5, 0.999)$, $\epsilon=1e^{-12}$.

\begin{figure*}[h!]
	\centering
	\subfloat[LeNet]{\includegraphics[width=.28\textwidth]{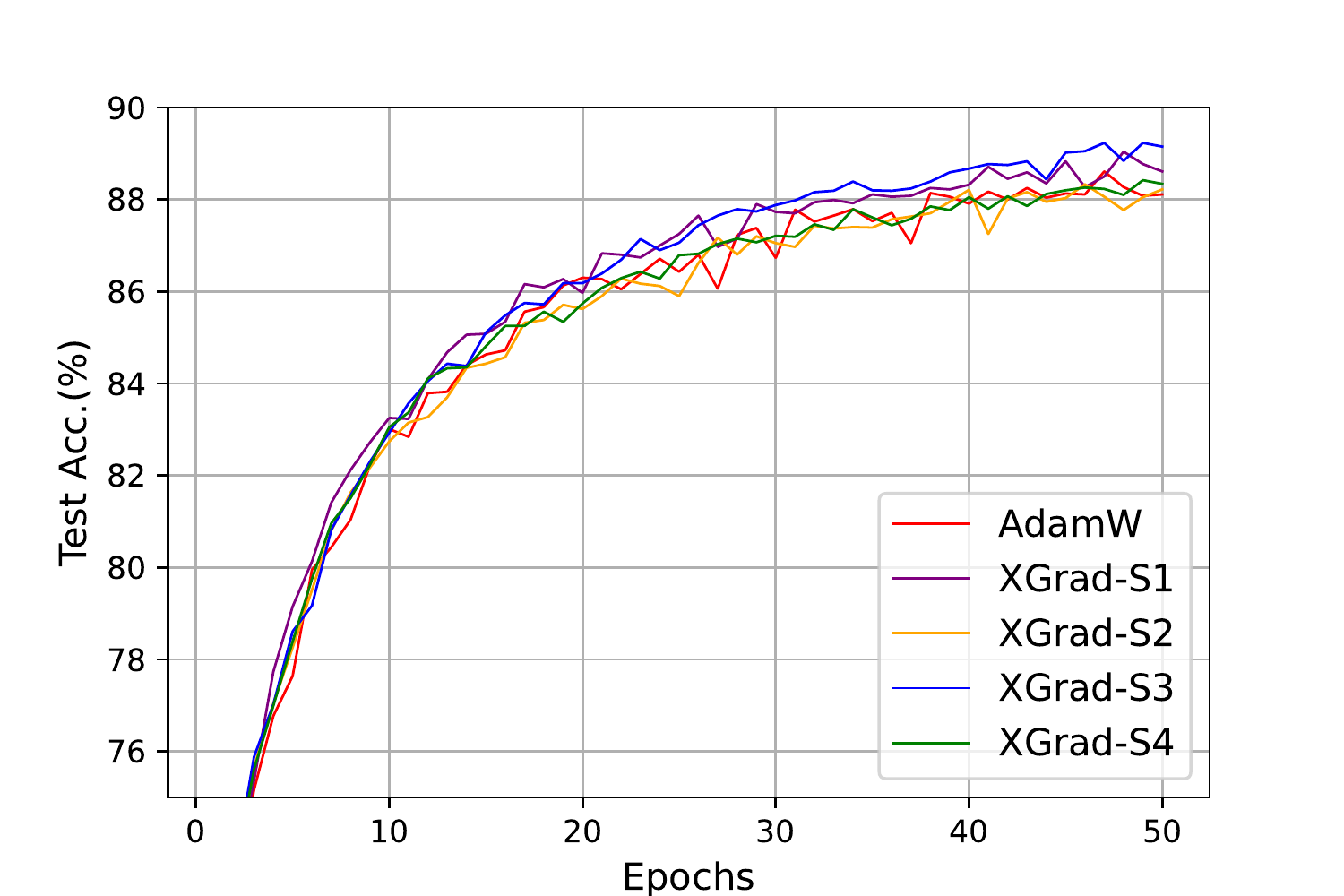}\label{comp-adamw-lenet-acc}}
	\subfloat[ResNet-34]{\includegraphics[width=.28\textwidth]{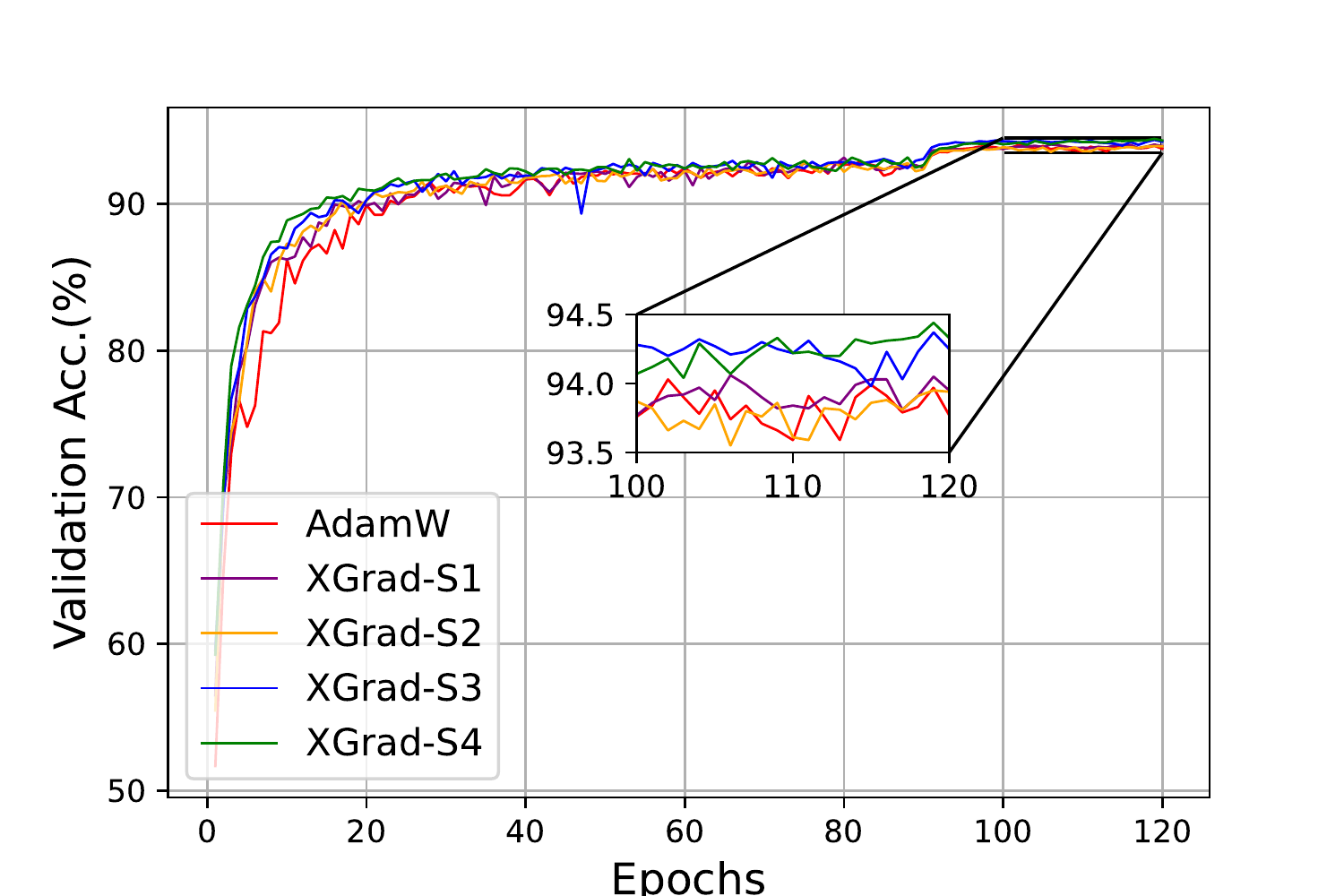}\label{comp-adamw-resnet34-acc}}
	\subfloat[DenseNet-121]{\includegraphics[width=.28\textwidth]{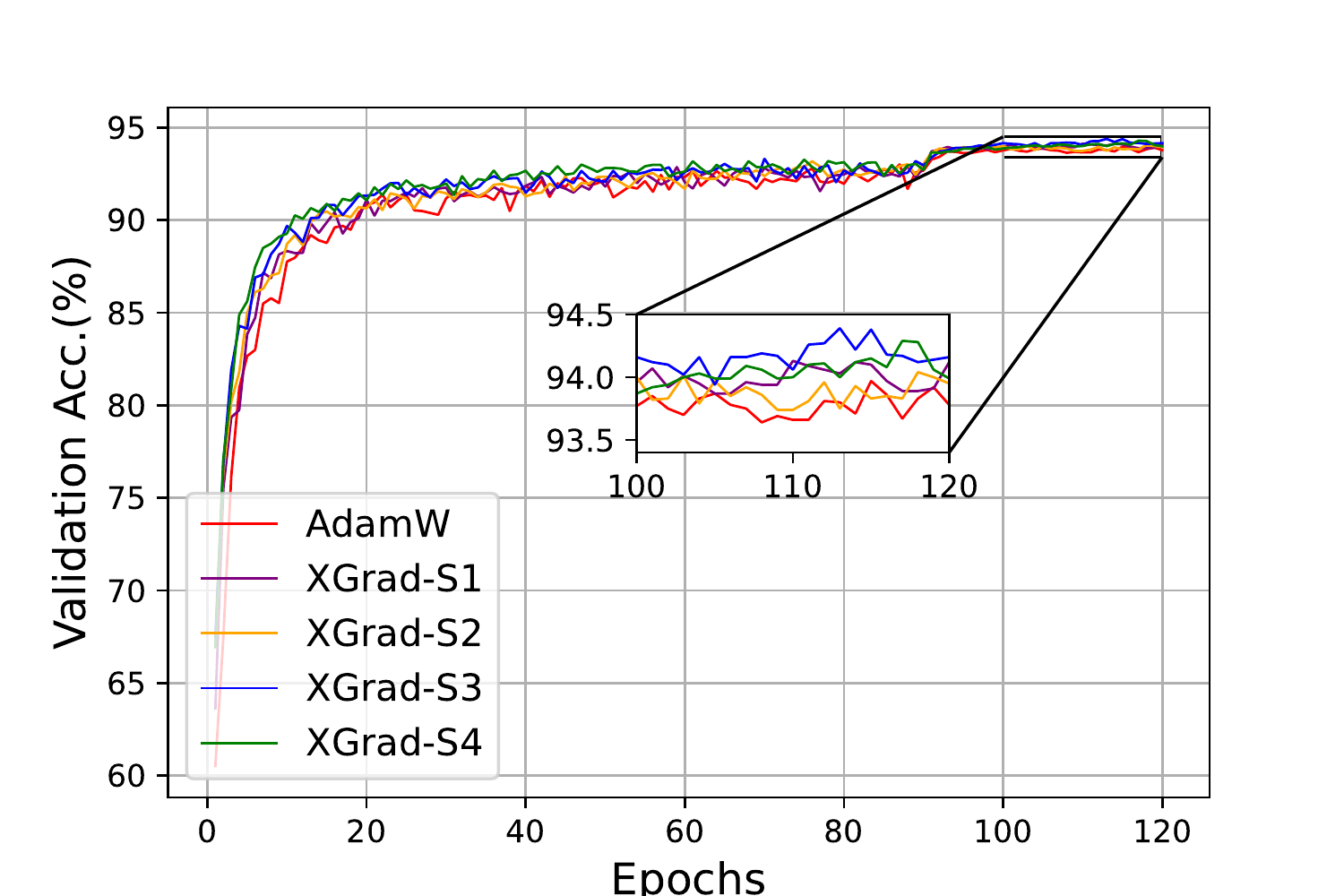}\label{comp-adamw-densenet-acc}}
	\quad
	\subfloat[Inception-V3]{\includegraphics[width=.28\textwidth]{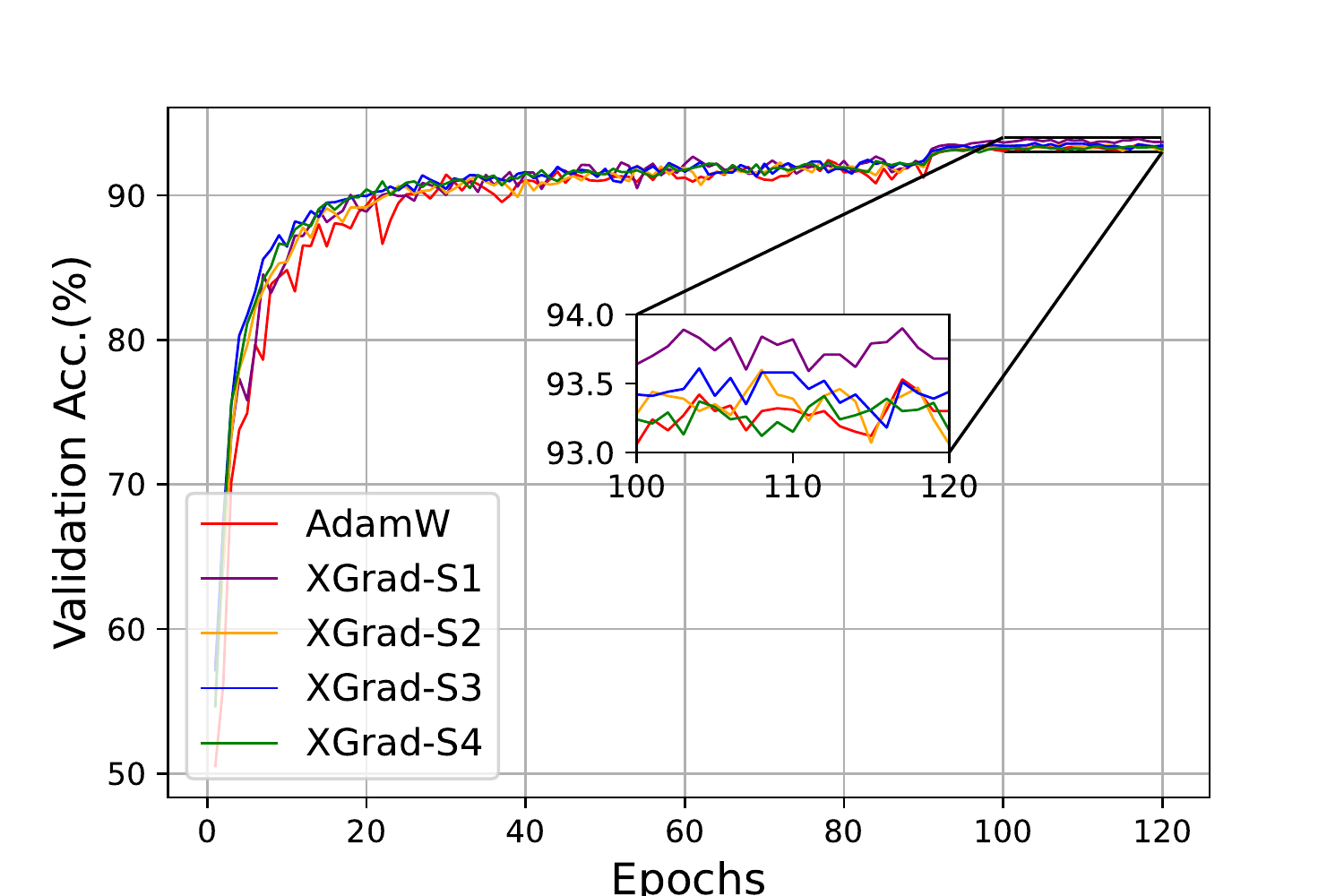}\label{comp-adamw-inceptionv3-acc}}
	\subfloat[LSTM-1]{\includegraphics[width=.28\textwidth]{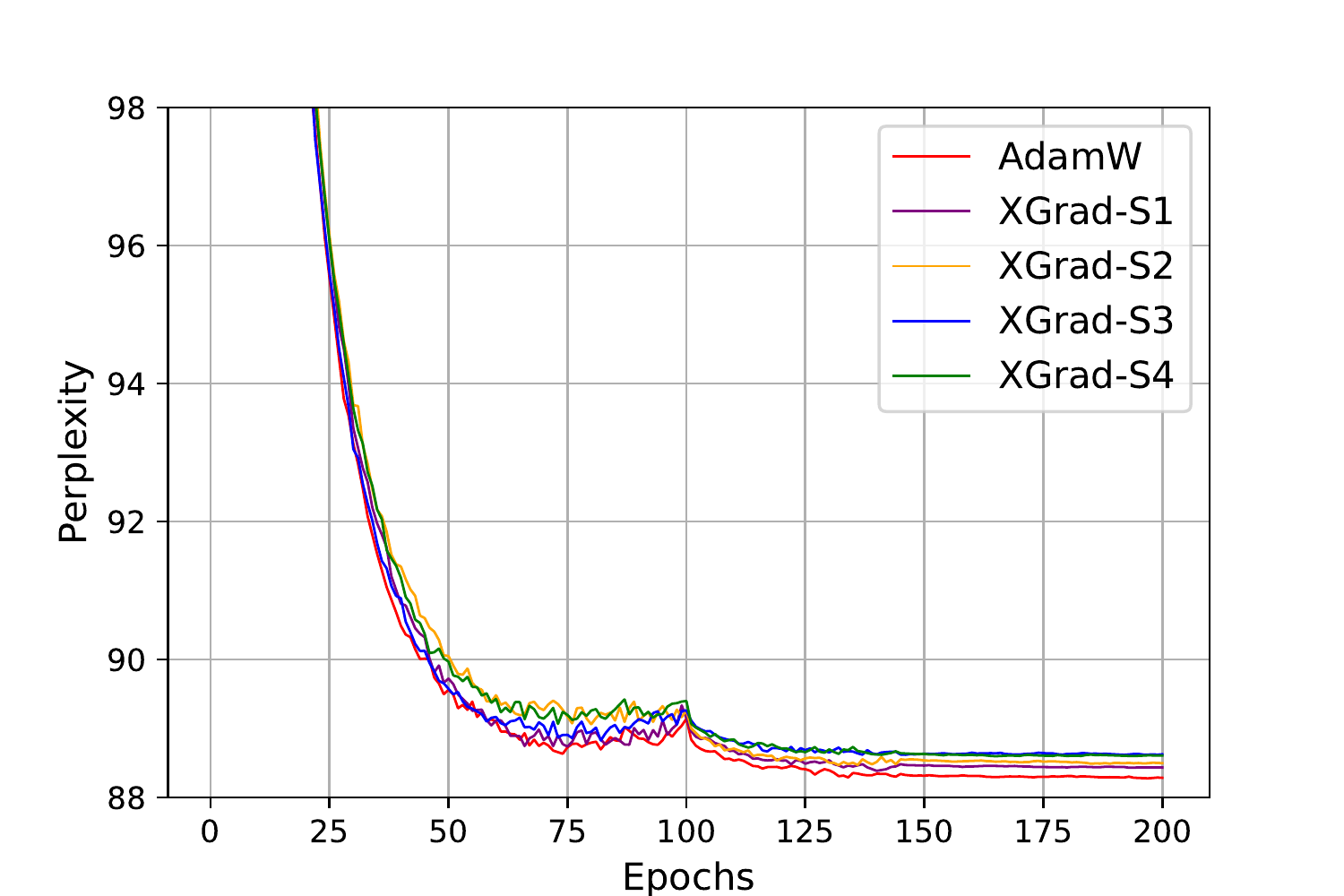}\label{comp-adamw-lstm1}}
	\subfloat[LSTM-2]{\includegraphics[width=.28\textwidth]{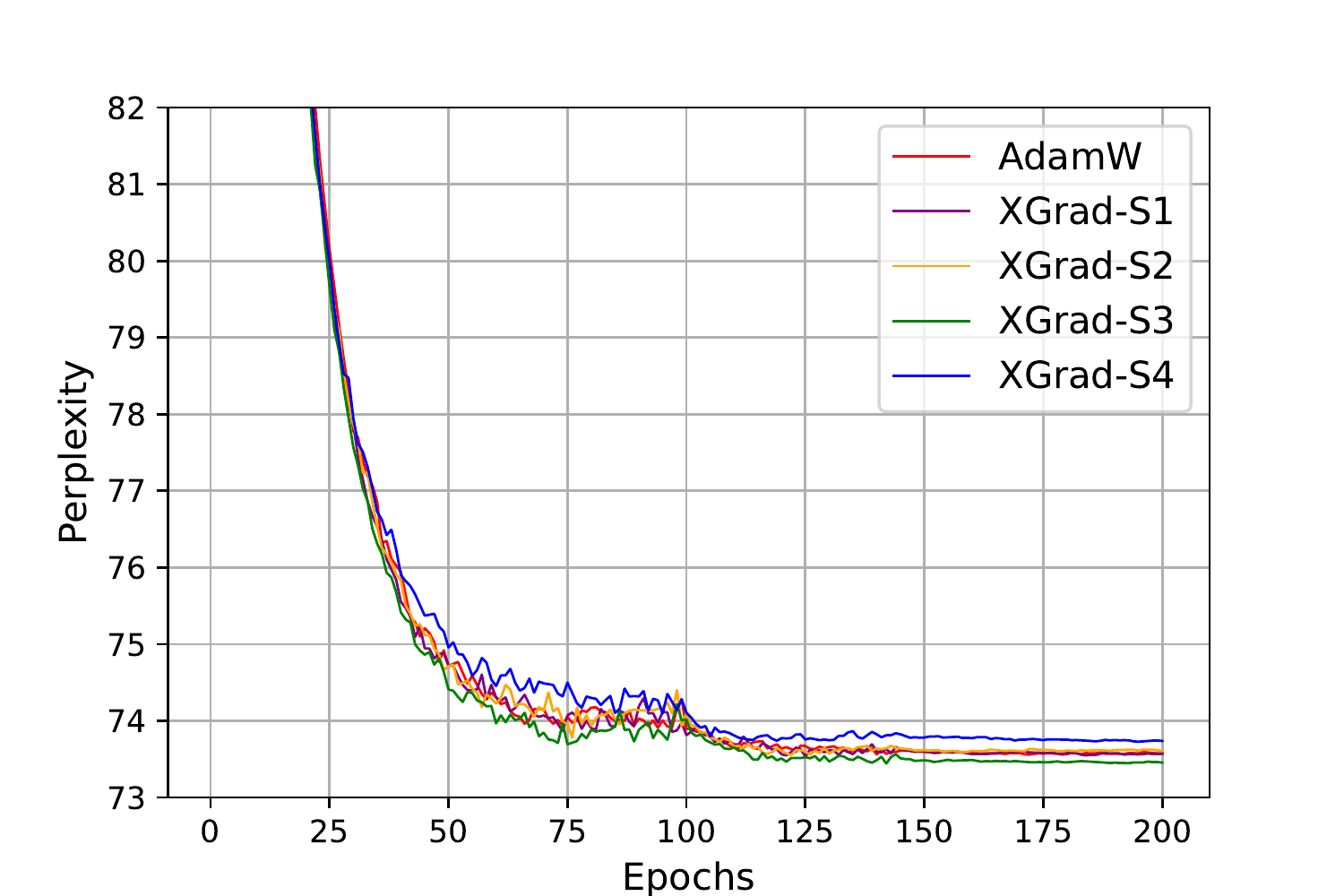}\label{comp-adamw-lstm2}}
	\quad
	\subfloat[LSTM-3]{\includegraphics[width=.28\textwidth]{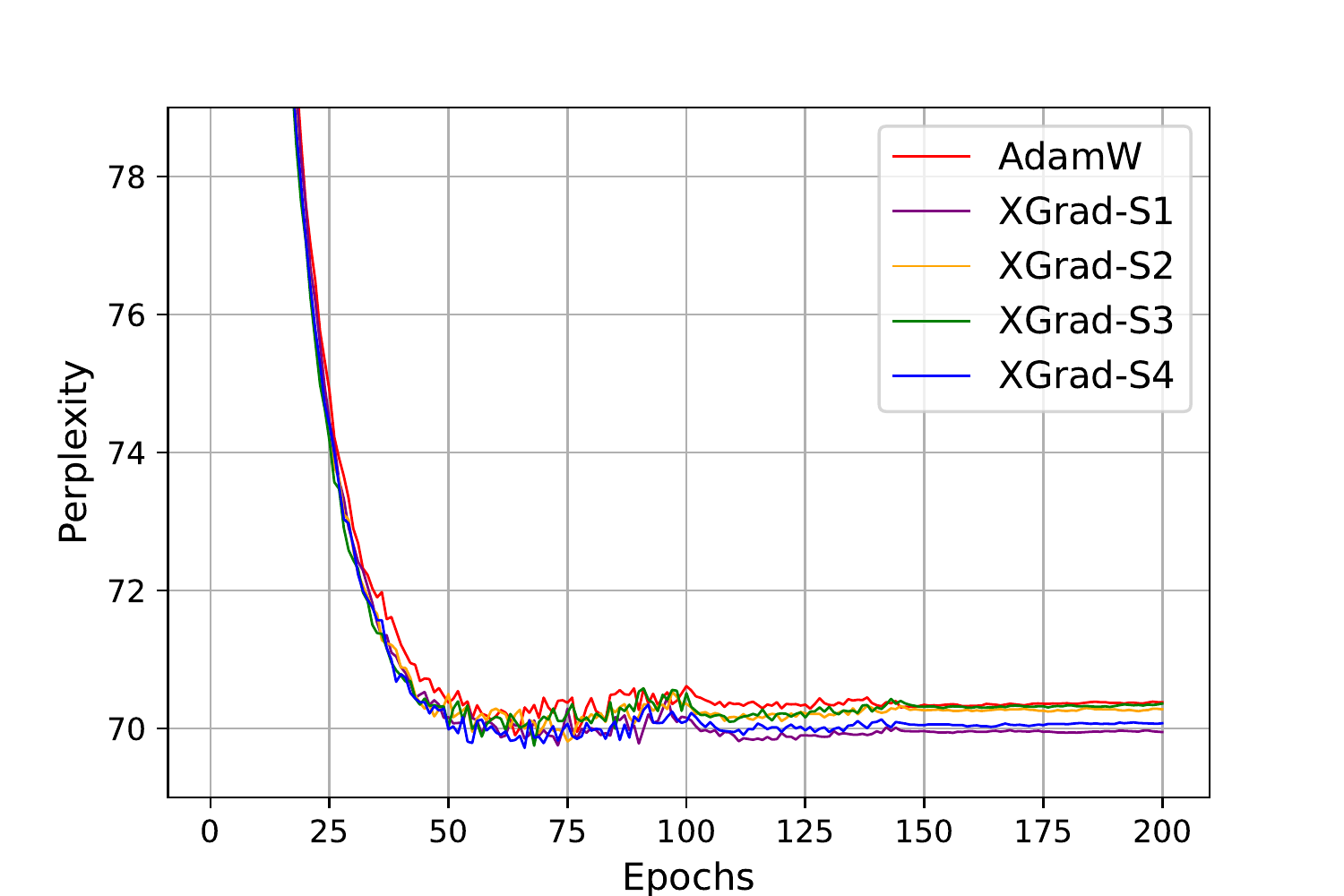}\label{comp-adamw-lstm3}}
	\subfloat[GNMT-8]{\includegraphics[width=.28\textwidth]{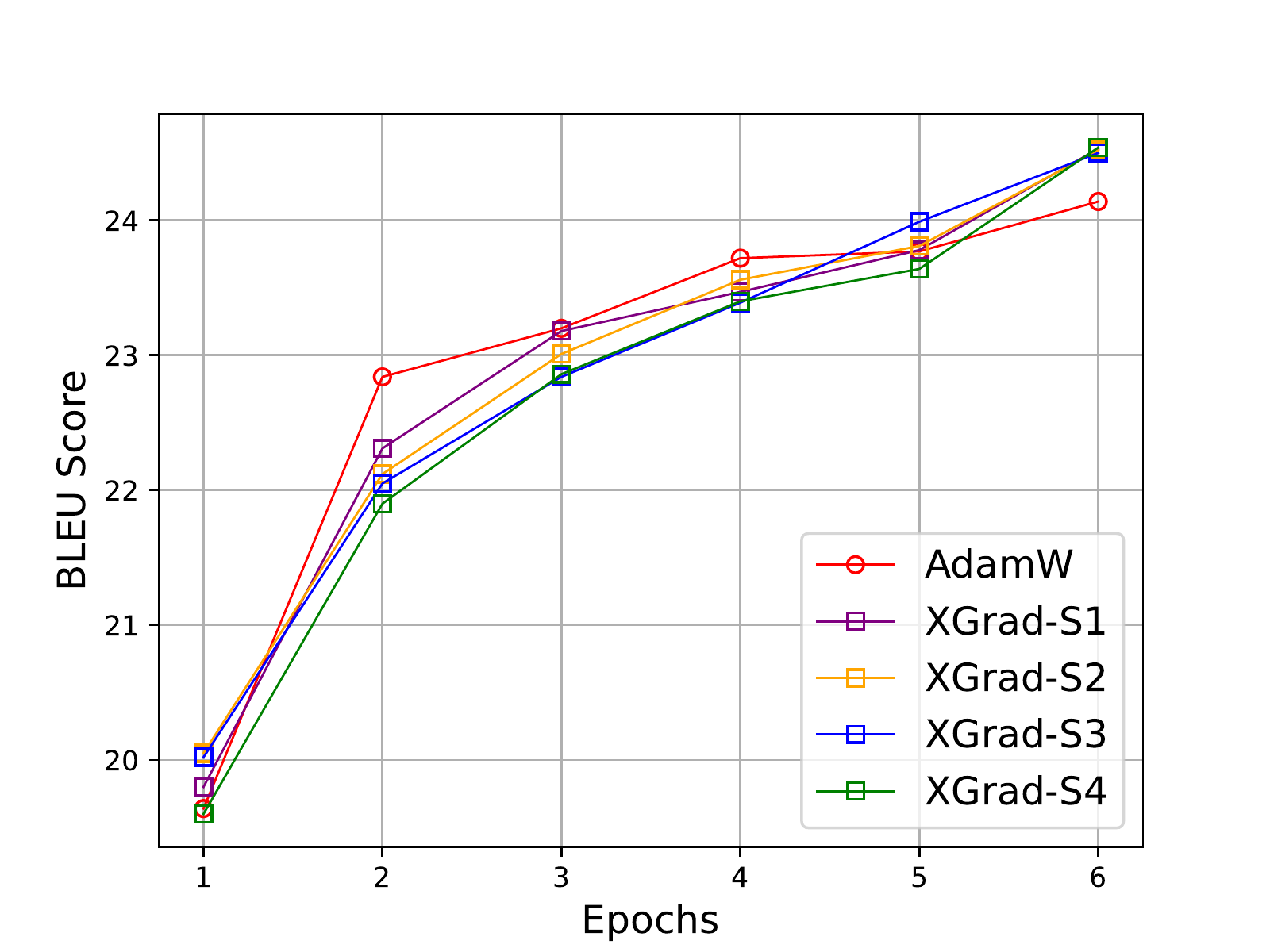}\label{comp-adamw-gnmt8}}
	\subfloat[VAE]{\includegraphics[width=.28\textwidth]{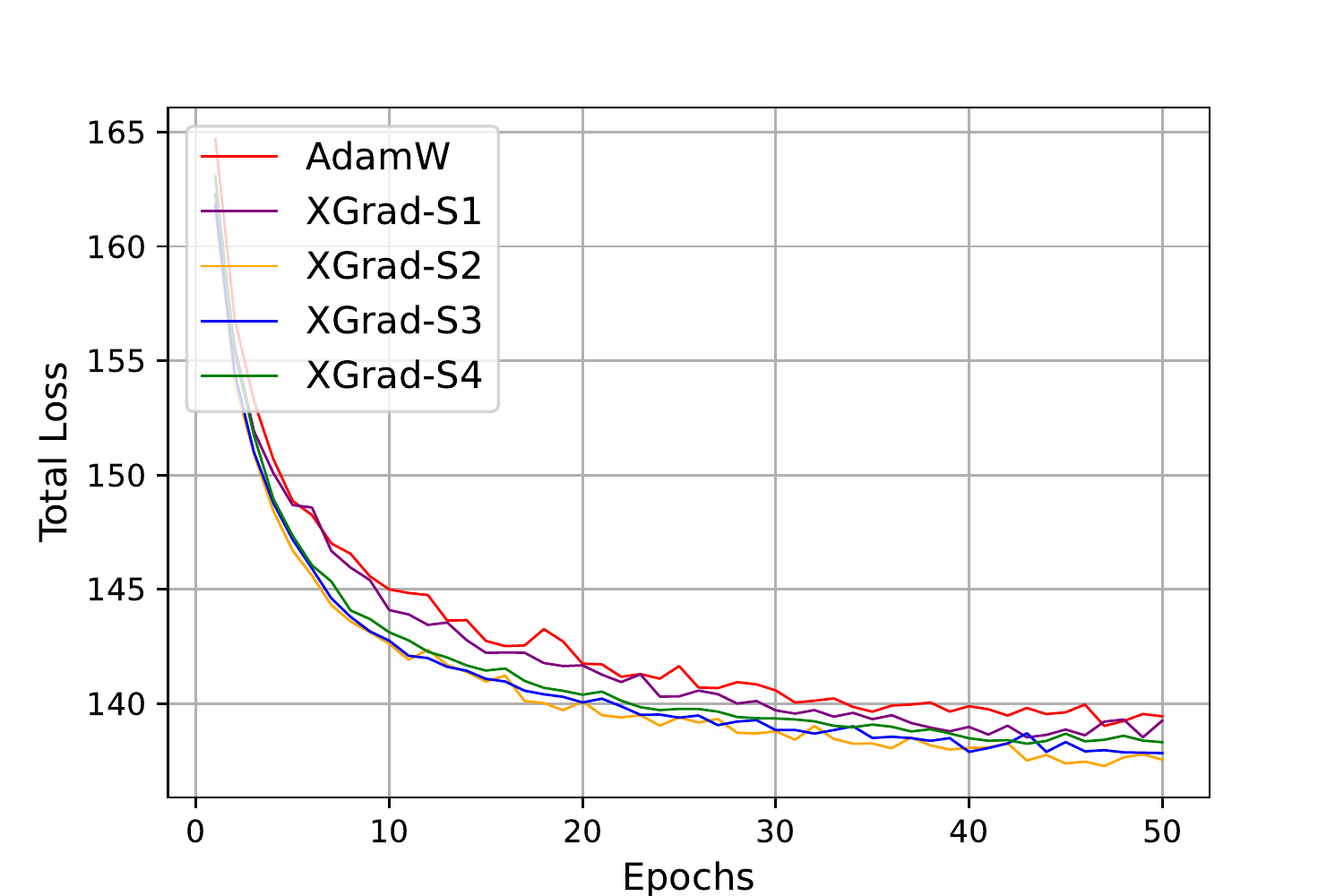}\label{comp-adamw-vae}}
	\caption{Comparison of AdamW and XGrad.  Figures~\ref{comp-adamw-lenet-acc}, ~\ref{comp-adamw-resnet34-acc}, ~\ref{comp-adamw-densenet-acc}, and~\ref{comp-adamw-inceptionv3-acc}: Top-1 accuracy (higher is better) vs. Epochs; Figures~\ref{comp-adamw-lstm1}, ~\ref{comp-adamw-lstm2}, and~\ref{comp-adamw-lstm3}: Perplexity (lower is better) vs. Epochs; Figure~\ref{comp-adamw-gnmt8}: BLEU score (higher is better) vs. epochs; Figure~\ref{comp-adamw-vae}: Total loss (lower is better) vs. Epochs.}
	\label{comp-adamw-acc-cifar10}
\end{figure*}

\begin{table*}[h!]
	\centering
	\caption{Summarization of best model accuracy of XGrad and AdamW. Maximum top-1 accuracy for LeNet, ResNet-34, DenseNet-121,  and Incepiton-V3; Minimum perplexity for LSTM-1/2/3; Maximum BLEU score for GNMT-8; Maximum Dev set accuracy for BERT$_{\text{BASE}}$; Minimum total loss for VAE; Minimum FID score for WGAN. The best model accuracy results are highlighted in boldface.}
	\label{table:adamw}
	\setlength{\tabcolsep}{1.8mm}
	\begin{tabular}{c|ccccccccccc}
		\toprule
		Optimizers & LeNet & ResNet-34 & DenseNet-121 & Inception-V3  & LSTM-1  & LSTM-2 &  LSTM-3  &GNMT-8  & BERT$_{\text{BASE}}$   & VAE &WGAN \\
		\midrule
		\multicolumn{12}{c}{Best Accuracy} \\
		\midrule
		\makecell{AdamW} & 88.61\% & 94.03\% & 93.97\% & 93.53\% &\textbf{88.28}& 73.56 & 69.90 & 24.14 & 85.29\%  & 139.03 &97.02    \\
		XGrad-S1 & 89.04\% & 94.06\%  & 94.13\% &  \textbf{93.90}\% &  88.39& 73.53 & 69.75 &  24.53 & 85.05\% &  138.53 &\textbf{74.94}   \\
		XGrad-S2 & 88.33\% & 93.95\% & 94.04\% & 93.60\% & 88.47& 73.55 & 69.81&  24.52 & 84.56\% & \textbf{137.27} & 75.42    \\
		XGrad-S3 & \textbf{89.23}\% & 94.37\% & \textbf{94.39}\% & 93.61\% & 88.62& \textbf{73.44} &  69.75 & 24.50  & 84.31\% & 137.84 &  79.18   \\
		XGrad-S4 & 88.42\%  & \textbf{94.44}\% & 94.29\% & 93.41\%  & 88.60 & 73.73 & \textbf{69.72} & \textbf{24.54}   & \textbf{87.01}\% & 138.25&  89.40   \\
		\bottomrule
	\end{tabular}
\end{table*}

Figure~\ref{comp-adam-acc-cifar10} illustrates the learning curves about model accuracy vs. epochs. Table~\ref{table:adam} summarizes the obtained best model accuracy. 
The experiment results verify the effectiveness of XGrad in boosting the convergence and generalization of Adam. First, Figures~\ref{comp-adam-lenet-acc}, \ref{comp-adam-resnet34-acc}, \ref{comp-adam-densenet-acc}, and~\ref{comp-adam-inceptionv3-acc} validate the superiority of XGrad over Adam when training CNN models. XGrad always tends to achieve higher accuracy than Adam when training them with the same number of epochs. Figures~\ref{comp-adam-lstm1}, \ref{comp-adam-lstm2}, and \ref{comp-adam-lstm3} illustrate similar phenomena. XGrad tends to demonstrate a smaller perplexity value with the increase of the epochs. The learning curves depicted in Figures~\ref{comp-adam-gnmt8} and~\ref{comp-adam-vae} show that XGrad tends to get a higher BLEU score and lower total loss than Adam. The experiment results shown in Table~\ref{table:adam} again validate the effectiveness of XGrad. Compared to Adam, XGrad respectively achieves an accuracy improvement of 0.71\%, 0.85\%, 1.00\%, and 0.47\% when training LeNet, ResNet-34, DenseNet-121, and Inception-V3. This means that XGrad leads to an average of 0.76\% accuracy improvement over Adam. Meanwhile, XGrad achieves 0.20, 0.57, and 0.46 less perplexity than Adam when training LSTM-1, LSTM-2, and LSTM-3, respectively. XGrad also obtains a 0.74 higher BLEU score than Adam when training GNMT-8 and gets 0.49\% higher accuracy than Adam when training the BERT$_{\text{BASE}}$. Furthermore, XGrad gets 1.37 less total loss than Adam when training VAE and also obtains a much lower FID score when training WGAN (72.90 vs. 95.60).

\subsubsection{Comparisons of XGrad with  AdamW}
In this section, we again selected LeNet, ResNet-34, DenseNet-121, Inception-V3,  LSTM-1, LSTM-2, LSTM-3, GNMT-8, BERT$_{\text{BASE}}$, VAE, and WGAN as the benchmark models and evaluated them with the same experimental settings as described in Section~\ref{sec:comp-xgrad-adam}.  Figure~\ref{comp-adamw-acc-cifar10} depicts the learning curves about model accuracy vs. epochs. Table~\ref{table:adamw} summarizes the obtained best model accuracy. 

We can reach the following conclusions based on the observation of the experiment results. First, XGrad performs better than AdamW on the image classification tasks. On average, XGrad yields 0.46\% (up to 0.62\%) top-1 accuracy improvement over AdamW. Second, XGrad also outperforms AdamW on NLP tasks. XGrad performs slightly better than AdamW on the language modeling tasks. In particular, XGrad is inferior to AdamW when training LSTM-1 (88.39 vs. 88.28) but gets 0.12 and 0.18 less perplexity than AdamW when training LSTM-2 and LSTM-3, respectively. XGrad also achieves a higher BLEU score (24.54 versus 24.14) when training GNMT-8 and obtains 1.72\% higher accuracy than AdamW when training BERT$_{\text{BASE}}$ on MRPC. Third,  for both the image classification and NLP tasks, the weight prediction step has a slight effect on the model accuracy generated by the XGrad. However, the performance of XGrad varies sharply with different weight prediction steps when training sophisticated DNN models such as BERT$_{\text{BASE}}$ and WGAN. 



\begin{figure*}[h!]
	\centering
	\subfloat[LeNet]{\includegraphics[width=.28\textwidth]{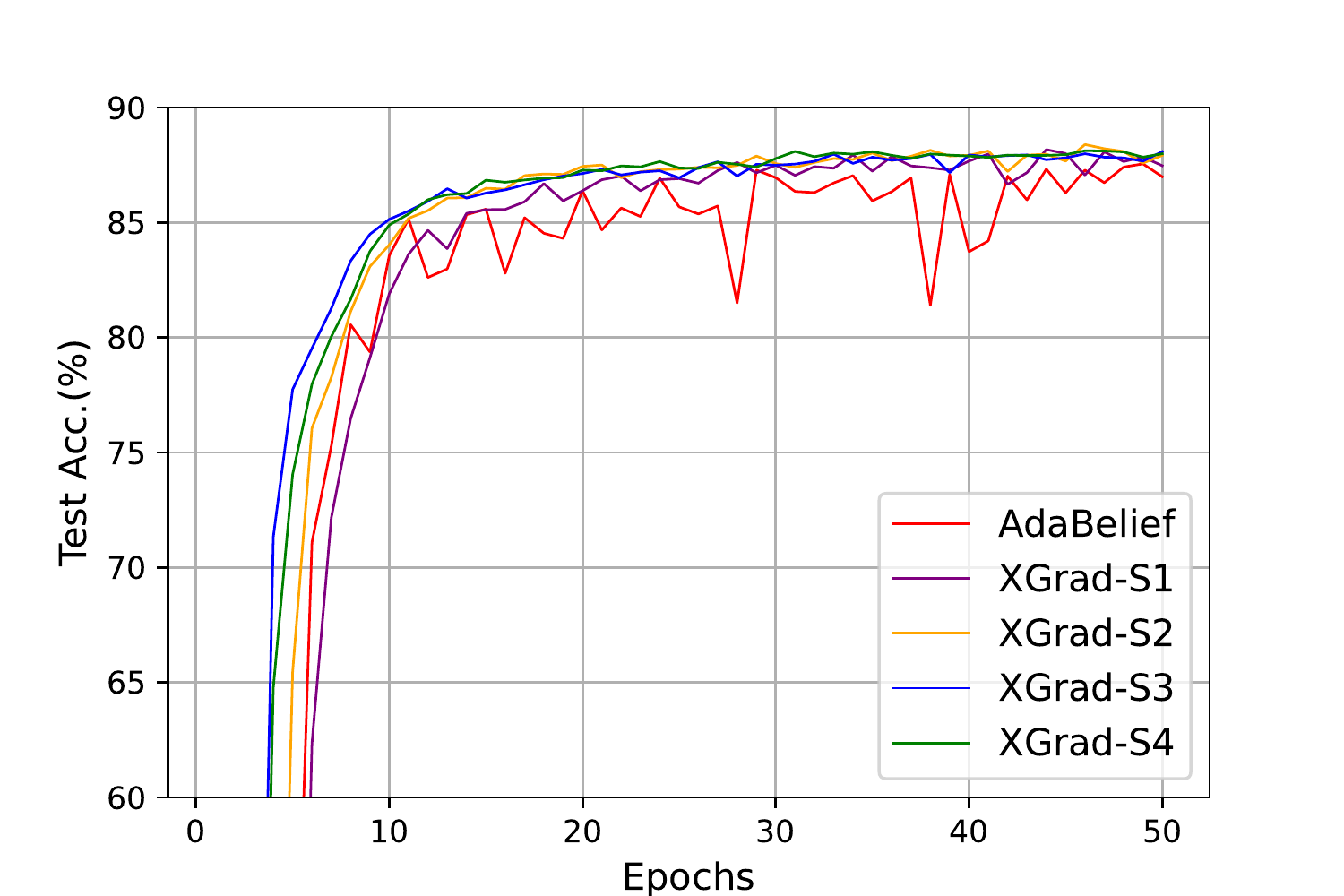}\label{comp-adabelief-lenet-acc}}
	\subfloat[AlexNet]{\includegraphics[width=.28\textwidth]{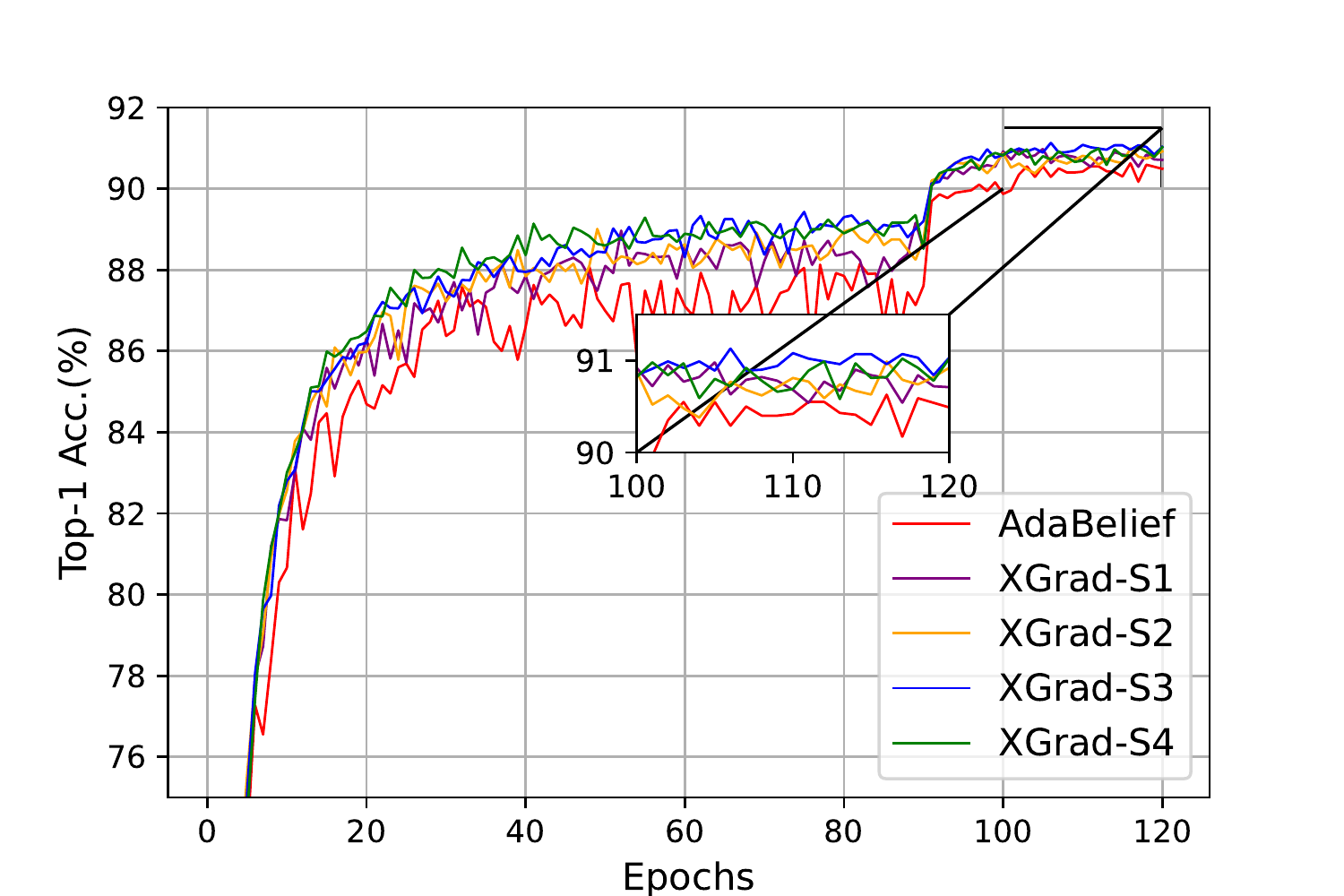}\label{comp-adabelief-alexnet-acc}}
	\subfloat[VGG-11]{\includegraphics[width=.28\textwidth]{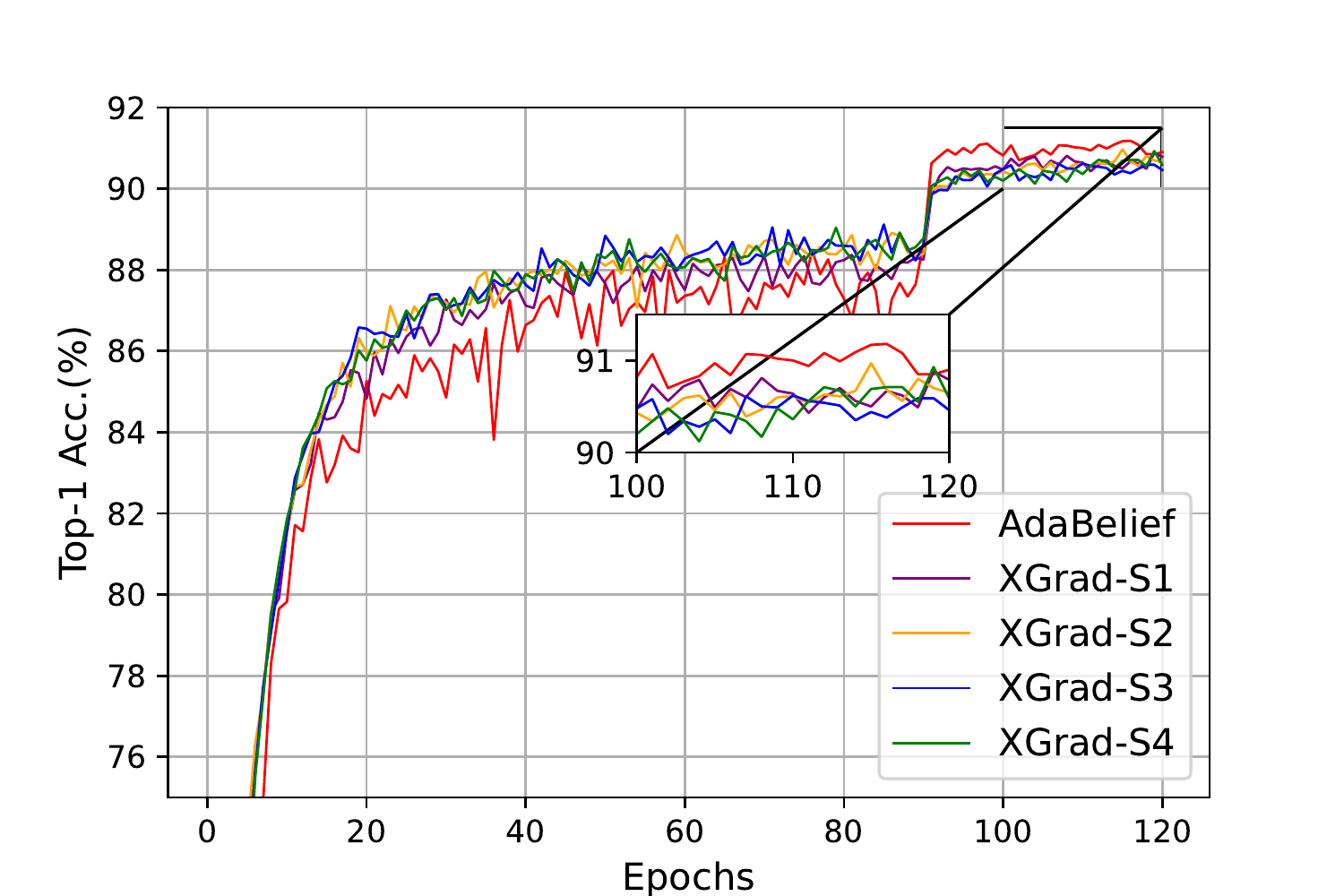}\label{comp-adabelief-vgg11-acc}}
	\quad
	\subfloat[ResNet-34]{\includegraphics[width=.28\textwidth]{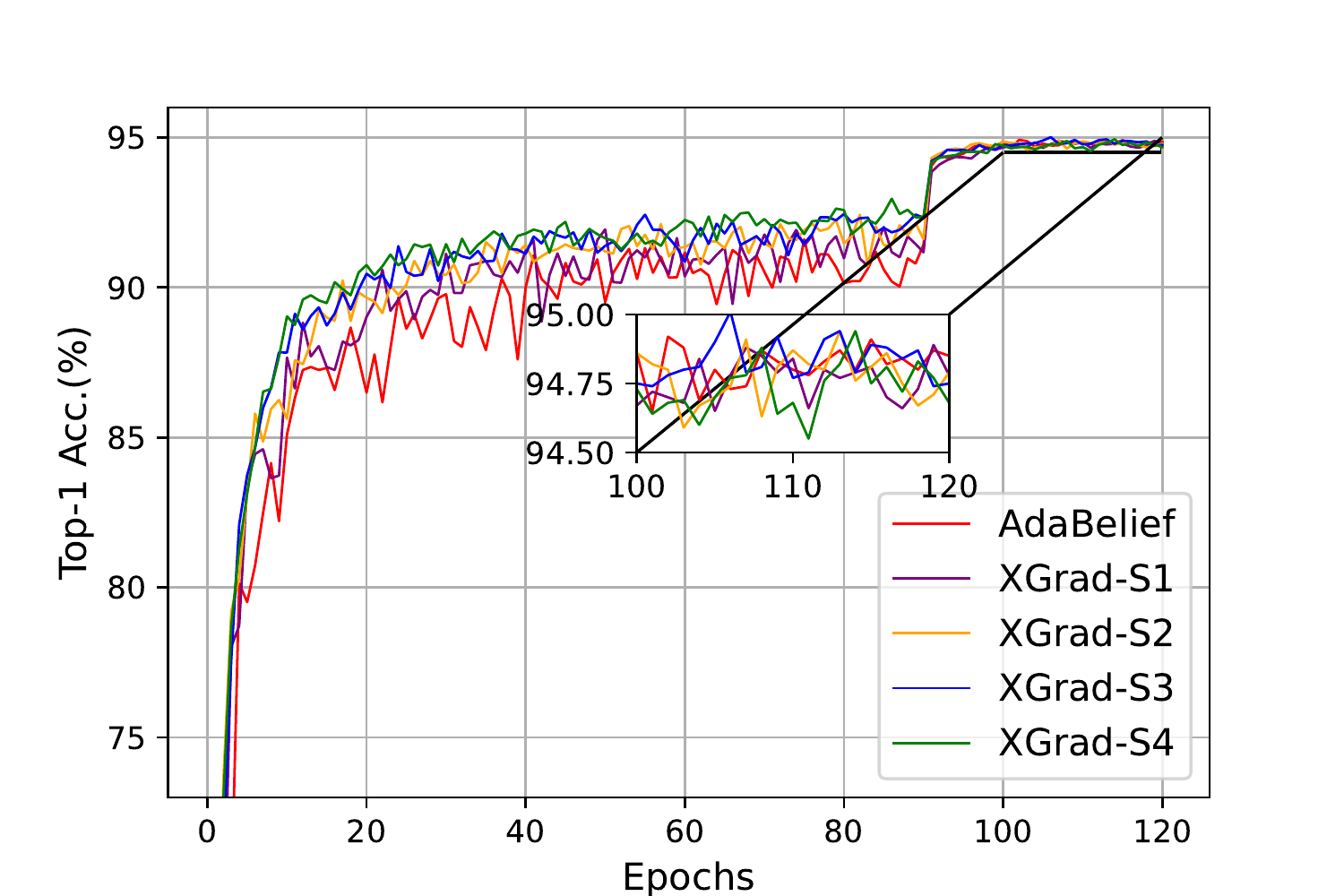}\label{comp-adabelief-resnet34-acc}}
	\subfloat[DenseNet-121]{\includegraphics[width=.28\textwidth]{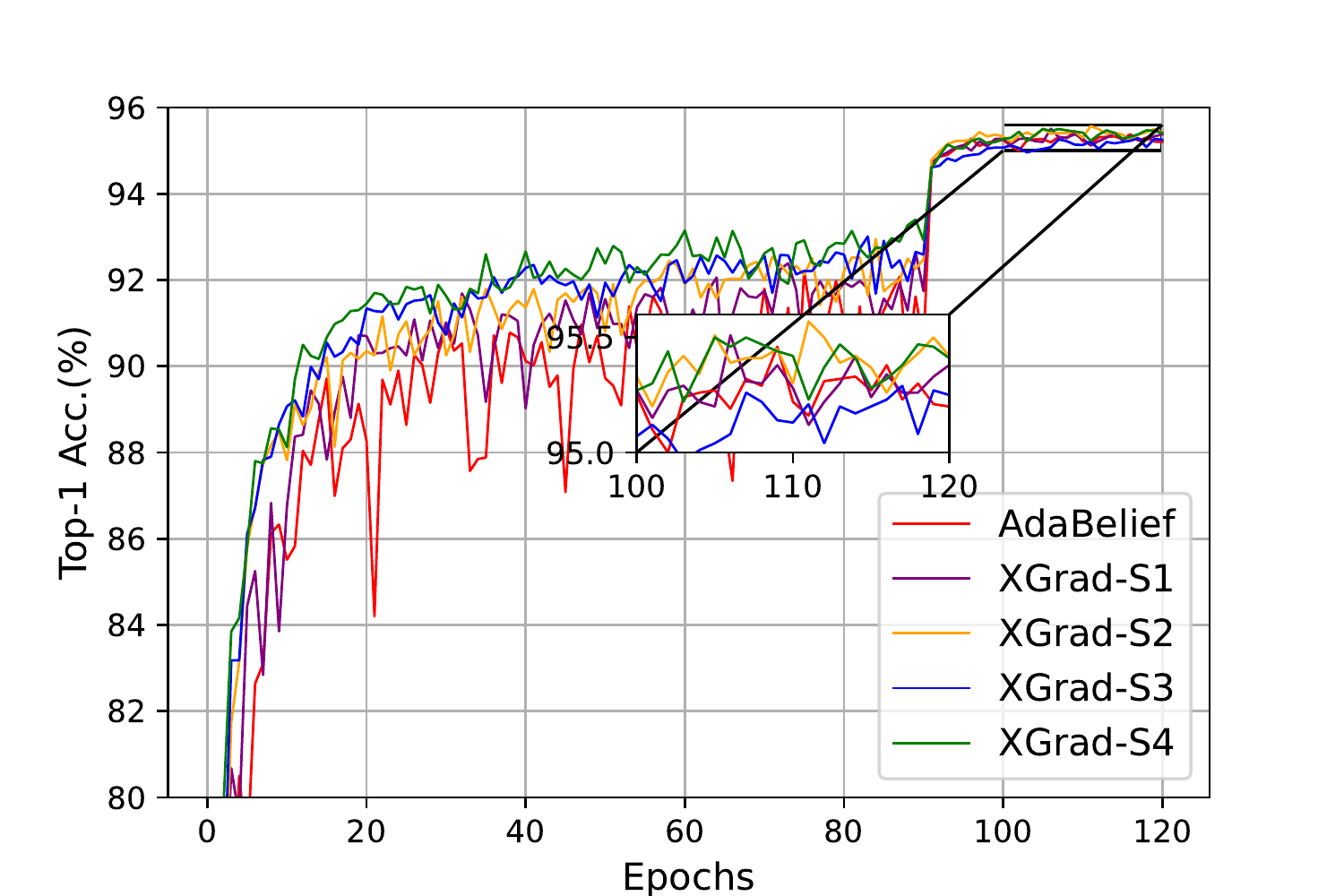}\label{comp-adabelief-densenet-acc}}
	\subfloat[LSTM-1]{\includegraphics[width=.28\textwidth]{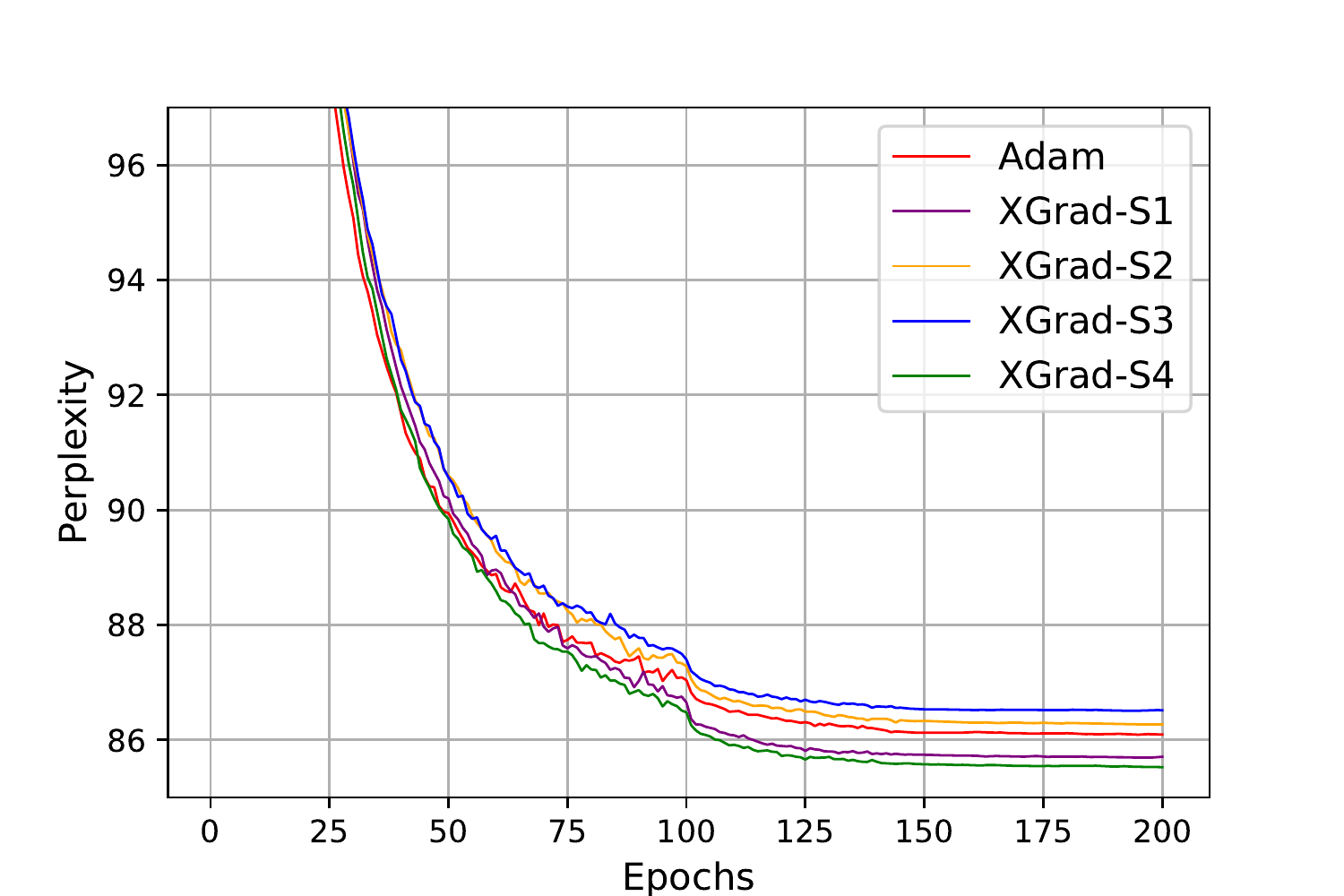}\label{comp-adabelief-lstm1}}
	\quad
	\subfloat[LSTM-2]{\includegraphics[width=.28\textwidth]{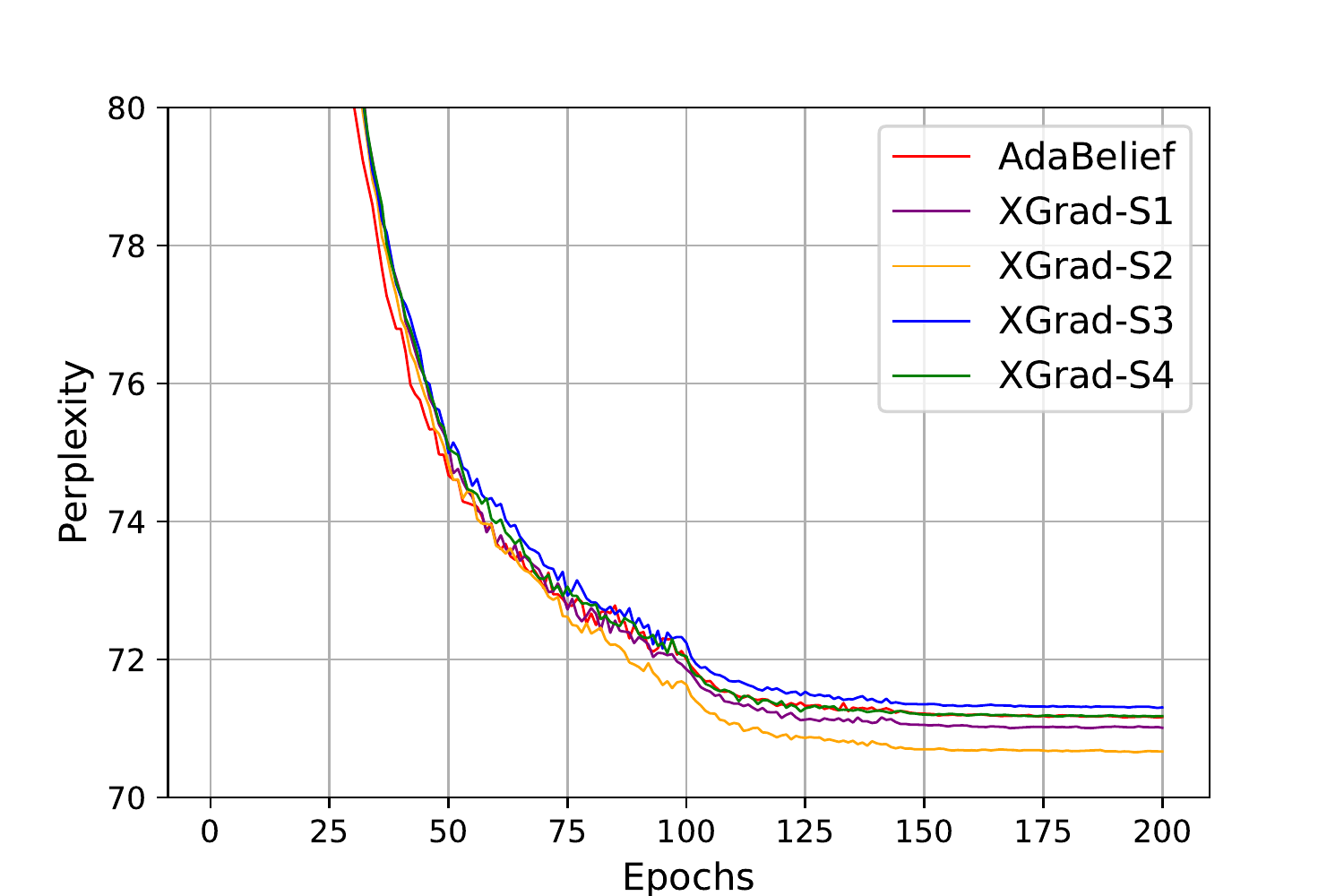}\label{comp-adabelief-lstm2}}
	\subfloat[LSTM-3]{\includegraphics[width=.28\textwidth]{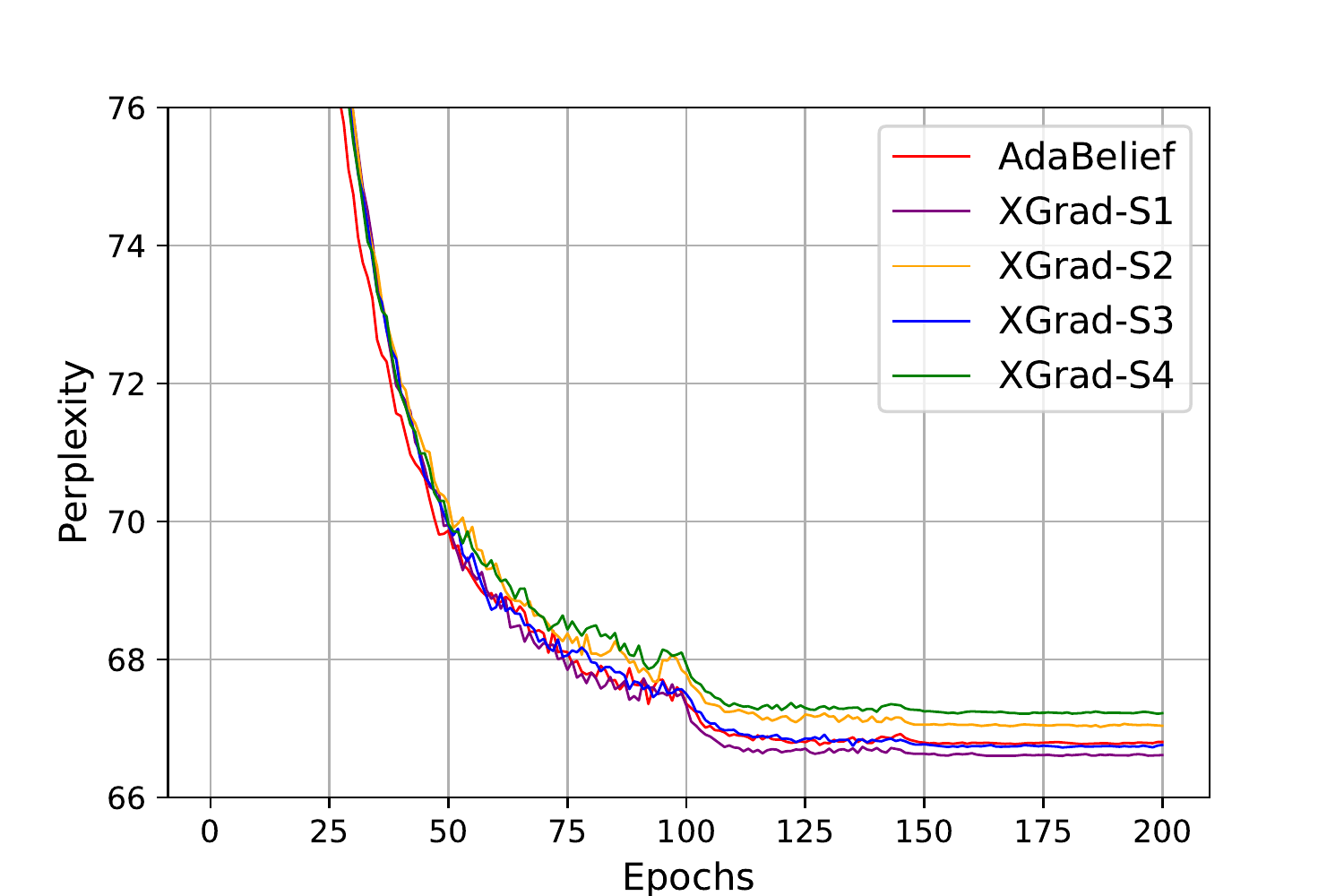}\label{comp-adabelief-lstm3}}
	\subfloat[VAE]{\includegraphics[width=.28\textwidth]{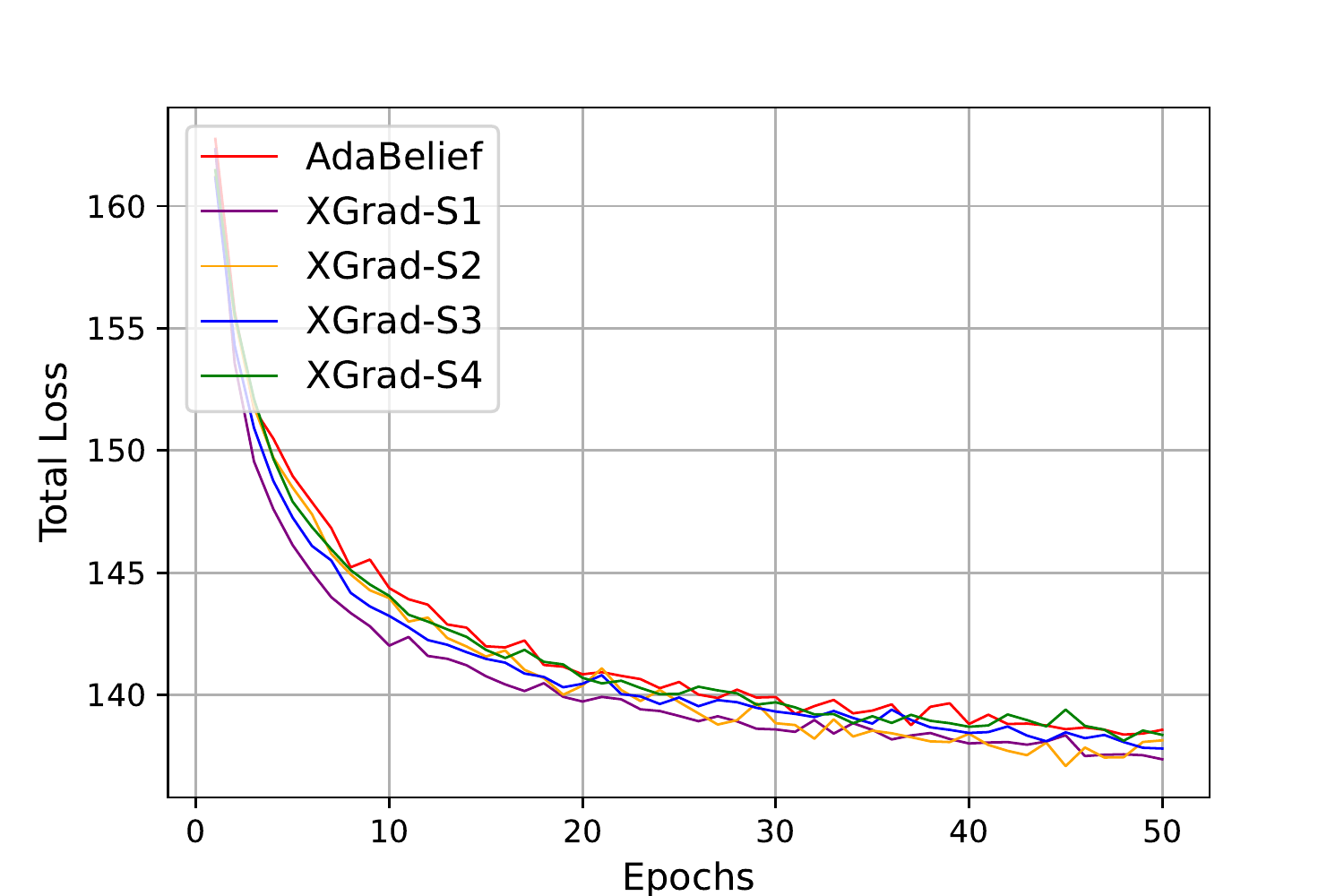}\label{comp-adabelief-vae}}
	\caption{Comparison of AdaBelief and XGrad. Figures~\ref{comp-adabelief-lenet-acc}, ~\ref{comp-adabelief-alexnet-acc}, ~\ref{comp-adabelief-vgg11-acc}, ~\ref{comp-adabelief-resnet34-acc}, and~\ref{comp-adabelief-densenet-acc}: Top-1 accuracy (higher is better) vs. Epochs; Figures~\ref{comp-adabelief-lstm1}, ~\ref{comp-adabelief-lstm2}, and~\ref{comp-adabelief-lstm3}: Perplexity (lower is better) vs. Epochs; Figure~\ref{comp-adabelief-vae}: Total loss (lower is better) vs. Epochs.}
	\label{comp-adabelief-acc-cifar10}
\end{figure*}

\begin{table*}[h!]
	\centering
	\caption{Summarization of best model accuracy of XGrad and AdaBelief. Maximum top-1 accuracy for LeNet, AlexNet, VGG-11, ResNet-34, and DenseNet-121; Minimum perplexity for LSTM-1/2/3; Maximum Dev set accuracy for BERT$_{\text{BASE}}$; Minimum total loss for VAE; Minimum FID score for WGAN. The best model accuracy results are highlighted in boldface.}
	\label{table:adabelief}
	\setlength{\tabcolsep}{2.0mm}
	\begin{tabular}{c|ccccccccccc}
		\toprule
		Optimizers & LeNet & AlexNet &VGG-11 &ResNet-34 & DenseNet-121  &LSTM-1  & LSTM-2  & LSTM-3  & BERT$_{\text{BASE}}$ & VAE &WGAN   \\
		\midrule
		\multicolumn{12}{c}{Best Accuracy} \\
		\midrule
		\makecell{AdaBelief} & 87.55\%  &  90.63\%& \textbf{91.18}\%  & 94.92\% & 95.46\% & 86.09 & 71.16&  66.76 &  84.56\%  & 138.38 & 86.66  \\
		XGrad-S1 & 88.17\% & 90.98\%   & 90.88\% & 94.89\%  & 95.51\% &  85.69 & 71.00  & \textbf{66.60} & 83.09\% & 137.37 & 86.05  \\
		XGrad-S2 & \textbf{88.39}\%& 90.99\%&  90.97\% & 94.94\%&  \textbf{95.57}\%& 86.27  & \textbf{70.66} &67.02 & 84.07\%  & \textbf{137.09} & \textbf{83.97}   \\
		XGrad-S3 &  88.08\%& \textbf{91.13}\% & 90.62\% & \textbf{95.01}\%  &  95.29\% & 86.51 & 71.29& 66.72 & 84.31\%  & 137.81&  97.42 \\
		XGrad-S4 & 88.12\% &  91.02\% &  90.93\% & 94.94\% &  95.50\%& \textbf{85.52}&71.17&  67.21 & \textbf{85.54}\% & 138.13 & 89.79   \\
		\bottomrule
	\end{tabular}
\end{table*}









\subsubsection{Comparisons of XGrad and AdaBelief}\label{sec:comp-xgrad-adabelief}
In this section, we compared XGrad with AdaBelief using 11 different DNN models including LeNet, AlexNet, VGG-11, ResNet-34, DenseNet-121, LSTM-1, LSTM-2, LSTM-3, BERT$_{\text{BASE}}$, VAE, and WGAN. 
Figure~\ref{comp-adabelief-acc-cifar10} shows learning curves about model accuracy vs. epochs. Table~\ref{table:adabelief} lists the obtained best model accuracy.

The learning curves depicted in Figure~\ref{comp-adabelief-acc-cifar10} again validate the effectiveness of XGrad. Given the same number of training epochs, XGrad always tends to achieve better model accuracy than AdaBelief. When training LeNet, AlexNet, VGG-11, ResNet-34, and DenseNet-121, XGrad respectively yields a top-1 accuracy improvement of 0.84\%, 0.50\%, -0.21\%, 0.09\%, and 0.11\% over AdaBelief, leading to an average of 0.27\% top-1 accuracy improvement. When training LSTM-1, LSTM-2, and LSTM-3, XGrad consistently achieves lower perplexity than AdaBelief, averaging 0.41 less perplexity than AdaBelief. For training BERT$_{\text{BASE}}$ on MRPC, XGrad achieves 0.98\% higher Dev set accuracy than AdaBelief. Moreover, compared to AdaBelief, XGrad respectively gets 1.29 less total loss when training VAE and 2.69 less FID score when training WGAN.


\subsubsection{Comparisons of XGrad and AdaM3}\label{sec:comp-xgrad-adam3}
\begin{figure*}[h!]
	\centering
	\subfloat[LeNet]{\includegraphics[width=.25\textwidth]{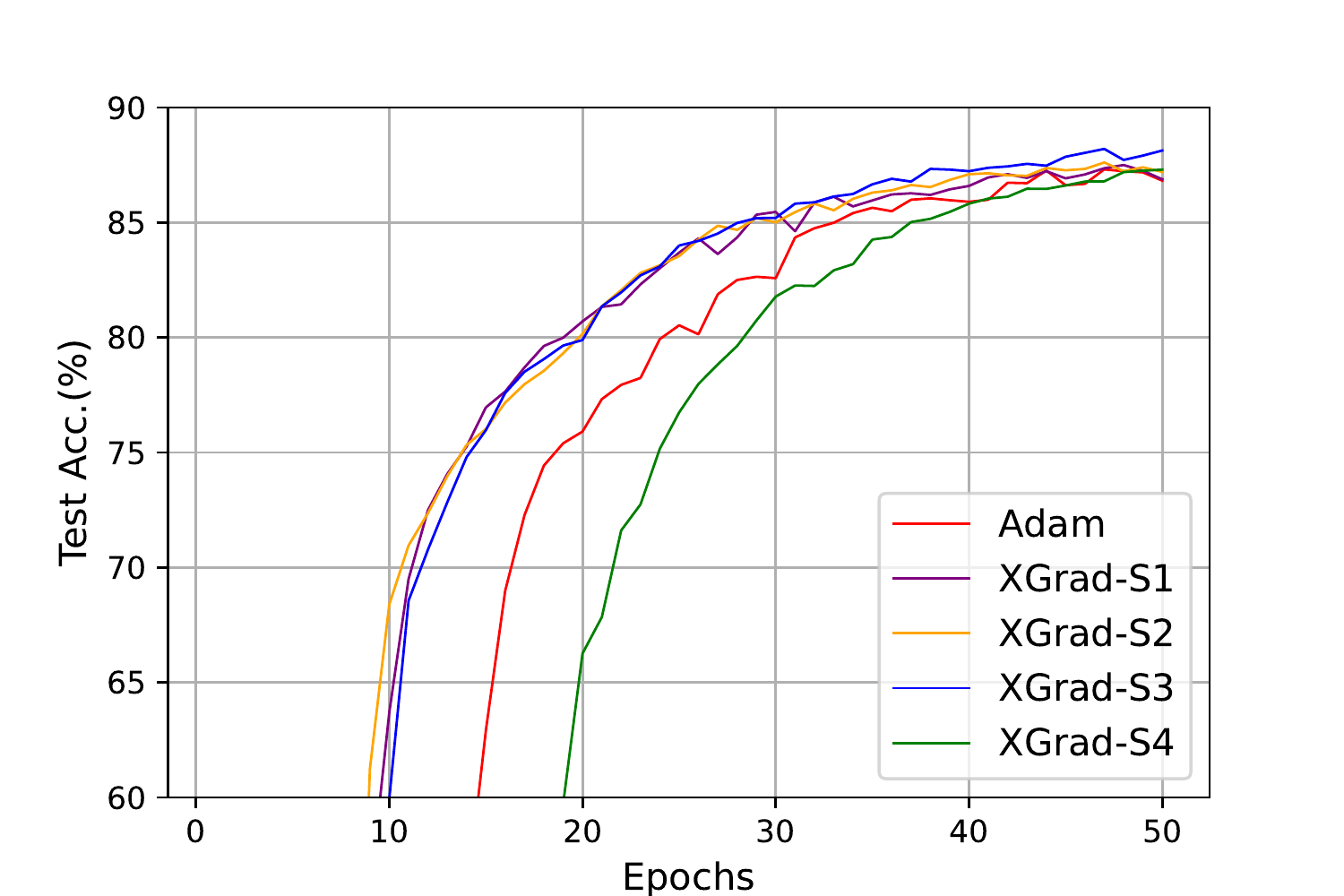}\label{comp-adam3-lenet-acc}}
	\subfloat[AlexNet]{\includegraphics[width=.25\textwidth]{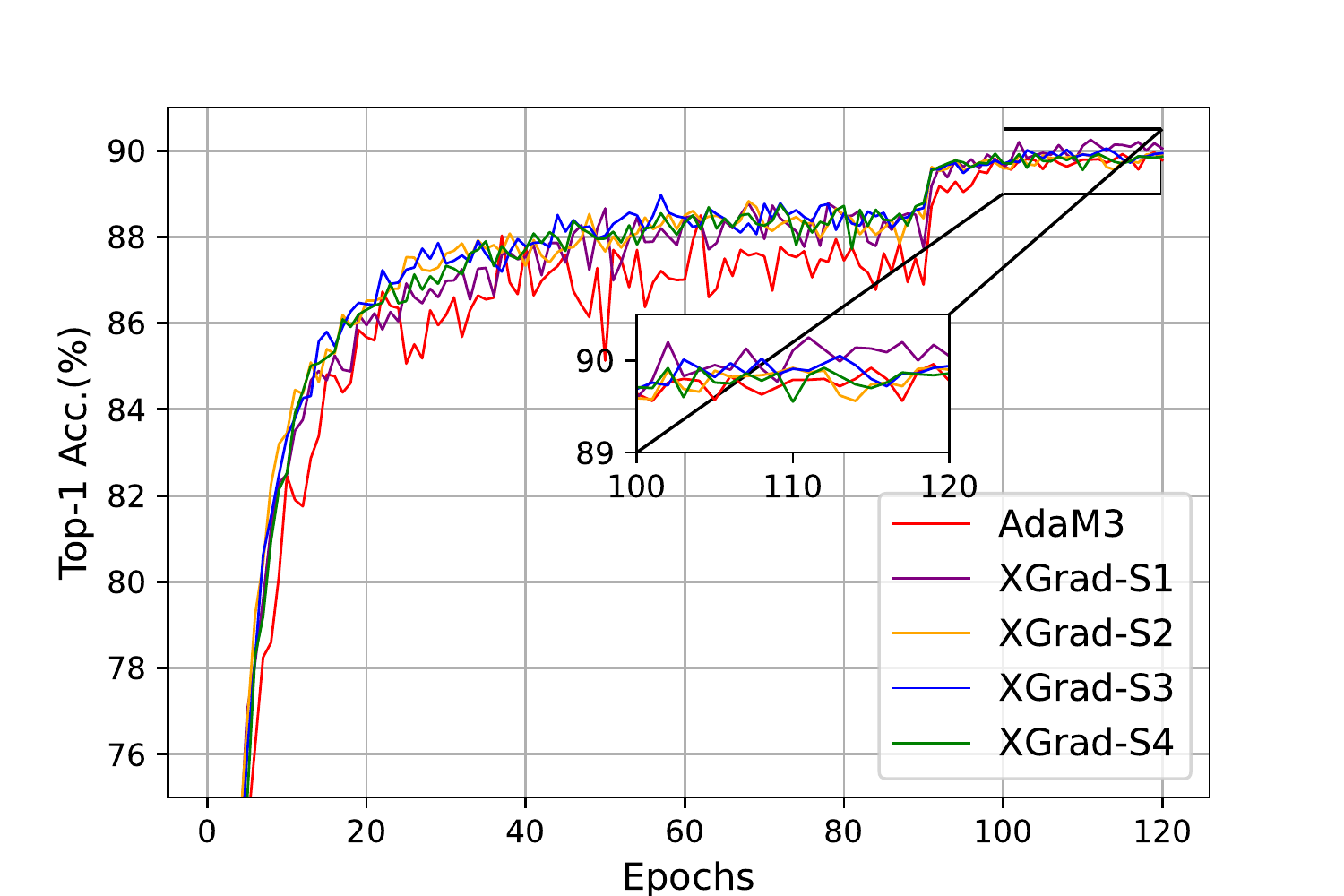}\label{comp-adam3-alexnet-acc}}
	\subfloat[VGG-11]{\includegraphics[width=.25\textwidth]{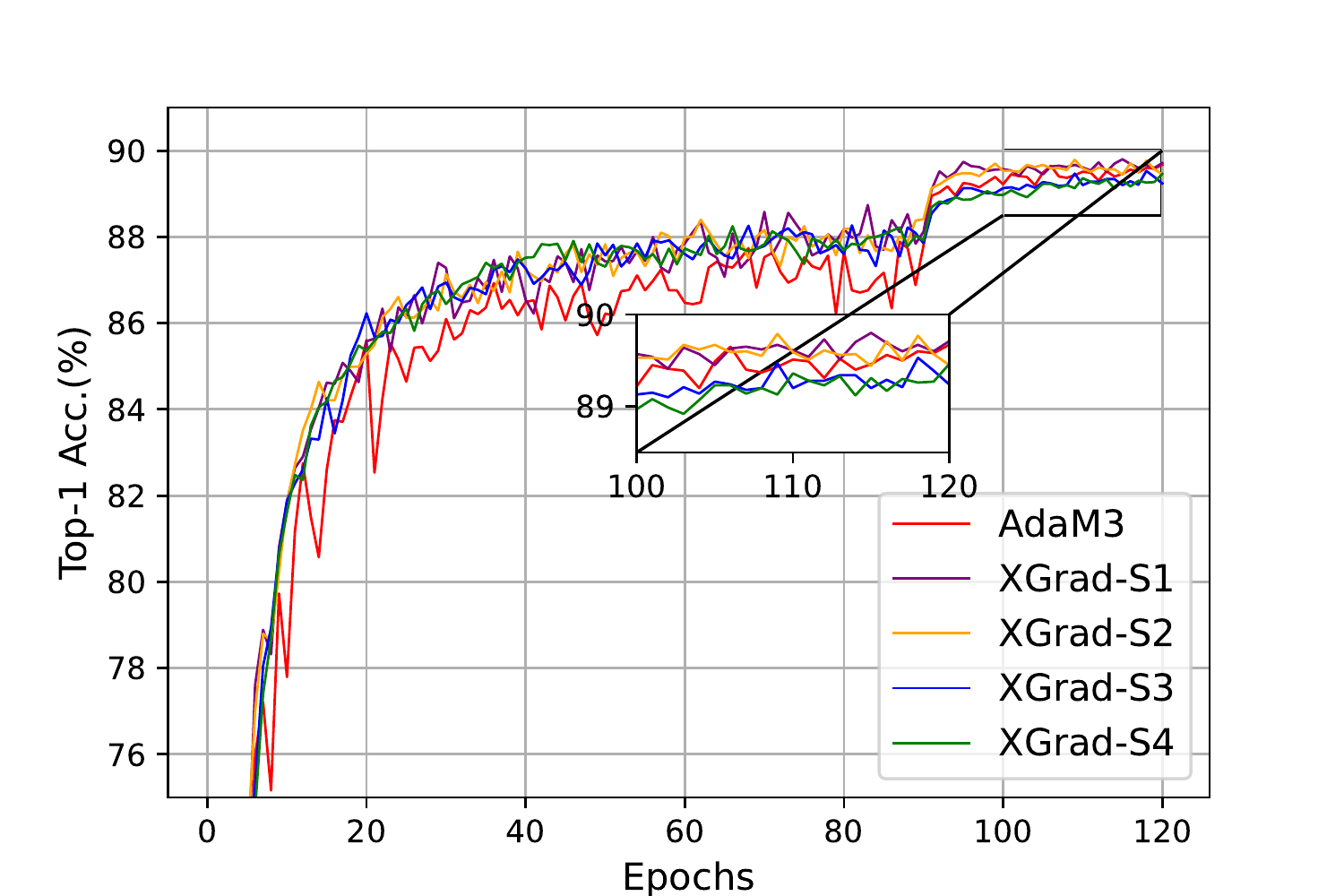}\label{comp-adam3-vgg11-acc}}
	\quad
	\subfloat[ResNet-34]{\includegraphics[width=.25\textwidth]{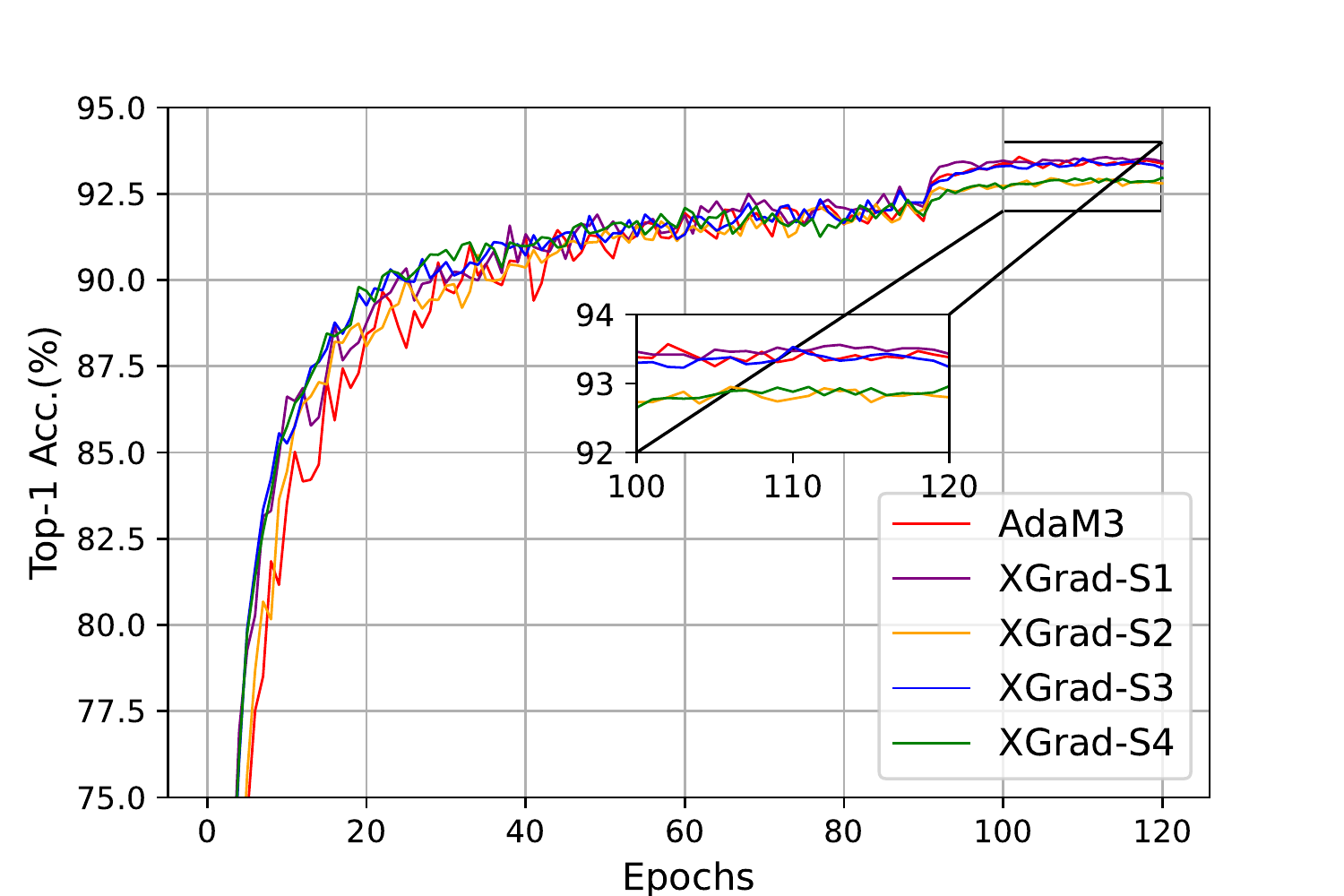}\label{comp-adam3-resnet34-acc}}
	\subfloat[DenseNet-121]{\includegraphics[width=.25\textwidth]{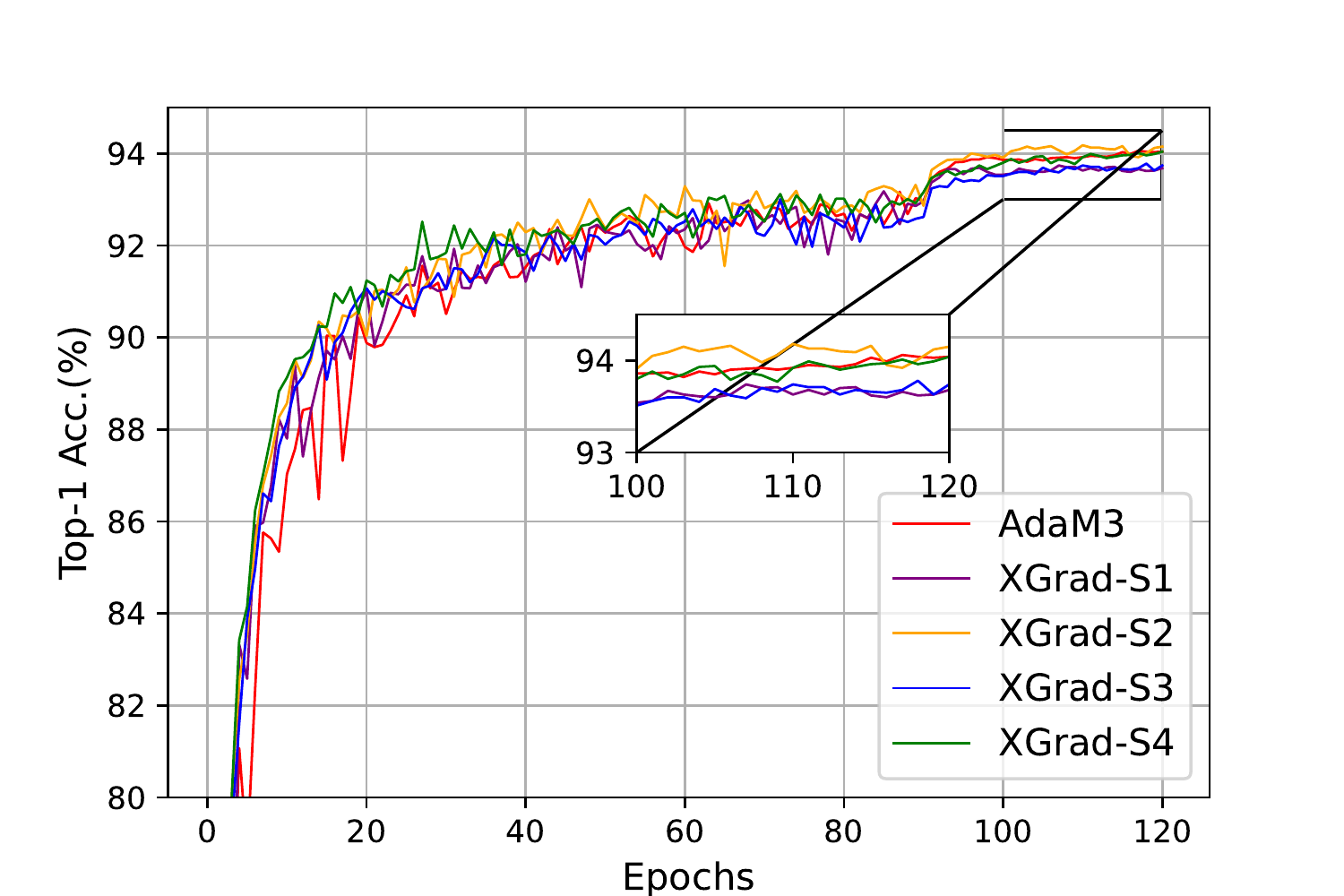}\label{comp-adam3-densenet-acc}}
	\subfloat[LSTM-1]{\includegraphics[width=.25\textwidth]{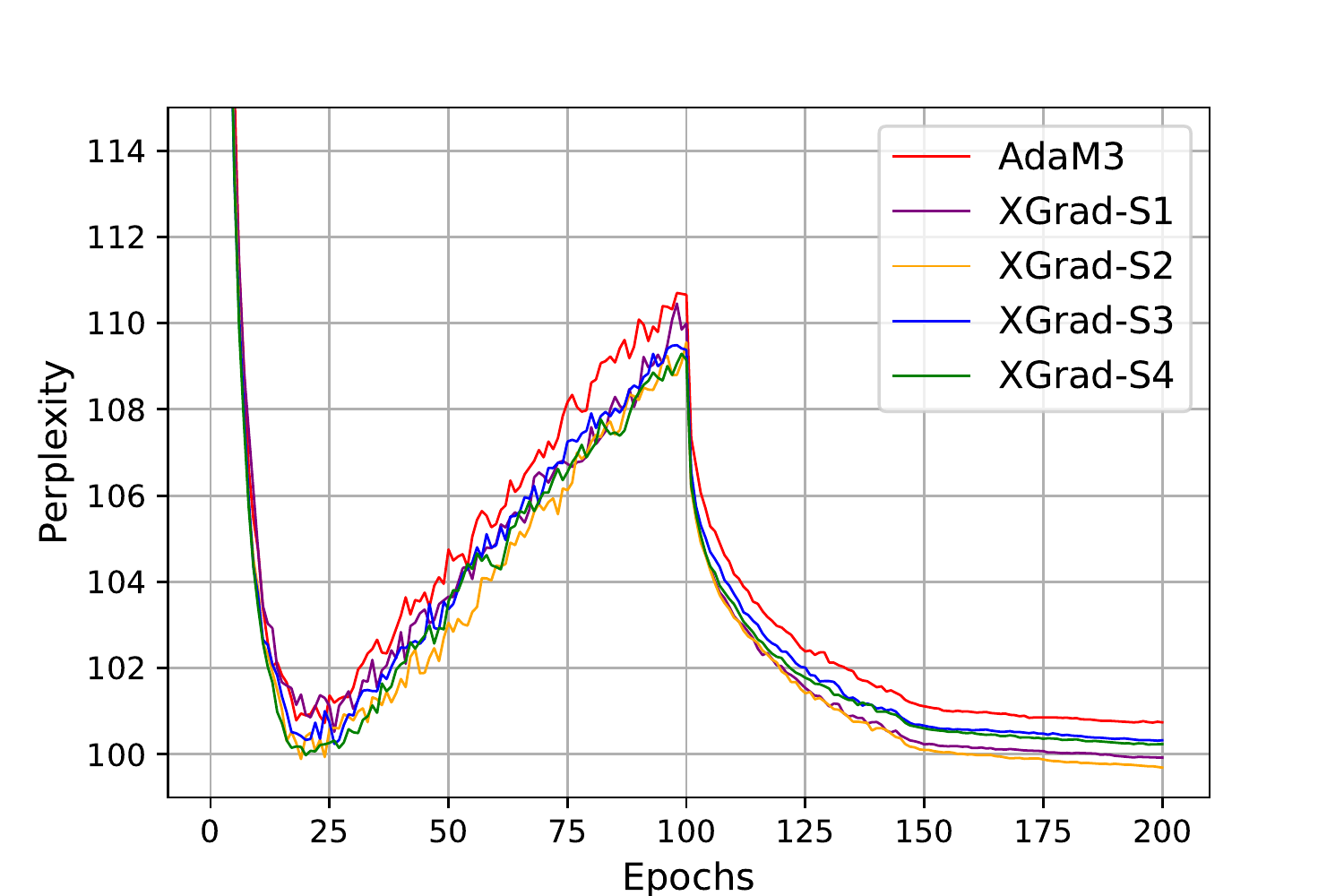}\label{comp-adam3-lstm1}}
	\quad
	\subfloat[LSTM-2]{\includegraphics[width=.25\textwidth]{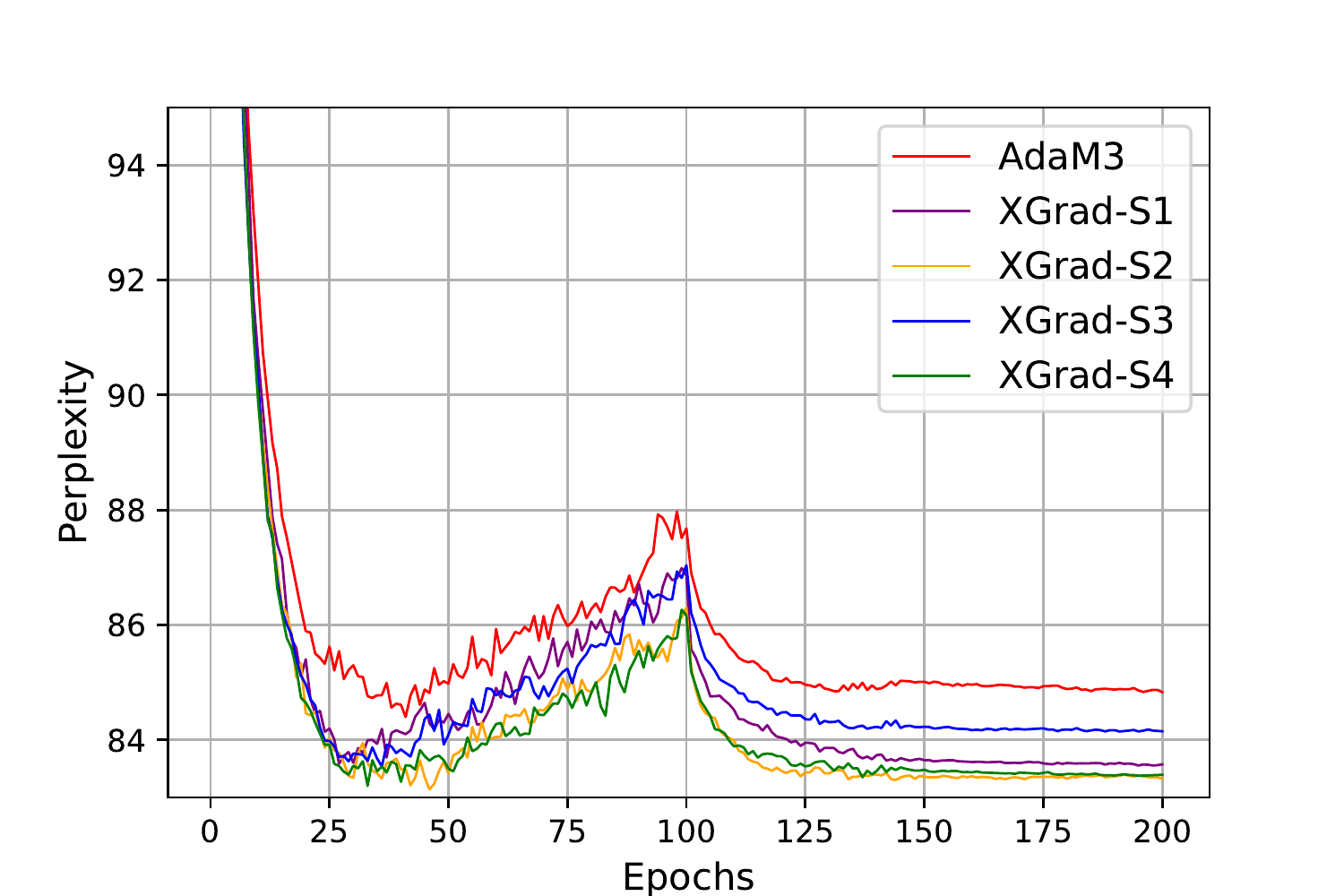}\label{comp-adam3-lstm2}}
	\subfloat[LSTM-3]{\includegraphics[width=.25\textwidth]{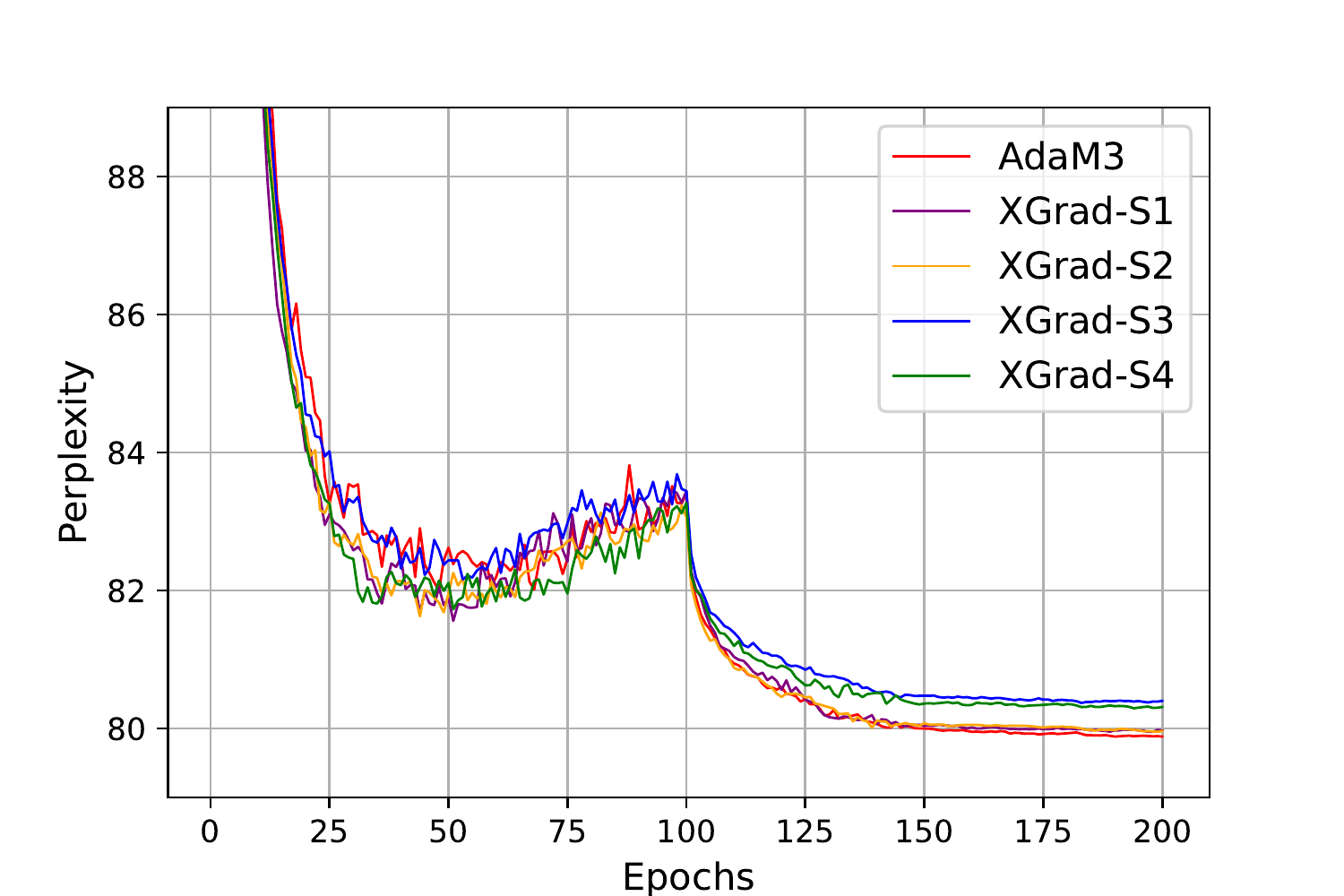}\label{comp-adam3-lstm3}}
	\subfloat[VAE]{\includegraphics[width=.24\textwidth]{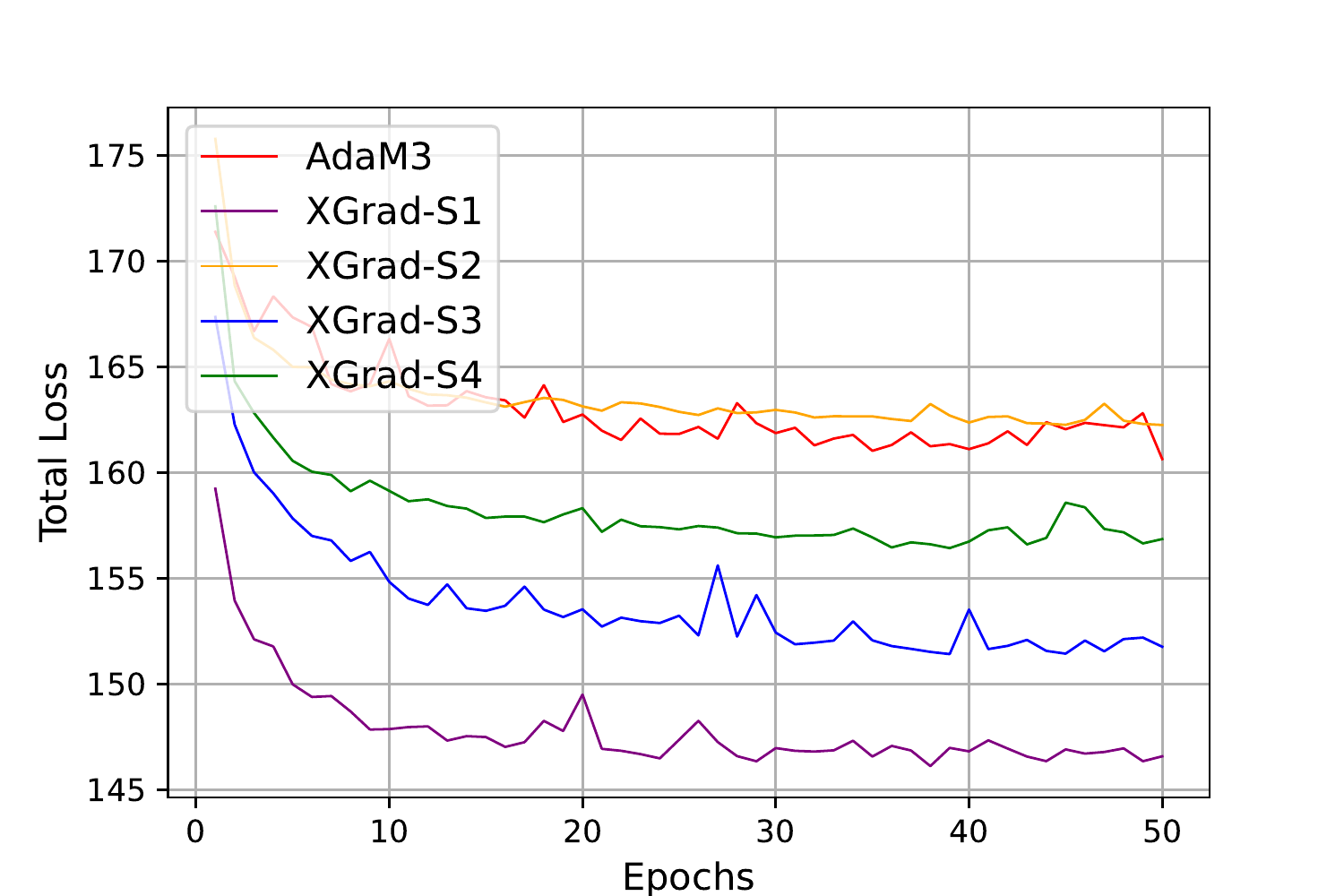}\label{comp-adam3-vae}}
	\caption{Comparison of AdaM3 and XGrad. Figures~\ref{comp-adam3-lenet-acc}, ~\ref{comp-adam3-alexnet-acc}, ~\ref{comp-adam3-vgg11-acc}, ~\ref{comp-adam3-resnet34-acc}, and~\ref{comp-adam3-densenet-acc}: Top-1 accuracy (higher is better) vs. Epochs; Figures~\ref{comp-adam3-lstm1}, ~\ref{comp-adam3-lstm2}, and~\ref{comp-adam3-lstm3}: Perplexity (lower is better) vs. Epochs; Figure~\ref{comp-adam3-vae}: Total loss (lower is better) vs. Epochs.}
	\label{comp-adam3-acc-cifar10}
\end{figure*}

As with Section~\ref{sec:comp-xgrad-adabelief}, we again selected LeNet, AlexNet, VGG-11, ResNet-34, DenseNet-121, LSTM-1, LSMT-2, LSTM-3, BERT$_{\text{BASE}}$, VAE, and WGAN as the benchmark DNN models. Figure~\ref{comp-adam3-acc-cifar10} depicts the learning curves about model accuracy vs. epochs and Table~\ref{table:adam3} summarizes the obtained best accuracy.

Similar conclusions can be drawn from the observation of the experiment results shown in Figure~\ref{comp-adam3-acc-cifar10} and Table~\ref{table:adam3}. Compared to AdaM3, XGrad generates a top-1 accuracy improvement of 0.89\%, 0.29\%, 0.13\%, -0.01\%, and 0.12\% when training LeNet, AlexNet, VGG-11, ResNet-34, and DenseNet-121, respectively. This gives rise to a top-1 accuracy improvement of 0.28\% on average. On language modeling tasks, XGrad achieves less perplexity than AdaM3 when training LSTM-1 and LSTM-2  and is slightly inferior to AdaM3 when training LSTM-3. Especially when training BERT$_{\text{BASE}}$, XGrad achieves significant accuracy improvement over AdaM3 (68.38 vs. 80.64). On image generalization tasks, XGrad gets 14.51 less total loss when training VAE and 22.29 less FID score when training WGAN. Again, the experiment results demonstrate that the model accuracy obtained by XGrad varies with the settings of the weight prediction step, especially when training WGAN.



\begin{table*}[h!]
	\centering
	\caption{Summarization of best model accuracy of XGrad and AdaM3. Maximum top-1 accuracy for LeNet, AlexNet, VGG-11, ResNet-34, and DenseNet-121; Minimum perplexity for LSTM-1/2/3; Maximum Dev set accuracy for BERT$_{\text{BASE}}$; Minimum total loss for VAE; Minimum FID score for WGAN. The best model accuracy results are highlighted in boldface.} 
	\label{table:adam3}
	\setlength{\tabcolsep}{2.0mm}
	\begin{tabular}{c|ccccccccccc}
		\toprule
		Optimizers & LeNet &AlexNet&VGG-11 & ResNet-34 & DenseNet-121 & LSTM-1  & LSTM-2 & LSTM-3 & BERT$_{\text{BASE}}$ & VAE   &WGAN     \\
		\midrule
		\multicolumn{12}{c}{Best Accuracy} \\
		\midrule
		\makecell{AdaM3} & 87.31\%& 89.96\%& 89.67\%  & \textbf{93.57}\%  &  94.06\% &   100.73 &  84.40 & \textbf{79.88}& 68.38\% &160.64  &  104.56    \\
		XGrad-S1 & 87.50\%& \textbf{90.25}\% &  \textbf{89.80}\%& 93.56\% & 93.74\% & 99.92 &  83.56& 79.95 & \textbf{80.64}\%  & \textbf{146.13} &  85.10 \\
		XGrad-S2& 87.61\%& 89.92\%&89.79\% & 92.95\% & \textbf{94.18}\% &   \textbf{99.69}& \textbf{83.14} &  79.95 & 80.39\%  & 162.25 & \textbf{82.27}   \\
		XGrad-S3 &\textbf{88.20}\%& 90.05\% &89.53\% & 93.53\% & 93.78\% &  100.24 & 83.53 & 80.37  & 83.82\% & 151.42 & 119.86    \\
		XGrad-S4& 87.30\% & 89.93\% & 89.46\% &  92.96\%&  94.04\%&  99.98  &  83.20&  80.29& 68.38\% & 156.43& 119.42    \\
		\bottomrule
	\end{tabular}
\end{table*}

\subsubsection{Discussion about the effectiveness of XGrad}
The experiment results demonstrate the effectiveness of XGrad in boosting the convergence and generalization of the gradient-based optimizers including SGDM, Adam, AdamW, AdaBelief, and AdaM3. In the vast majority of cases, XGrad tends to achieve better model accuracy than the base optimizers on all evaluated DNN models, verifying that XGrad can help the optimization method train to obtain more optimal DNN parameters.

However, applying weight prediction doesn't mean that the optimization approach can always move along the optimal trajectories.  Summing up the experiment results, we see that the performance of XGrad varies with the selection of weight prediction steps and the evaluated tasks. Different weight prediction steps may lead to quite a different model accuracy, and some choices of step size may even result in model accuracy lower than that of the base optimizer. For example, when comparing XGrad with Adam with the LSTM-1 model, XGrad-S1 achieves lower perplexity than Adam while XGrad with step sizes 2, 3, and 4 performs worse than Adam (see Table~\ref{table:adam}). Furthermore, the experiment results also demonstrate that on sophisticated tasks (e.g., BERT$_{\text{BASE}}$ and WGAN), the performance of XGrad varies sharply when selecting different weight prediction steps. This reveals that applying weight prediction is easy to mislead the optimization approach when training on sophisticated tasks.

We note that the superiority of XGrad over the base optimizers is validated through extensive experimental evaluations. For most cases, XGrad can obtain better model accuracy than the base optimizer without weight prediction. Studying the optimal weight prediction step size and the impact of complex training tasks on weight prediction performance will be the main focus of future research.






\begin{table*}[h!]
	\centering
	\caption{1-epoch running time (in seconds) comparison of XGrad vs. SGDM.}
	\label{table:comp-cost-sgdm}
	\setlength{\tabcolsep}{3.5mm}
	\begin{tabular}{c|ccccccc}
		\toprule
		Optimizers  & VGG-16 & ResNet-34 & ResNet-101 & GoogleNet & DenseNet-121 & Inception-V3 & ViT \\
		\midrule
		\makecell{SGDM} & \textbf{12.59}s  &  \textbf{36.20}s  & \textbf{131.73}s & \textbf{94.81}s & \textbf{82.41}s  &  \textbf{191.81}s & \textbf{40.34}s \\
		XGrad & 13.29s & 38.25s  & 139.09s  & 99.43s & 92.70s & 196.42s & 41.60s \\
		\bottomrule
	\end{tabular}
\end{table*}

\begin{table*}[h!]
	\centering
	\caption{1-epoch running time (in seconds/minutes) comparison of XGrad vs. Adam.}
	\label{table:comp-cost-adam-adamw}
	\setlength{\tabcolsep}{3.5mm}
	\begin{tabular}{c|ccccccc}
		\toprule
		Optimizers  & ResNet-34 & DenseNet-121 & Inception-V3 & ViT  & LSTM-1 & LSTM-2 & GNMT-8 \\
		\midrule
		\makecell{Adam} & \textbf{37.01}s  &  \textbf{95.34}s  & \textbf{200.63}s  &\textbf{40.72}s  & \textbf{21.03}s & \textbf{25.38}s& \textbf{386}m \\
		XGrad & 44.36s & 115.99s   & 207.78s & 43.54s & 24.96s & 26.43s & 400m \\
		\bottomrule
	\end{tabular}
\end{table*}

\begin{figure*}[h!]
	\centering
	\subfloat[XGrad vs. SGDM]{\includegraphics[width=.85\textwidth]{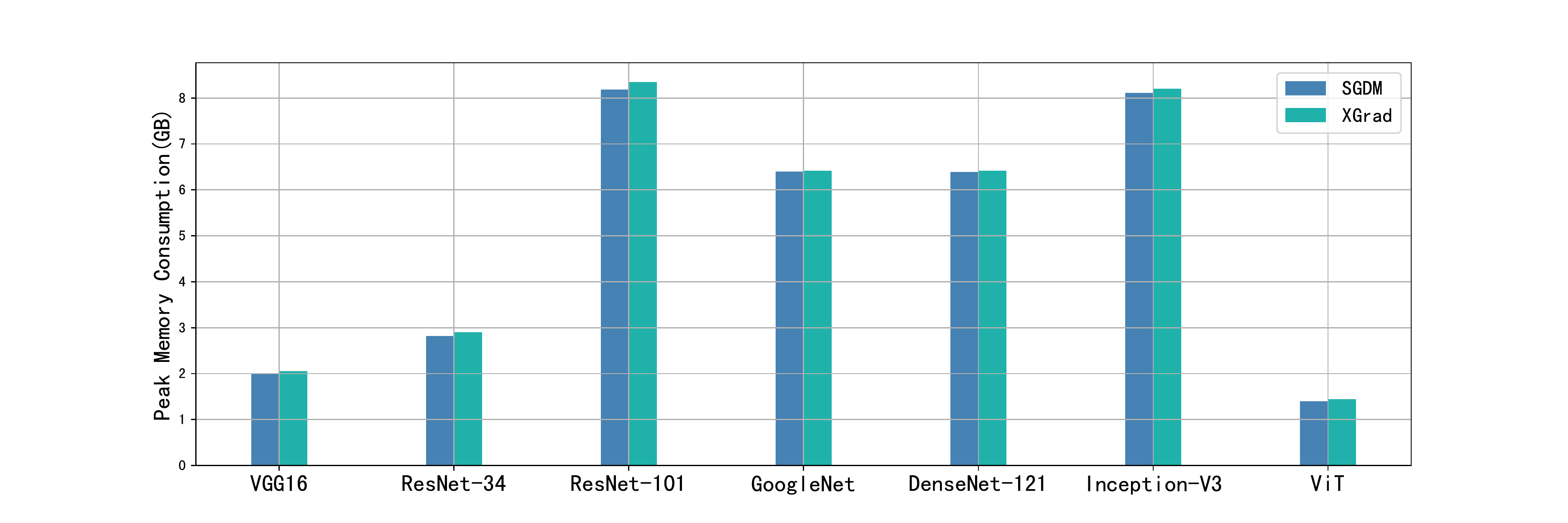}\label{comp-sgd-memory}}
	\quad
	\subfloat[XGrad vs. Adam]{\includegraphics[width=.85\textwidth]{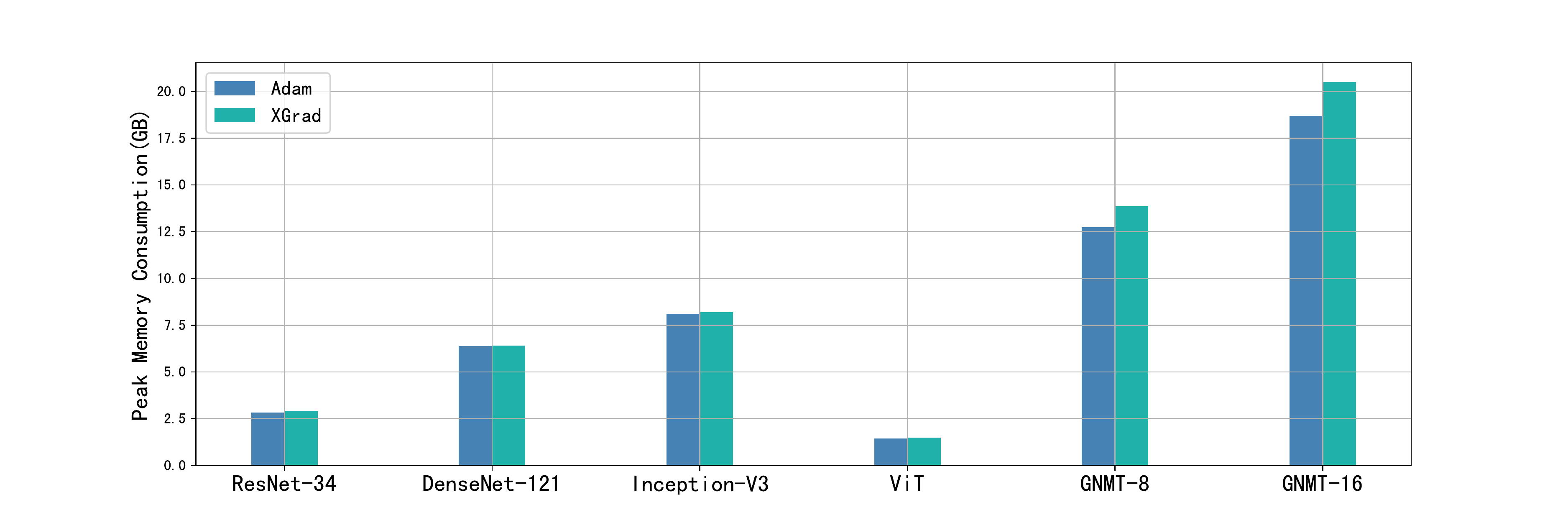}\label{comp-adam-memory}}
	\caption{Comparisons in terms of Peak GPU memory consumption (in GB). Figure~\ref{comp-sgd-memory}: XGrad vs. SGDM; Figure~\ref{comp-adam-memory}: XGrad vs. Adam.}
	\label{fig:comp-memory}
\end{figure*}

\subsection{Computational cost}\label{exp:comp_cost}
In this section, we evaluate the computational cost of XGrad.  We mainly compare XGrad with SGDM and Adam. For the comparison of XGrad and SGDM, we select VGG-16, ResNet-34, ResNet-101, GoogleNet, DenseNet-121, Inception-V3, and ViT as the benchmark models. For the comparison of XGrad and Adam, we select ResNet-34, DenseNet-121, Inception-V3, ViT, LSTM-1, LSTM-2, and GNMT-8 as the benchmark models. For the comparison of XGrad and SGDM, we trained all the DNN models on CIFAR-10  with the same experimental settings described in Section~\ref{sec:comp-xgrad-sgdm}. For the comparison of XGrad and Adam, we evaluated all the DNN models with the same experimental settings described in Section~\ref{sec:comp-xgrad-adam}. Table~\ref{table:comp-cost-sgdm} reports the running time of 1-epoch training when comparing XGrad with SGDM. Table~\ref{table:comp-cost-adam-adamw} summarizes the experimental results of the comparison of XGrad and Adam.

The experimental results shown in both Tables~\ref{table:comp-cost-sgdm} and~\ref{table:comp-cost-adam-adamw} demonstrate that XGrad always incurs longer training time than the original gradient-based optimizer. XGrad incurs an average of 5.67\%  (up to 12.48\%) longer training time than SGDM. In comparison with Adam, XGrad leads to 11.21\% longer training time on average. The experimental results are understandable as performing weight prediction in each iteration leads to additional computations compared to the original optimizers. Nevertheless, we see that the increased computations are quite limited and sometimes can even be neglected.


\subsection{Memory consumption}\label{exp:memory_cost}
In this section, we evaluate the memory consumption of XGrad. We mainly compare the peak GPU memory consumption of XGrad with that of SGDM and Adam. Likewise, the comparisons are divided into two groups: XGrad vs. SGDM and XGrad vs. Adam. For the comparison of XGrad vs. SGDM, we evaluated seven CNN models on CIFAR-10 including VGG-16, ResNet-34, ResNet-101, GoogleNet, DenseNet-121, Inception-V3, and ViT. For the comparisons of XGrad vs. Adam, the evaluated DNN models include ResNet-34, DenseNet-121, Inception-V3, ViT, LSTM-2, and GNMT-8.  The experimental settings are the same as in Section~\ref{exp:comp_cost}.

Figure \ref{fig:comp-memory} illustrates the peak GPU memory consumption of XGrad and the base optimizers including SGDM and Adam. From the observation of Figure \ref{comp-sgd-memory}, we can immediately reach the following conclusions. First, for any evaluated model, XGrad always consumes almost the same (only slightly more) GPU memory as SGDM. On average, XGrad only incurs 1.8\% (up to 3.0\%) more GPU memory than SGDM. We can draw similar conclusions from the experimental results shown in Figure \ref{comp-adam-memory}. XGrad leads to an average of 5.0\% (up to 9.7\%) more GPU memory consumption than Adam.

Overall, the experimental results shown in Figure~\ref{fig:comp-memory} reveal that XGrad incurs slightly more GPU memory consumption than the original baseline optimizers. This is reasonable because XGrad requires storing more intermediate variables in GPU memory to achieve weight prediction. However, we note that the increased GPU memory consumption is quite limited.






\section{Related Work}\label{sec:related_work}

The research on optimization methods always remains a hotspot in the field of deep learning as the convergence trait of the used optimizer directly affects convergence and the accuracy of the model. Generally, the training procedure of a DNN model is built upon backpropagation~\cite{hinton2006fast} where each layer of neural network weights is fine-tuned along the backward propagation of ``errors''.  Backpropagation always comes along with gradient descent. Currently, first-order gradient methods, such as SGDM~\cite{sutskever2013importance} and adaptive methods~\cite{zeiler2012adadelta,kingma2014adam,zhuang2020adabelief} are the most widely used deep learning optimization methods.


SGDM has been regarded as the default optimizer, especially for image classification tasks~\cite{simonyan2014very,szegedy2015going,he2016deep,krizhevsky2017imagenet}. 
The momentum term of SGDM accumulates the gradient information from all previous iterations, making the weights update along the way of inertia directions. Another noteworthy feature of SGDM is that a unified learning rate for all parameters is used throughout the entire training period.


Noticing that the learning rate plays a significant role in DNN training, many researchers turn to studying adaptive methods (also known as adaptive learning methods), which compute a specific learning rate for each individual parameter. In 2011, Duchi et al.~\cite{duchi2011adaptive} proposed AdaGrad, which dynamically adjusts the learning rate according to the history gradients from previous iterations and utilizes the quadratic sum of all previous gradients to update the model parameters. Zeiler~\cite{zeiler2012adadelta} proposed AdaDelta, seeking to alleviate the continual decay of the learning rate of AdaGrad. AdaDelta does not require manual tuning of a learning rate and is robust to noisy gradient information. Tieleman and Hinton~\cite{tieleman2012lecture} refined AdaGrad and proposed RMSprop. The same as AdaGrad, RMSprop adjusts the learning rate via element-wise computation and then updates the variables. One remarkable feature of RMSprop is that it can avoid decaying the learning rate too quickly. In order to combine the advantages of both AdaGrad and RMSprop, Kingma and Ba~\cite{kingma2014adam} proposed another famous adaptive gradient method, Adam, which has become the most important choice for training many Transformer-based DNN models~\cite{tieleman2012lecture,kingma2014adam,loshchilov2017decoupled}.  Loshchilov and Hutter~\cite{loshchilov2017decoupled} found that the major factor of the poor generalization of Adam is due to that $L_2$ regularization for it is not as effective as for its competitor SGDM. They further proposed decoupled weight decay regularization for Adam, which is also known as AdamW. The experimental results demonstrate that AdamW substantially improves the generalization performance of Adam and illustrates competitive performance as  SGDM~\cite{sutskever2013importance} when tackling image classification tasks. Currently, Adam as well as AdamW have become the default optimizer for DNN training. To simultaneously achieve fast convergence and good generalization, Zhuang \etal~\cite{zhuang2020adabelief} proposed another adaptive gradient method called AdaBelief, which adapts the stepsize according to the ``belief'' in the current gradient direction. Noticing that the second-moment estimation of Adam is a more favorable option for learning rate scaling than that of the raw gradient, Wang \etal~\cite{wang2021rethinking} proposed AdaMomentum. Reddi \etal~\cite{reddiconvergence} studied why RMSProp and Adam probably do not converge to an optimal solution and proposed a new exponential moving average variant: AMSGrad. Considering that Nesterov's accelerated gradient~\cite{nesterov1983method} has a better bound than gradient descent, Timothy Dozat incorporated Nestorv momentum into Adam and proposed a new optimization method named NAdam~\cite{dozat2016incorporating}. Other proposed adaptive methods include Yogi~\cite{zaheer2018adaptive}, AdaBound~\cite{luo2019adaptive}, RAdam~\cite{liu2019variance},  etc. Chen \etal~\cite{chen2018convergence} studied the convergence of adaptive gradient methods including the popular algorithms such as Adam, AMSGrad, and AdaGrad.

Notably, SGDM as well as all adaptive methods share a common feature: weight updates are performed in a continuous manner and each mini-batch training always utilizes currently available weights to do both forward pass and backward propagation. Weight prediction was previously used to overcome the weight inconsistency and weight staleness issues in the asynchronous pipeline parallelism. Chen et al.~\cite{chen2018efficient} used the smoothed gradient to replace the true gradient to predict future weights when using  SGDM~\cite{sutskever2013importance} as the optimizer. Guan et al.~\cite{guan2019xpipe} proposed using the update values of Adam~\cite{kingma2014adam} to perform weight predictions. Yet, both approaches use weight prediction to ensure effective parameter learning in the asynchronous pipeline training rather than studying the impact of weight prediction on the optimizers.

\section{Concluding remarks}\label{sec:conclusion}
To further boost the convergence and generation of popular gradient-based optimizers, in this paper, we introduce weight prediction into the DNN training and propose a new DNN training framework called XGrad. Our proposal can improve the convergence and generalization of popular gradient-based optimizers with a cost of slightly increasing the training time and GPU memory consumption. The remarkable feature of our proposal is that we perform both forward pass and backward propagation using the future weights which are predicted according to the specific update rule of the used optimizer. In particular, we construct the mathematical relationship between currently available weights and future weights and devise an effective way to incorporate weight prediction into DNN training. The proposed framework covers many of the most frequently used optimizers such as SGDM, RMSprop, Adam, AdamW, AdaBelief, and AdaM3. 

XGrad is easy to implement and works well in boosting the convergence of DNN training. Extensive experimental results on image classification, natural language processing, and image generalization tasks verify the effectiveness of our proposal. We believe that other adaptive gradient methods such as AdaBound, RAdam, Yogi, Lion~\cite{chen2023symbolic}, etc can be easily incorporated into our proposed framework.

For future work, we would like to explore the internal mechanism of weight prediction from a mathematical perspective. Furthermore, we will study the relationship between the weight prediction steps and the learning rate, as well as their impacts on the convergence and generalization of DNN training, especially when training on sophisticated tasks such as BERT and GAN.

\section*{Acknowledgments}
Lei Guan thanks Kangkang Deng at the National University of Defense Technology for stimulating discussions about the extragradient method and its relationship with XGrad. Lei Guan thanks Shaofeng Zhang at Ant Group for helping run the BERT benchmark.


%

\bibliographystyle{IEEEtran}
\bibliography{IEEEabrv, ref}


 
\vspace{11pt}

\vspace{11pt}


\vfill

\end{document}